\documentclass[twocolumn,english]{IEEEtran}
\usepackage{amsmath,graphicx}
\usepackage{blkarray}
\newcounter{example}[section]
\newenvironment{example}[1][]{\refstepcounter{example}\par\medskip
   \noindent \textbf{Example~\theexample. #1} \rmfamily}{ $\blacksquare$\medskip}

\DeclareMathOperator*{\minimize}{minimize}
\DeclareMathOperator*{\maximize}{maximize}

\DeclareMathOperator*{\argmin}{argmin}
\usepackage[T1]{fontenc}
\usepackage[nolist]{acronym}
\usepackage{babel}
\usepackage{units}
\usepackage{amsthm,amssymb,bm}
\usepackage{amsbsy}
\usepackage{algorithm,algorithmic}
\newtheorem{definition}{Definition}
\usepackage[unicode=true,
bookmarks=true,bookmarksnumbered=true,bookmarksopen=true,bookmarksopenlevel=1,
 breaklinks=true,pdfborder={0 0 0},backref=true,colorlinks=false]
 {hyperref}
\usepackage{aircraftshapes} 
\usepackage{tikz}
\usepackage{textcomp}
\usepackage{graphicx} 
\usepackage{epsfig} 
\usepackage{pgfplots}
\usepackage{caption}
\usepackage{subcaption}
\pgfdeclarelayer{foreground}
\pgfsetlayers{background, main, foreground}
\usetikzlibrary{quotes, angles, backgrounds, arrows, automata, shapes, positioning, calc, through, spy, decorations.markings}

\tikzset{
    imglabel/.style={
      rectangle,
      inner sep=2pt,
      text=black,
      minimum height=1em,
      text centered,
      fill=white,
      fill opacity=1.0,
      text opacity=1,
      anchor=south west,
    },
  }

\tikzset{
	state/.style={
		rectangle,
		draw=black, very thick,
		minimum height=1.0em,
		text centered,
	},
}

\title{When Robotics Meets Wireless Communications: An Introductory Tutorial}
\usepackage{amsfonts}
\usepackage{mathrsfs}
\usepackage{latexsym}
\usepackage{color}

\def\xx{\mathsf{x}}
\def\yy{\mathsf{y}}
\def\zz{\mathsf{z}}

\def\eu{\mathfrak}

\newcommand{\dif}[1]{\mathrm{d}{#1}}
\newcommand{\der}[2]{\frac{\dif{#1}}{\dif{#2}}}
\newcommand{\ders}[3]{\frac{\mathrm{d}^{{#3}}{#1}}{\mathrm{d}{#2}^{{#3}}}}

\author{Daniel Bonilla Licea$^{1,2}$\thanks{This work was partially funded by the European Union under the project Robotics and advanced industrial production (reg. no. CZ.02.01.01/00/22\_008/0004590), by the Czech Science Foundation (GAČR) under research project no. 23-07517S, and by CTU grant no SGS23/177/OHK3/3T/13.}~\IEEEmembership{Member,~IEEE}, Mounir Ghogho$^{{3,4}}$,~\IEEEmembership{Fellow~Member,~IEEE}, and Martin Saska$^{2}$, \\
\IEEEauthorblockA{{$^1$College of Computing, Mohammed VI Polytechnic University,
Ben Guerir, Morocco}\\$^2${Czech Technical University in Prague, Czech Republic}}\\
$^3${International University of Rabat, College of Engineering \& Architecture, TICLab, Morocco} \\
{$^4$ School of Electronic and Electrical Engineering, University of Leeds, UK }\\
daniel.bonilla@um6p.ma, m.ghogho@ieee.org, martin.saska@fel.cvut.cz}

\begin{document}
\maketitle
	\begin{acronym}
        \acro{AoA}[AoA]{Angle of Arrival}
        \acro{AoD}[AoD]{Angle of Departure}
        \acro{AoI}[AoI]{Age of Information}
	    \acro{AWGN}[AWGN]{Additive White Gaussian Noise}
        \acro{AF}[AF]{Amplify-and-Forward}
	    \acro{BS}[BS]{Base Station}
        \acro{BER}[BER]{Bit Error Rate}
        \acro{CaPP}[CaPP]{Communications-aware Path Planning}
        \acro{CaR}[CaR]{Communications-assisted Robotics}
  		\acro{CaTP}[CaTP]{Communications-aware Trajectory Planning}
        \acro{CNR}[CNR]{Channel-to-noise Ratio}
        \acro{DDR}[DDR]{Differential Drive Robot}
        \acro{DC}[DC]{Direct Current}
        \acro{DF}[DF]{Decode-and-Forward}
        \acro{HL}[HL]{Hovering Location}
        \acro{IoT}[IoT]{Internet of Things}
        \acro{ICC}[ICC]{Information Causality Constraint}
	    \acro{LoS}[LoS]{Line of Sight}
        \acro{LAP}[LAP]{Low-altitude aerial platform}
        \acro{MR}[MR]{Mobile Robot}
        \acro{MILP}[MILP]{Mixed-Integer Linear Programming}
        \acro{MIP}[MIP]{Mixed-Integer Programming}
        \acro{MTZ}[MTZ]{Miller-Tucker-Zemlin}
        \acro{PPP}[PPP]{Poison Point Process}
        \acro{PRR}[PRR]{Packet Reception Ratio}
        \acro{PWM}[PWM]{Pulse-Width Modulation}
        \acro{PDF}[p.d.f.]{Probability Density Function}
        \acro{PMF}[p.m.f.]{Probability Mass Function}
        \acro{POI}[POI]{Points of Interest}
        \acro{QoS}[QoS]{Quality of Service}
        \acro{RaC}[RaC]{Robotics-assisted Communications}
        \acro{RF}[RF]{Radio Frequency}
        \acro{ROS}[ROS]{Robot Operating System}
	    \acro{RSS}[RSS]{Received Signal Strength}
	    \acro{RSSI}[RSSI]{Received Signal Strength Index}	
	    \acro{SNR}[SNR]{Signal-to-Noise Ratio}
   	    \acro{SINR}[SINR]{Signal-to-Interference-and-Noise Ratio}
  	    \acro{TOMR}[TOMR]{Three Wheeled Omnidirectional Robot}    
	    \acro{TP}[TP]{Trajectory Planning}    
		\acro{UAV}[UAV]{Unmanned Aerial Vehicle} 		
 		\end{acronym}
\begin{abstract}
The importance of ground \acfp{MR} and \acfp{UAV} within the research community, industry, and society is growing fast. Nowadays, many of these agents are equipped with communication systems that are, in some cases, essential to successfully achieve certain tasks. In this context, we have begun to witness the development of a new interdisciplinary research field at the intersection of robotics and communications. This research field has been boosted by the intention of integrating \acp{UAV} within the 5G and 6G communication networks, and will undoubtedly lead to many important applications in the near future. Nevertheless, one of the main obstacles to the development of this research area is that most researchers address these problems by oversimplifying either the robotics or the communications aspects. Doing so impedes the ability to reach the full potential of this new interdisciplinary research area. In this tutorial, we present some of the modelling tools necessary to address problems involving both robotics and communication from an interdisciplinary perspective.
As an illustrative example of such problems, we focus on the issue of communication-aware trajectory planning in this tutorial.
\end{abstract}

\begin{IEEEkeywords}
Communication, control, trajectory planning, robot, UAV.
\end{IEEEkeywords}

\section{Introduction}
\label{sec:Intro}
Interdisciplinary research involving communications and robotics is gaining momentum, as evidenced by the steady increase in publications from the robotics~\cite{Gasparri2016TRO,Gao2017IEEETM, Varadharajan2020RAL,CATP19}, control \cite{CATP3,Muralidharan2017ACC,CATP18,Ali2019TCNS,Kantaros2019IEEETAC}, and communications~\cite{uav3, Wu2018TWC, mobb10,LiuICCC2019,Ahmed2020CL, Dabiri2020CL,Khalil2020ITGCN,Zhou2020ACM} communities, which are dealing with~\acfp{MR} and communications issues. Some reasons behind the growing interest include the emergence of 5G technology that aims to integrate~\acfp{UAV} into the cellular communication network~\cite{Mishra2020CN, uav26b} and the growing importance of robotic swarms \cite{Chung18TRO}. From an application perspective, we can divide these interdisciplinary problems into two categories: \acf{RaC} and \acf{CaR}. 
\par
In \ac{RaC} applications, generally one or multiple \acp{MR} are incorporated into a communication network with the intent to improve the performance of the latter; the \acp{MR} typically operate as mobile relays, data ferries, or mobile flying \acfp{BS} on-board \acp{UAV}. The main objective of \ac{RaC} is to control the behavior of the added \ac{MR} in order to improve the performance of the communications network. In traditional mobile communications, the transceiver's position is considered {\it uncontrollable} and random. In \ac{RaC}, the transceiver's position is {\it controllable}, and thus constitutes an additional communication system parameter to be optimized. This small, but important difference opens up new and exciting possibilities in the design of communication systems involving \acp{MR}. For example, in the context of diversity techniques for the compensation of small-scale fading, it is widely accepted that designing the diversity branches in such a way as to make the channels statistically independent maximizes the diversity gain, and thus performance. When using a transceiver mounted on an \ac{MR}, we demonstrated in \cite{MD2} that by controlling and adapting the \ac{MR}'s position, we can obtain diversity branches that yield a higher diversity gain than when the branches are designed to be statistically independent. This short example provides a brief glimpse into the possibilities of going beyond the  theoretical bounds derived from classical communications problems when the mobility of the transceiver is treated as an additional controllable parameter of the communications system.
\par
In \ac{CaR} applications, the communication capabilities of the \ac{MR} are leveraged to help the robotic system to better perform tasks. Communication is an essential component that enables multi-robot applications~\cite{CATP2} and~\ac{UAV} swarms~\cite{Chung18TRO,UAV25,Varadharajan2020RAL}. Communication between the \acp{MR} allows the exchange of different types of information, such as: relative localization, allowing for the creation of formations or the navigation of surroundings in a coordinated manner; sensing data, which could be used in mapping applications; and signalling data, used to coordinate the behaviour of a robotic team. Often in \ac{CaR} applications, the \acp{MR} must execute certain robotic tasks while considering the communication quality and swarm connectivity to ensure adequate behaviour of the robotic team. 
\par
To efficiently address the problems encountered in \ac{RaC} and \ac{CaR} applications, we argue that a good understanding of {\em both} communications and robotics is required. Indeed, as will be described later, there is often a deep entanglement between the communications and robotics aspects in such problems. 
Considering only the communications aspects and oversimplifying the \ac{MR}'s model might produce an energy inefficient solution that wastes too much energy in motion, or even an unfeasible solution for a real \ac{MR} due to a breach of its own mechanical constraints. On the other hand, focusing on the robotics aspects and oversimplifying the communication model may lead the \ac{MR} to fail to complete its task due to unexpected communication failures, which may arise from poorer connectivity or a lower bit rate than those expected with the oversimplified model. Therefore, an interdisciplinary approach is essential when dealing with \ac{CaR} and \ac{RaC} problems in order to propose functionally adequate solutions and to fully exploit all underlying opportunities in this new research area.  
The importance of such an interdisciplinary approach has also been recently recognized in \cite{SIMULATIONS1,SIMULATIONS2}, where the authors proposed a simulation framework that allows for coordinating a robotics simulator (e.g. \ac{ROS}), a communications network simulator, and an antenna simulator. This enables to accurately simulate the dynamics of the robot and the communications channel.
\par
In the literature, however, \ac{CaR} and \ac{RaC} problems are often not addressed with such an interdisciplinary approach. Indeed, oversimplified models are often adopted for either the communications or the robotics aspects. This oversimplification causes the researchers to miss interesting results and opportunities, or even to derive techniques that would fail when tested on real robots equipped with real communication systems. 
Some tutorials have recently been published on communications-aware robotics problems. However, they still simplify either the robotics or communications aspects. For instance, the tutorial \cite{uav26b} discusses \ac{UAV} communications with great detail, but treats the control and robotics aspects in a superficial manner. The authors in \cite{uav26b} mention that, to the best of their knowledge, no rigorous expression for the \ac{UAV} energy consumption for a given trajectory has been derived. As we will show in subsection \ref{sec:uav}, such a statement is imprecise and comes from a lack of understanding of the \ac{UAV} dynamic models and control theory. On the other hand, the robotics community generally oversimplifies the communication model. For instance, the authors of \cite{DiskModel1} consider the problem of a team of data-gathering \acp{MR} and assume a binary disk model \cite{DiskModel0} for the communication channel. In this model, the communication is perfect as long as two \acp{MR} remain within a certain distance of each other. Such a model is far from reality, as we shall see in section \ref{Comm:channel}.
\par
To our knowledge, in the literature, there are no surveys or tutorials which have taken an interdisciplinary approach to address \ac{CaR} or \ac{RaC} problems. This tutorial aims to contribute to closing this gap and raising awareness of the importance of an interdisciplinary approach when tackling these problems. To illustrate the opportunities and challenges of this approach, we focus on the important problem of \acf{TP} within the context of \ac{CaR} and \ac{RaC} applications, henceforth referred to as \acf{CaTP}. It is our hope that this tutorial will inspire new research in this exciting and promising, yet underdeveloped area. 

{ It is noteworthy that this tutorial serves as an introductory exposition, omitting advanced wireless communication technologies such as multi-antenna communication, which is often the focus of 5G and 6G research. Despite this intentional exclusion, a measure taken to simplify notations and derivations, the content of this tutorial strives to endow the reader with foundational knowledge crucial for the exploration of issues related to CaR and RaC applications within the contexts of 5G and 6G systems. Additionally, it is pertinent to underscore that several scholarly works focused on 6G integrate certain channel models elucidated within the scope of this tutorial.}

\par


\par
{  This tutorial is organized as follows. In section \ref{sec:RMod}, we describe in detail the basic and crucial mobile robotics aspects and models. Section \ref{sec:Mod:comms} describes the different aspects of communications systems that are relevant to \ac{CaTP} problems and discusses modelling the wireless communication channels. In section \ref{sec:CATP}, the general structure of \ac{CaTP} problems, the different types of optimization targets and the constraints involved in these problems are described. In section \ref{sec:ExamplesDetails}, we present the structure of the \ac{CaTP}, discuss a variety of examples, and present some open research problems and opportunities related to \ac{CaTP}. In Section \ref{sec:summary}, the reader can find a concise summary of \ac{CaTP} structure that can be used as a navigation tool within this tutorial. Finally, section \ref{sec:Discussion} provides conclusions.}
\section{Mobile Robot Modeling}
\label{sec:RMod}
\par
Mathematically modelling a phenomenon or object consists of describing it in terms of certain aspects of interest and under certain conditions. We can divide the model into the following four components.

{\bf Object representation}: in our case, the object to be modelled is the \ac{MR}. The object representation is a mathematical abstraction of the physical \ac{MR} that describes properties that are relevant for the problem at hand. In the context of trajectory planning problems, the MR can be described as a single point\footnote{This holds only if the obstacles and the other \acp{MR} are dilated to avoid collisions with the considered \ac{MR}.} with orientation, even if the real object itself occupies non-zero physical space. This is a suitable representation of the MR for trajectory planning problems, although it may be inadequate for other problems, such as the mechanical analysis of the MR's frame.

{\bf Model Input}: the model input is a set containing all controllable variables and whose effect on the object representation will be captured by the mathematical model. There might be other controllable variables which affect the object representation, but the model input includes only the variables considered in the mathematical model.

{\bf Range of validity}: this consists of all of the conditions under which the model adequately describes the behaviour of the object representation, e.g. the assumptions on the considered scenarios and ranges of variations for the variables forming the model input. The model input together with the range of validity form the {\bf model input space}. This space is the set of all valid values that the model input can take.  

{\bf Input-to-object relation}: this is the mapping rule that relates any element within the model input space to a corresponding element within the {\bf object representation space}. In the case of analytical models, this relation can take the form of explicit mathematical equations, while in the case of numerical methods, it can take the form of numerical-lookup tables or neural networks. \qedsymbol{}

\par
Simplicity is always a desirable property in modelling as it brings tractability, facilitates mathematical analysis, lowers computational requirements, and allows for better problem understanding. One of the fundamental problems in modelling is finding a compromise between simplicity and accuracy that reflects the difference between the behaviour of the mathematical model and the real object. Simple models tend to be highly inaccurate and have reduced ranges of validity; highly accurate models with large ranges of validity tend to be very complex and have little tractability, hence the importance of selecting an adequate level of model complexity. Oversimplification of the model occurs when it is used beyond its range of validity or when variables that have a significant impact on the object representation are disregarded.
\par
Let us analyse this problem in more detail by discussing three important consequences of using oversimplified models for robots in the context of \ac{CaTP}.

{\bf Involuntary energy waste}:
oversimplifying the motion-induced energy consumption model while searching for minimum energy trajectories may significantly degrade performance. This is illustrated with the scenario described in \cite{UAV6}, where a rotary-wing \ac{UAV} (i.e. a \ac{UAV} with propellers and thus hovering capability) has to collect data from a sensor network. The authors optimize the \ac{UAV} trajectory to accomplish this task in minimum time and assume that the energy consumed by the \ac{UAV} is proportional to the time that the \ac{UAV} spends in the air. In other words, the authors implicitly model the \ac{UAV} energy consumption as being proportional to the flying time, thus expecting that a real \ac{UAV} following the trajectory optimized according to the assumed energy model would indeed accomplish its assigned task while draining the minimum energy from its battery. 
\par
For those who are unfamiliar with aeronautics or with aerial robotics, the energy model mentioned above might seem reasonable. However, in \cite{UAV11}, the authors present an aerodynamic power consumption model for a rotary-wing \ac{UAV} that shows that the power consumption also depends on the horizontal speed of the \ac{UAV}. This model contains a term that increases with the horizontal \ac{UAV} speed due to the blade's drag. Hence, a \ac{UAV} executing the minimum time trajectory will travel as fast as possible and consume a large amount of energy instead of minimizing energy consumption. {Furthermore, in \cite{DaiIEEETWC2023}, the authors demonstrate the benefits of incorporating wind effects into the energy consumption model for rotary-wing aerial vehicles operating in windy conditions. 
Using simulations, they found that a multirotor's trajectory optimized according to a wind-dependent model is 100\% more energy efficient than a model that ignores the wind effect.} 
\par
{\bf ii. Unfeasibility of designed trajectory}: another problem of oversimplified models is the possibility of obtaining a trajectory that does not satisfy robot motion constraints. This may lead to collisions with obstacles or among robots, as well as failures to perform the planned tasks. This occurs when the \ac{MR}'s motion model disregards the specific kinematic or dynamic constraints of the MR under consideration.
\par
For instance, assume that we want to design a trajectory for a car-like MR to pass in a predefined order through a set of points of interest $\{\mathbf{o}_k\}_k^K$ in a minimum time. We first model the car-like \ac{MR} with a single integrator, i.e. the position $\mathbf{p}(t)$ of the robot at time $t$ is described by:
\begin{equation}
\label{eq:1}
\mathbf{p}(t)=\mathbf{p}(0)+\int_{0}^t\mathbf{v}(\tau)\mathrm{\tau},
\end{equation} 
where $\mathbf{p}(0)$ is the \ac{MR}'s initial position and $\mathbf{v}(\tau)$ is its velocity at time $\tau$, which can be directly controlled. To make the model more realistic, we take into account the maximum translational speed of the robot, and thus add the following constraint $\|\mathbf{v}(\tau)\|_2<V_{max}$. According to this model, the optimum trajectory that allows the \ac{MR} to pass through all the points of interest in the minimum time is a piece-wise linear path linking consecutive points of interest, requiring the \ac{MR} to use the maximum linear speed possible, $V_{max}$, at all times. 
\par
Following this piece-wise linear path continuously at a constant linear speed without stopping until reaching the end of the path would require, in general, that the \ac{MR} instantaneously changes direction at each point of interest without stopping. However, this is an impossible manoeuvre for a car-like \ac{MR} due to its dynamic and kinematic constraints. Its dynamic constraints will not allow abrupt changes in velocity (whether the change is in direction or in magnitude) and its kinematic constraints will limit the curvature of the turns it can execute. Now, assume that we allow the \ac{MR} to stop at each point of interest of the piece-wise linear path. This would now require the robot to rotate on the spot at each point of interest in order to change direction and then continue its path. Yet, this manoeuvre is also impossible for the car-like MR due to the kinematic constraints imposed by its own physical geometry.
\par
Clearly, the model (\ref{eq:1}) used in this example oversimplifies the motion capabilities of the car-like \ac{MR}. Thus, the solution for the trajectory planning problem derived from such a model results in an unfeasible trajectory for the real car-like \ac{MR}. Some post-processing, e.g. smoothing the path, could be done on the resulting trajectory to make it feasible for the real \ac{MR}, however, this would still require considering the \ac{MR}'s kinematic and/or dynamic constraints. 
\par

\par
{
{\bf Reduced performance}: 
A further consequence of oversimplifying models is a degradation in performance. Some impairments may not be captured by overly simplified models. Thus, oversimplified models may cause performance degradation in some cases due to unexpected impairments not predicted by the model.
\par
For instance, in \cite{BonillaIEEECL2023}, we optimized the trajectory of a fixed-wing \ac{UAV} acting as a communication relay between two ground nodes. We used aerodynamic equations to model the \ac{UAV} motion, and we considered the position of the antenna on the UAV airframe. This model allows to predict the airframe-shadowing phenomenon (a blockage of the \ac{LoS} between the \ac{UAV} and the ground nodes produced by the \ac{UAV} airframe). 
The trajectory optimized with this elaborated model was shown to be up to 143.16\% more energy efficient than one optimized with an oversimplified model that is unable to predict airframe shadowing. 
\qedsymbol{}
\par
{In some cases, increasing the complexity of the MR models may bring only small benefits to the performance of the system while increasing significantly the complexity of the problem}. For instance, in \cite{Rob2}, the authors considered a three-wheeled omnidirectional \ac{MR} and compared two approaches of finding a minimum energy trajectory. The first trajectory was optimized using a certain dynamic model of the robot; the second trajectory used the same dynamic model of the robot, except that it neglected the term accounting for the Coriolis force. Both trajectories were then tested on a real \ac{MR}. The results showed that when tracking the first trajectory optimized with the simpler dynamic model, the \ac{MR} consumed just slightly more energy than when it tracked the trajectory optimized with the more complex model. 
\par
As we have explained above, the oversimplification of MR models can have serious consequences, thus the importance of selecting an adequate model complexity. In order to help researchers with no (or little) robotics background, the rest of this section provides a general description of mathematical models describing the motion and energy consumption for three popular MRs: ground wheeled robots, rotary-wing(s) aerial robots, and fixed-wing aerial robots. We will present an overview of the different types of models, their implications, and their limitations. This does not constitute an exhaustive list of all the mathematical models associated with these MRs, but rather an introductory presentation for common models used in trajectory planning. Readers familiar with this research area can skip to section \ref{sec:Mod:comms} where we discuss the modelling of communication systems and the wireless channel.
\par
We must mention that selecting the appropriate complexity of the models is not always an easy task and sometimes does not have a clear, or even a unique {\it correct} answer. This is highly dependent on the particular and specific conditions of the problem to be solved. In this tutorial, we aim to raise awareness about the problems related to oversimplification, while underlining that the solution to the latter is not an overcomplexification. Indeed, overly complex models might solve some issues due to oversimplification, but may also cause other problems, such as a lack of tractability and the high computational burden  of the solutions.
\subsection{Wheeled Mobile Robots (WMRs)}
\label{sec:RMod:GWR}
\par
WMRs \cite{Mob3} can be implemented in a simple and relatively inexpensive way, making them highly popular for numerous applications. They can be classified according to the number of wheels, their wheel type, and spatial configuration within the WMR's frame. These aspects determine the kinematic constraints of the robot. Therefore, before discussing the kinematic model, we briefly present the wheels used in typical WMRs. Some common types include \cite{Rob1}:
1) {\bf Conventional wheels}: these wheels have two degrees of freedom (DOFs), including rotation around its axis line and rotation around its contact point with the floor.
2) {\bf Omnidirectional wheels}: this is a special type of wheel constructed using a main conventional wheel with embedded small rollers whose orientation is not aligned with that of the main wheel. This special configuration allows the omnidirectional wheel to operate not only as a conventional wheel, but also to advance along the direction orthogonal to the main wheel's orientation. These wheels can be actuated (i.e. mechanically coupled to a motor's shaft).
\par
Kinematics is the field of mechanics that studies the motion of objects without considering the effect of forces. WMR kinematic models describe the relationship between the angular speed of the WMR's actuated wheels and the velocity of the WMR's object representation, i.e., the translational and angular velocities of $\mathbf{p}$. There are two types of kinematic models: the forward kinematic model and the inverse kinematic model. Forward kinematic models describe the WMR's velocities in terms of the angular speeds of the wheels, i.e. the wheels' angular speeds are the model's inputs, and the WMR's velocities are the model's output. Inverse kinematic models describe the opposite relation: the WMR's velocities are the model's inputs, and the angular speeds of the wheels are the model's outputs. The forward kinematic model is useful for analyzing the movement of the WMR generated by a particular control signal applied to its motors. The inverse kinematic model is useful to make the WMR's controller track a desired trajectory (i.e. the model's input is the desired trajectory and the model's output is the required wheels' angular speeds).
\par
In the following sections, we will discuss the above-mentioned kinematic models for various WMRs.
\subsubsection{Forward Kinematic Model}
\label{sec:MR:Kin}

As mentioned before, although WMRs are not single points in the plane, but rather systems that cover a certain area, their object representation consists of their horizontal position $\mathbf{p}$ and orientation $\phi$ (WMRs usually operate in 2D). The point $\mathbf{p}$ is a point on the WMR's frame whose exact location is selected to facilitate kinematic and dynamic analysis. For instance, in the \ac{TOMR} of Fig. \ref{Figure2}, the point $\mathbf{p}$ is chosen to be the geometrical center of the MR, but in the \ac{DDR}, the point $\mathbf{p}$ is the center of the axis-line of the actuated wheels, rather than the geometrical center of the robot, see Fig. \ref{Figure1}. Other choices for $\mathbf{p}$ are also possible, but these would complicate the mathematical analysis of the motion models.
\par
Each type of wheel and their location within the WMR's frame determines the forward kinematic model, and each type imposes different constraints. For instance, a conventional wheel is allowed rotational slippage around its contact point with the floor, but is not allowed translational slippage. The key hypotheses in the kinematic model derivation are the following: (i) the WMR's frame is solid and suffers no deformation; (ii) all the wheels are in contact with the floor at all times, which implies the consideration of a flat floor; (iii) each wheel is in contact with the floor only at a single point. 
For more details about the derivations of forward kinematic models, the interested reader can consult \cite{Rob1}.
\par
{
In terms of number of wheels, the unicycle is the simplest WMR whose kinematic model is described in \cite{bookR1}. Then we have the bicycle which is a vehicle that has a back conventional wheel and a front steerable conventional wheel. Its kinematic model will depend if the driving wheel is the front wheel or the back wheel, see \cite{bookR1} for more details. }
\par
\begin{figure}[htp!]              
\centerline{{\includegraphics[scale=0.3]{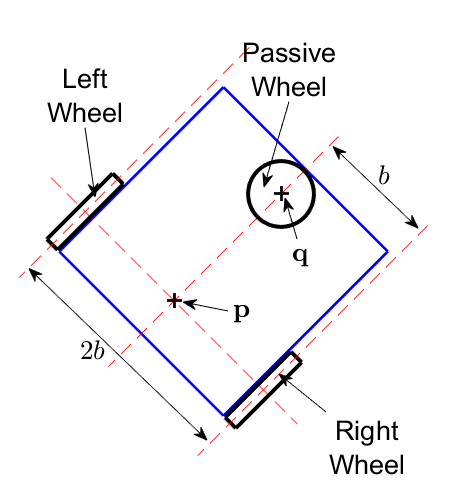}}}\vspace{-0.25cm}
    \caption{
   \small{Differential Drive robot diagram.}}
  \label{Figure1}
\end{figure}
Now, we consider the \ac{DDR}, see Fig. \ref{Figure1}. This robot has two conventional wheels, of radii $r$, which are actuated by separate motors. The horizontal position of the \ac{DDR} is determined by the point $\mathbf{p}$ located in the center of the line linking the points of contact of both wheels with the floor. The distance between these points is $2b$. A third unactuated wheel making contact with the floor at point $\mathbf{q}$ provides mechanical stability; it is chosen so that it does not impose any kinematic constraints. The corresponding kinematic model for the \ac{DDR} is \cite{Rob5}:
\begin{equation}
\label{kin:DDR}
\left[
\begin{array}{c}
\dot{x}\\
\dot{y}\\
\dot{\theta}
\end{array}
\right]
=
\frac{r}{2}
\left[
\begin{array}{cc}
\cos(\theta)& \cos(\theta)\\
\sin(\theta) & \sin(\theta)\\
\frac{1}{b} & -\frac{1}{b}
\end{array}
\right]
\left[
\begin{array}{c}
\omega_R\\
\omega_L
\end{array}
\right],
\end{equation}
where $\omega_R\in\mathbb{R}$ and $\omega_L\in\mathbb{R}$ are the angular speeds of the right and left wheels, $x$ and $y$ are the $\mathrm{x}$ and $\mathrm{y}$ positions of the \ac{DDR}, respectively, and $\theta$ is its orientation. The model input vector in (\ref{kin:DDR}) is two dimensional, while the \ac{DDR} configuration (i.e. the output vector) is three dimensional. This implies that the \ac{DDR} translational velocity (i.e. $[\dot{x},\ \dot{y}]$) and its angular speed $\dot{\theta}$ cannot be controlled independently. The \ac{DDR} translational speed (i.e. the magnitude of the \ac{DDR} translational velocity) and its angular speed are expressed as:
\begin{equation}
\label{kin:diff}
\left[
\begin{array}{c}
v\\
\dot{\theta}
\end{array}
\right]
=
\frac{r}{2}
\left[
\begin{array}{cc}
1&1\\
1&-1
\end{array}
\right]
\left[
\begin{array}{c}
\omega_R\\
\omega_L
\end{array}
\right].
\end{equation}
Note that while the model (\ref{kin:DDR}) is not invertible, the model (\ref{kin:diff}) is invertible. This implies that we can fully set independently the angular speed and translational speed of the robot, but we cannot fully set independently its angular speed and its translational velocity.
\par
{
Another common WMR is the car-like robot. It has two rear conventional wheels which are mechanically coupled to the same motor, and also has two conventional steerable front wheels which are mechanically coupled. Its kinematic model is described in \cite{Rob4}.}
\begin{figure}[htp!]              
\centerline{{\includegraphics[scale=0.17]{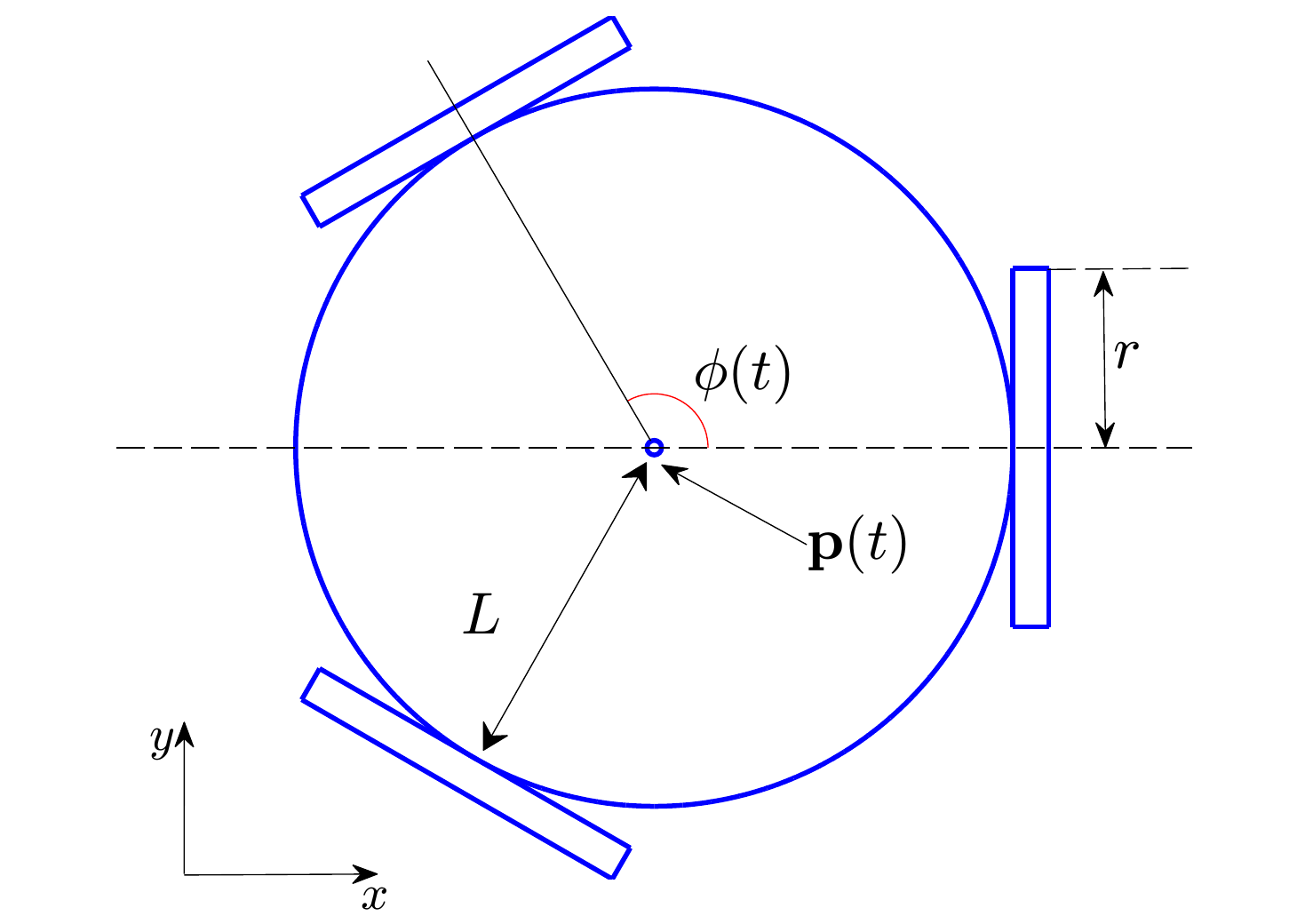}}}\vspace{-0.25cm}
    \caption{
   \small{Three-wheeled omnidirectional robot diagram.}}
  \label{Figure2}
\end{figure}
\par
Another class of WMR includes omnidirectional robots. They have the particularity of using omnidirectional wheels to enable the robot's omnidirectional motion, i.e., moving in any direction at any time. One popular MR belonging to this class is the \ac{TOMR}, see Fig. \ref{Figure2}. Its kinematic model is \cite{Rob3}:
\begin{equation}
\label{kin:TOMR}
\left[
\begin{array}{c}
\dot{x}\\
\dot{y}\\
\dot{\phi}
\end{array}
\right]
=
r
\left[
\begin{array}{ccc}
0&1&L\\
-\sin\left(\frac{\pi}{3}\right) & -\cos\left(\frac{\pi}{3}\right)&L\\
\sin\left(\frac{\pi}{3}\right) & -\cos\left(\frac{\pi}{3}\right)&L
\end{array}
\right]^{-1}
\left[
\begin{array}{c}
\omega_1\\
\omega_2\\
\omega_3
\end{array}
\right],
\end{equation}
where $r$ is the radii of the omnidirectional wheels, $\omega_i$ is the angular speed of the $i$th wheel, and $L$ is the distance from the wheels to the robot's center. Unlike the kinematic models of the \ac{DDR} and of the car-like robot, that of the \ac{TOMR} is fully invertible analytically. Hence, we can always independently determine the translational velocity and the angular speed of the MR to find the corresponding exact angular speed of each wheel. Therefore, the \ac{TOMR} does not present any kinematic restriction on its motion, but dynamic constraints still need to be considered, as will be described in section \ref{sec:MR:Dyn}.
\par
\par
\subsubsection{Inverse Kinematic Model}
\label{sec:MR:InKin}
As explained in the previous section, the forward kinematic model is obtained by analysing the MR's geometry and applying the motion constraints imposed by the wheels. The inverse kinematic models are obtained by inverting the forward kinematic model. 
\par
In some cases, the derivation of the inverse kinematic model is straightforward. For instance, consider the kinematic model of the \ac{TOMR} (\ref{kin:TOMR}). This is a linear model involving an invertible matrix, and thus the inverse kinematic model for the three-wheeled omnidirectional MR is:
\begin{equation}
\label{INkin:TOMR}
\left[
\begin{array}{c}
\omega_1\\
\omega_2\\
\omega_3
\end{array}
\right]
=
\frac{1}{r}
\left[
\begin{array}{ccc}
0&1&L\\
-\sin\left(\frac{\pi}{3}\right) & -\cos\left(\frac{\pi}{3}\right)&L\\
\sin\left(\frac{\pi}{3}\right) & -\cos\left(\frac{\pi}{3}\right)&L
\end{array}
\right]
\left[
\begin{array}{c}
\dot{x}\\
\dot{y}\\
\dot{\phi}
\end{array}
\right].
\end{equation}
Similarly, the inverse kinematic model for the \ac{DDR} can be derived from inverting (\ref{kin:diff}):
\begin{equation}
\label{INkin:DDR}
\left[
\begin{array}{c}
\omega_R\\
\omega_L
\end{array}
\right]
=
\frac{b}{r}
\left[
\begin{array}{cc}
\frac{1}{b}&1\\
\frac{1}{b}&-1
\end{array}
\right]
\left[
\begin{array}{c}
v\\
\dot{\theta}
\end{array}
\right].
\end{equation}
Finally, some MRs have complex forward kinematic models that are difficult to invert analytically due to nonlinearities and singularities. In such cases, machine learning is a good option for the derivation of the inverse kinematic model \cite{InverseKinematic1}.
\subsubsection{Dynamic Model}
\label{sec:MR:Dyn}
Dynamic models consider the forces exerted by the robot on its environment. They relate the WMR motion to the torque exerted by each motor and the electric signals producing it. WMR can be modelled as a nonlinear dynamic system whose general form is \cite{syst1}:
\begin{equation}
\label{dyn:1}
\dot{\mathbf{z}}=f(\mathbf{z},\mathbf{u}),
\end{equation}
where $\mathbf{z}$ is the {\it state vector} of the WMR\footnote{The state vector is a vector composed of a set of state variables. Broadly speaking, a set of state variables is the minimum set of variables required (in addition to the input) to uniquely determine the state vector. For more information on the subject, see \cite{syst1}.}, where $\mathbf{u}$ is the input signal vector and $f(\cdot)$ is a general function, which can be linear or non-linear. Let us begin the discussion of dynamic models with the popular {\it pure integrator model}:
\begin{equation}
\label{dyn:Integrator}
\frac{\mathrm{d}^n\mathbf{p}(t)}{\mathrm{d} t^n}=\mathbf{u}(t),\ \ \ n\in\mathbb{N}^+,
\end{equation}
where $\mathbf{u}$ is the control signal and $\mathbf{p}$ is the WMR position. (\ref{dyn:Integrator}) models a generic omnidirectional robot. It is also a very popular model due to its simplicity, which eases the theoretical analyses. This allows analytical results to be derived with less effort than with the use of more elaborate dynamic models (at the expense of accuracy, however). 
\par
Next, we discuss the order $n$ in the model (\ref{dyn:Integrator}). If $n=1$, we obtain the {\it single integrator model}, which allows for abrupt velocity changes. This model can describe with some accuracy the WMR motion under any of the following conditions: (i) the WMR speed is constant or changes slowly; (ii) the WMR dynamics is fast enough to follow the input. In \cite{CATP2}, the authors used this model to design trajectories, which were then tested experimentally on real WMRs.
\par
The model (\ref{dyn:Integrator}) with $n=2$, which is referred to as the {\it double integrator model}, limits the WMR acceleration. It is also a very common model in mobile robotics \cite{CATP3}, \cite{CATP7}. 
\par
The {\it pure integrator model} is a very simple general model used to describe a variety of WMRs, but greatly lacks accuracy. We will now begin to discuss more complex and specific dynamic models.
\par
Most WMRs use \ac{DC} motors to drive their wheels \cite{Rob5} as they are cheap, easy to control, and efficient. The WMR's dynamic model can be derived by first modeling the \ac{DC} motor and then by combining the motor models while considering the MR frame. Following such a procedure, a dynamic model for the \ac{DDR} is found to be \cite{Rob5}:
\begin{eqnarray}
\label{dyn:DDR:1}
\dot{\mathbf{z}}&=&\mathbf{A}\mathbf{z}+\mathbf{B}\mathbf{u},\\
\label{dyn:DDR:2}
\dot{\mathbf{p}}&=&\left[\begin{array}{ccc}
\cos(\phi)& \sin(\phi)& 0,\\
0& 0& 1,
\end{array}\right]^\mathrm{T}\mathbf{z},\\
\mathbf{z}&=&\mathbf{T}_q  \left[\begin{array}{c}\omega_R\\ \omega_L\end{array}\right],
\end{eqnarray}
\begin{equation}
\mathbf{T}_q =\frac{r}{2}\left[\begin{array}{cc}
1&1\\
b^{-1}&-b^{-1}
\end{array}\right],
\end{equation}
\begin{eqnarray}
\label{dyn:DDR:3}
\mathbf{p}=[\mathrm{x},\ \mathrm{y}, \phi]^\mathrm{T},\quad
\mathbf{z}=[v,\ \dot{\phi}]^\mathrm{T},\quad
\mathbf{u}=[u_R,\ u_L]^\mathrm{T},
\end{eqnarray}
where $u_R$ and $u_L$ are the normalized input \ac{DC} voltage to the motors (i.e. $|u_R|\leq $$=$$1$ and$|u_L|\leq $$=$$1$);  $\omega_R$ and $\omega_L$ are the angular speeds of the right and left wheels, respectively; $\mathbf{p}$ is the pose of the WMR (i.e. the WMR's horizontal position and orientation); $v$ and $\dot{\phi}$ are the translational speed and the angular speed (around its center) of the WMR, respectively. The matrices $\mathbf{A}$ and $\mathbf{B}$ contain information related to the inertia of the MR, its weight, the friction, the battery voltage level, and other electromechanical parameters of the motors.
\par
From this dynamic model, we observe that:
1) The \ac{DDR}'s state vector $\mathbf{z}$ is linear w.r.t. the control input $\mathbf{u}$, see (\ref{dyn:DDR:1}), but the pose is nonlinear w.r.t. $\mathbf{z}$, see (\ref{dyn:DDR:2}).
2) The \ac{DDR}'s velocity, encoded in $\mathbf{z}$, is related to the input $\mathbf{u}$ by a first-order linear system and its pose is related to $\mathbf{u}$ by a second-order nonlinear system. This implies that the input $\mathbf{u}$ controls the \ac{DDR}'s acceleration, and so the state $\mathbf{z}$ cannot change abruptly. Consequently, the WMR has the following limitations: (i) it cannot change the speed abruptly, thus requiring a non-zero deceleration time to stop; (ii) it cannot change direction abruptly.
3) If $\dot{\mathbf{z}}$ in (\ref{dyn:DDR:1}) is significantly small, then $\mathbf{z}\approx-\mathbf{A}^{-1}\mathbf{B}\mathbf{u}$. In other words, if the \ac{DDR}'s acceleration is significantly small, then the WMR state $\mathbf{z}$ almost becomes linear w.r.t. the control signal $\mathbf{u}$. As such, the dynamic \ac{DDR} motion model (\ref{dyn:DDR:1})-(\ref{dyn:DDR:3}) does not add additional constraints and can be sufficiently modelled by the kinematic model.
\par
The dynamic model of the \ac{TOMR} is \cite{Rob2}:
 \begin{eqnarray}
\label{dyn:TOMR:1}
\dot{\mathbf{z}}&=&\mathbf{A}(\phi,\dot{\phi})\mathbf{z}+\mathbf{B}\mathbf{u},\\
\label{dyn:TOMR:2}
\mathbf{p}&=&\left[\begin{array}{cc}
\mathbf{I}_{3\times3}& \mathbf{O}_{3\times3}
\end{array}\right]\mathbf{z},
\end{eqnarray}
\begin{eqnarray}
\label{dyn:TOMR:3}
\mathbf{z}=[\mathrm{x},\ \mathrm{y}, \phi,\ \dot{\mathrm{x}},\ \dot{\mathrm{y}}, \dot{\phi}]^\mathrm{T},\quad
\mathbf{u}=[u_1,\ u_2\ u_3]^\mathrm{T},
\end{eqnarray}
where:
\begin{equation}
\label{dyn:TOMR:4}
\mathbf{A}(\phi,\dot{\phi})=
\left[
\begin{array}{cc}
\mathbf{O}_{3\times 3} &\mathbf{I}_{3\times 3}\\
\mathbf{O}_{3\times 3} &\mathbf{R}(\phi)\dot{\mathbf{R}}^\mathrm{T}(\phi)\dot{\phi}-\mathbf{C}
\end{array}\right],
\end{equation}
\begin{equation}
\label{dyn:TOMR:5}
\mathbf{R}(\phi)=
\left[\begin{array}{ccc}
\cos(\phi)& -\sin(\phi)& 0,\\
\sin(\phi)& \cos(\phi)& 0,\\
0& 0& 1,
\end{array}\right].
\end{equation}
The matrices $\mathbf{B}$ and $\mathbf{C}$ in (\ref{dyn:TOMR:1}) and (\ref{dyn:TOMR:4})  depend on the electromechanical characteristics of the particular \ac{TOMR}. The matrix $\mathbf{R}(\phi)\dot{\mathbf{R}}^\mathrm{T}(\phi)\dot{\phi}$ is related to the Coriolis force.
\par
As opposed to the \ac{DDR}, where the dynamic model described in (\ref{dyn:DDR:1}) is linear, the \ac{TOMR} dynamic model in (\ref{dyn:TOMR:1})-(\ref{dyn:TOMR:5}) is nonlinear w.r.t. the input. The nonlinearity comes from $\mathbf{A}(\phi,\dot{\phi})$ and, in particular, from the component $\mathbf{R}(\phi)\dot{\mathbf{R}}^\mathrm{T}(\phi)\dot{\phi}$ in (\ref{dyn:TOMR:3}) produced by the Coriolis force. After having discussed some dynamic models for WMRs, we continue with the models for describing their energy consumption due to motion.
\subsubsection{Energy Consumption model}
\label{Rob:ener}
Most MRs draw their energy from a battery. Due to the limited capacity of the battery, it is important to calculate the \ac{MR} energy consumption to determine their operation time. There are various approaches to derive energy consumption models, but we will discuss only some of the most common ones.
\par
{\bf Electric model}: most WMRs use \ac{DC} motors to actuate their wheels because of their advantages in comparison to other types of motors, as explained in \cite{Rob5}. In this approach, we first calculate the energy consumed by the $k$th motor as a function of the instantaneous electric power:
\begin{equation}
\label{eq:En:e1}
E_k=\int_t\mathfrak{p}_k(t)\mathrm{d}t,
\end{equation}
where $\mathfrak{p}_k(t)=\mathfrak{i}_k(t)\mathfrak{v}_k(t)$ is the instantaneous electric power consumed by the $k$th motor, $\mathfrak{i}_k(t)$ is the input current, and $\mathfrak{v}_k(t)$ is the \ac{DC} input voltage to the $k$th motor. Since this is a \ac{DC} motor, the input variable controlling it is the \ac{DC} input voltage (controlled using \ac{PWM}), which is given as:
\begin{equation}
\label{eq:En:e3}
\mathfrak{v}_k(t)=u_k(t)V_s,
\end{equation}
where $V_s$ is the amplitude of the \ac{PWM} signal and $u_k(t)\in[-1,1]$ is the normalized control signal. Circuit theory and electromechanical equations are then used to derive the equations relating the input current $\mathfrak{i}_k(t)$ to the angular speed $\omega_k$ of the wheel that is mechanically coupled with the motor's shaft. Thereafter, the dynamic model is used to relate the WMR state vector to the wheel's angular speed. Finally, these equations are combined to obtain an equation relating the WMR energy consumption to the input vector $\mathbf{u}$ and the WMR state vector. To illustrate this method, we briefly present the energy consumption models of a \ac{DDR} and of a TOMR. According to \cite{Rob7}, the \ac{DDR}'s energy consumption model is:
\begin{eqnarray}
\label{eq:En:e4}
E
&=&\int_t\left(k_1\|\mathbf{u}(t)\|^2-k_2\mathbf{z}^{\mathrm{T}}(t)\mathbf{T}_q^{-\mathrm{T}}\mathbf{u}(t)\right)\mathrm{d}t,
\end{eqnarray}
where the integrand corresponds to the electric power consumed and the parameters $k_1$ and $k_2$ depend on electromechanical properties of the motors. Variables $\mathbf{u}$, $\mathbf{z}$, and $\mathbf{T}_q$ have the same definition as in (\ref{dyn:DDR:1})-(\ref{dyn:DDR:3}). The energy consumption model for the three-wheeled omnidirectional \ac{MR} is given in \cite{Rob2} as:
\begin{eqnarray}
\label{eq:En:e5}
E
&=&\int_t\left(k_1\|\mathbf{u}(t)\|^2-k_2\dot{\mathbf{p}}^{\mathrm{T}}(t)\mathbf{R}(\phi)\mathbf{B}\mathbf{u}(t)\right)\mathrm{d}t,
\end{eqnarray}
where $\mathbf{u}(t)$, $\mathbf{p}(t)$, and $\mathbf{B}$ have the same definition as in (\ref{dyn:TOMR:1}), while $k_1$ and $k_2$ depend on the motor's electromechanical characteristics. In \cite{Mob6}, the authors present an energy consumption model for a car-like robot derived using the same method, however neglecting the energy required to steer.
\par
This model describes the electrical energy consumed at the input of the motors and takes into account the energy lost as heat within the motor. (\ref{eq:En:e4}) and (\ref{eq:En:e5}) are quadratic functions of the normalized control signal $\mathbf{u}(t)$.
\par
{\bf Physics approach}: another approach to describe the WMR energy consumption is to analyze the system from an external perspective and calculate the WMR's kinetic energy and the energy required to overcome the floor's friction. To illustrate this type of model, let us consider the energy consumption model of the \ac{DDR} presented in \cite{Rob6}:
\begin{equation}
\label{eq:En:k1}
E_{\rm motors}=E_{\rm kin}+E_{\rm res},
\end{equation}
where $E_{\rm kin}$ is the kinetic energy, $E_{\rm res}$ is the energy consumed by the motors to overcome the traction resistance presented by the friction of the floor, and $E_{\rm motors}$ is the energy consumed by the motors, which is {\it measurable} from an external point of view, i.e., without taking into account the internal losses. The kinetic energy is given by:
\begin{eqnarray}
\label{eq:En:k2}
E_{\rm kin}&=&\frac{1}{2}mv^2(t)+I\omega^2(t),\nonumber\\
&=&\int_t\{mv(t)a(t)+I\omega(t)\beta(t)\}\mathrm{d}t,
\end{eqnarray}
where $m$ is the total WMR's mass; $I$ is the total WMR's rotational inertia; $v$ and $\omega$ are the translational and rotational speeds, respectively; and $a(t)$ and $\beta(t)$ are the translational and rotational accelerations, respectively. The products $v(t)a(t)$ and $\omega(t)\beta(t)$ in (\ref{eq:En:k2}) can be negative when the WMR stops, meaning that the WMR's motors get some power back that {\bf could} be recovered by the WMR's electrical system. But, most WMR's do not have the electric systems to recover such power and thus, considering this limitation, (\ref{eq:En:k2}) becomes:
\begin{equation}
\label{eq:En:k3}
E_{\rm kin}=\int_t\{m\max(v(t)a(t),0)+I\max(\omega(t)\beta(t),0)\}\mathrm{d}t.
\end{equation}
The term describing the energy consumed to overcome friction in (\ref{eq:En:k1}) is:
\begin{equation}
E_{res}=2\mu mg\int_t\max(|v(t)|,b|\omega(t)|)\mathrm{d}t.
\end{equation}
The distance travelled by the \ac{DDR} is approximately $\int_t|v(t)|\mathrm{d}t$. The advantage of this model lies with its independence to the type of motor used by the WMR. It requires only the mass $m$ and the rotational inertia $I$ of the WMR, which can be estimated. The disadvantage is that it only considers the energy expenditure observed from the outside and does not take into account the internal losses of the motors and the circuitry. 
\par
{\bf Data-driven Model}: the energy consumption models mentioned above are derived from physical principles. Those theoretical models, for the sake of tractability, often ignore certain effects, such as nonlinearities of the motor or the influence of temperature on the electric resistances. Another approach consists of measuring the electric consumption of the WMR under different conditions, creating a dataset, and then training numerical models. The authors of \cite{Mob5} present a simple implementation of this approach by measuring the power consumed by \ac{DC} motors at different angular speeds. The authors note that a suitable model for the power consumption of the $k$th motor is:
\begin{equation}
\label{eq:En:B1}
P_k=\sum_{j=0}^6a_j\omega_k^j,
\end{equation}
where $\omega_k$ is the angular speed of the $k$th motor and $\{a_{j}\}_{j=0}^6$ are experimentally determined coefficients. The model in (\ref{eq:En:B1}) is then combined with the kinetic model, which relates the velocity of the WMR to the angular speed of the wheels $\omega_k$ in order to obtain the power consumed by the WMR. After integration over time, we obtain the consumed energy.
\par
In \cite{Mob7}, there are two examples of this type of energy consumption model. First, the authors measure the power consumption of the PPRK (a commercial \ac{TOMR} moving at a constant linear speed $v$. The measurements are done at different linear speeds up to a maximum value. After a numerical analysis of the measurements, the authors observed that the power consumed by the PPRK can be well-modelled by a fourth order polynomial function of the linear speed.
Next, the authors did the same for the Pioneer 3DX (a commercial \ac{DDR}) and measured its energy consumption while moving at a constant speed on a straight line. Due to limitations in the robot's internal system, the Pioneer 3DX could not be operated at its maximum linear speed. The experimental results showed that the power consumption of this robot can be modelled with an affine function of the linear speed.
\par
The derivation of this type of model can require significant time to gather data in the laboratory. Additionally, the generalisation of the derived models to conditions different from those of the modeling phase might exhibit an uncertain accuracy, e.g. when trying to use the model in \cite{Mob7} to predict the energy consumption of the Pioneer 3DX while moving along curves or at variable linear speed. On the other hand, one advantage of this type of model is that they do not require deep theoretical knowledge. Also, they can implicitly consider the various complex processes involved in energy consumption, which may be complicated to take into account in the theoretical models. For example in \cite{Mob5}, the authors mention that \ac{DC} motors can be modelled with a second-order polynomial of the angular speed $\omega$ from electromagnetic theory. Yet, measurements show that such a model is insufficient. For example, the sixth-order polynomial model introduced in (\ref{eq:En:B1}) exhibited a much better fit on real measurements. This may be due to the fact that the second-order model derived from theory overlooks certain processes, while the model in (\ref{eq:En:B1}) took them into account implicitly through the fitting process. 
\par
{\bf Restricted Domain Models}: some simple energy consumption models are derived from more complex models by constraining their range of validity. To illustrate this, let us consider the work in \cite{Mob8} where the authors use an electric energy consumption model for a \ac{DDR}, similar to the one used in (\ref{eq:En:e4}). They restrict the \ac{DDR} to move within a straight line with a trapezoidal linear speed profile. For such a trajectory, they analytically calculate the energy consumption starting from the electric energy consumption model to obtain the following expression:
\begin{equation}
\label{eq:En:H1}
E=\sum_{j=-1}^3a_j\omega^j,
\end{equation}
where $\omega$ is the angular speed of the \ac{DDR} wheels. The model (\ref{eq:En:H1}) exposes an interesting phenomenon that is sometimes overlooked: motors can be inefficient when operated at very low speeds. As the motor's speed decreases, the term $\omega^{-1}$ in (\ref{eq:En:H1}) dominates the energy consumption and grows very quickly. After the optimization of the trapezoidal speed profile, the authors found that the energy consumption is proportional to the distance travelled when the total duration of movement remains constant. Based on the result in (\ref{eq:En:H1}), the authors of \cite{CATP4} modeled the energy consumption as being proportional to the distance $D$ travelled by the WMR:
\begin{equation}
\label{eq:En:H2}
E=kD,
\end{equation}
where $k$ is a proportionality factor. The model (\ref{eq:En:H2}) has been adopted by many authors for simplicity and is sometimes derived in a similar manner in different papers, or sometimes just assumed as it is {\it intuitive} and simple, see  \cite{Mob4} and \cite{CATP1}. 

These models are, in general, simplifications of other more complex models under certain conditions. When choosing a model, we must consider where it comes from and how it was derived in order to see if the application in which we intend to use it is close enough to the conditions under which the model was derived. This is done so as to ensure that we do not operate the model outside its range of validity. Otherwise, if we neglect this aspect, then the model selected might deviate significantly from the true energy consumption. 
For example, while the model introduced in (\ref{eq:En:H2}) is appropriate when a WMR is moving at a constant speed, it may be quite inaccurate when the speed varies considerably over time.

{\bf Square norm:} lastly, another popular energy consumption model within the control theory community is :
\begin{equation}
\label{eq:En:N1}
E=\int_{0}^t\|\mathbf{u}(\tau)\mathrm{d}\|^2\tau,
\end{equation}
where $\mathbf{u}$ is the control signal of the WMR dynamic model. This is notably a simplification of the energy consumption models derived through circuit theory presented in (\ref{eq:En:e4}) and (\ref{eq:En:e5}). The model in (\ref{eq:En:N1}) becomes closer to the models (\ref{eq:En:e4}) and (\ref{eq:En:e5}) when the WMR operates at low speeds and/or the coefficient $k_2$ of those models is small. \qedsymbol{}

Although motion consumes a significant part of the energy in WMRs, there are also other processes which consume a non-negligible amount of energy. In \cite{Mob2}, the authors empirically evaluated the contribution of different processes to the energy consumption of a DDR and found that the sensors and microcontrollers also contribute significantly to energy consumption
The microcontroller's power consumption can be modelled as constants, since they usually perform low level tasks which are repetitive and relatively stable. On the other hand, the power consumption of the embedded computers can be modelled as a stochastic process as they usually perform tasks which depend more on exogenous inputs, and thus have a more variable behaviour.
\subsection{Rotary-wing UAVs}
\label{sec:uav}
The study of the motion of aerial vehicles is a complex subject that has been investigated since the early appearance of the first airplanes. There is a large body of literature on the aerodynamic aspects of these vehicles and their modeling. In this section, we will discuss the quadrotor aerial robot, which is a basic type of multirotor UAV.
\par
Multirotor aerial robots (also called rotary-wing aerial robots) are one of the most popular types of aerial robots nowadays. One of the most common type of these UAVs is the quadrotor, which is the subject of this section. 
\par
The WMRs discussed in the previous section operate in 2D and their configuration is fully described by their horizontal position and orientation. In contrast, UAVs operate in 3D. To fully describe their configuration, we need to specify their 3D position, as well as their attitude\footnote{Attitude, not to be confused with the altitude, is synonym with orientation.}, which can be expressed using either Euler angles or quaternions \cite{quat1,quat2}. In this tutorial, we will limit the discussion to models representing the attitude using Euler angles; details about the utilisation of quaternions to represent the attitude can be found in \cite{Diebel06representingattitude:}. 
\par
The configuration of the UAV can be described using Euler angles as:
\begin{equation}
\label{eq:UAV:1}
\mathbf{p}(t)=[x,\ y\ ,z,\ \theta,\ \phi,\ \psi]^\mathrm{T},
\end{equation}
where $[x,\ y\ ,z]^\mathrm{T}$ is the UAV center of mass and $\theta$, $\phi$, and $\psi$ are the Euler angles of roll, pitch, and yaw, respectively, which describe the orientation of the UAV (see \cite{UAV7} for a more detailed geometrical description of these angles). Note that $[x,\ y\ ,z]^\mathrm{T}$ is represented in a static coordinate frame attached to the world and not to the UAV itself, and $[\theta, \phi, \psi]^\mathrm{T}$ are represented in a coordinate frame attached to the center of mass of the UAV oriented in the same manner as the inertial coordinate frame. Before discussing the quadrotor dynamic model, we will briefly discuss the physical principles that allow it to fly. In a classic quadrotor, the four propellers lie in the same plane and are oriented vertically. When the propeller rotates, it creates thrust in the same direction as its orientation and with a sense opposing the gravity. The faster the propeller turns, the greater the thrust generated. When all four propellers produce the same thrust, the plane containing all four propellers is parallel to the floor, and the resulting thrust is vertical. If this thrust equals the gravitational force, then the quadrotor remains in the air hovering. When one propeller turns faster or slower than the other, the quadrotor tilts and the resulting thrust presents a horizontal component, making the UAV move in the horizontal plane. 
\par
Just as the dynamic models of the WMRs were derived starting from the electrical analysis of their motors, the derivation of the dynamic model for multirotors begins in the same way. Consider a dynamic model for the quadrotor that is used by many roboticists \cite{UAV9}:
\begin{eqnarray}
\label{eq:QR_model}
\left[\begin{array}{c}
\ders{\xx}{t}{2}
\\ \noalign{\smallskip}
\ders{\yy}{t}{2}
\\ \noalign{\smallskip}
\ders{\zz}{t}{2}
\end{array}\right] &=&
\left[\begin{array}{c}
\mathrm{c}(\phi)\mathrm{s}(\theta)\mathrm{c}(\psi) +
\mathrm{s}(\phi)\mathrm{s}(\psi)
\\ \noalign{\smallskip}
\mathrm{c}(\phi)\mathrm{s}(\theta)\mathrm{s}(\psi) -
\mathrm{s}(\phi)\mathrm{c}(\psi)
\\ \noalign{\smallskip}
\mathrm{c}(\phi)\mathrm{c}(\theta)
\end{array}\right]\frac{{u}_{\zz}}{m} -
\left[\begin{array}{c}
0 \\ 0 \\ g
\end{array}\right],
\nonumber\\
\end{eqnarray}
\begin{eqnarray}
\label{eq:QR_model2}
\left[\begin{array}{c}
\ders{\phi}{t}{2}
\\ \noalign{\smallskip}
\ders{\theta}{t}{2}
\\ \noalign{\smallskip}
\ders{\psi}{t}{2}
\end{array}\right] &=&
\left[\begin{array}{c}
\left(\frac{I_\yy-I_\zz}{I_\xx}\right)\,\der{\theta}{t}\,\der{\psi}{t} -
\frac{J}{I_\xx}\,\der{\theta}{t}\,{q}_{w}
\\ \noalign{\smallskip}
\left(\frac{I_\zz-I_\xx}{I_\yy}\right)\,\der{\phi}{t}\,\der{\psi}{t}+
\frac{J}{I_\yy}\,\der{\phi}{t}\,{q}_{w}
\\ \noalign{\smallskip}
\left(\frac{I_\xx-I_\yy}{I_\zz}\right)\,\der{\phi}{t}\,\der{\theta}{t}
\end{array}\right] +
\left[\begin{array}{c}
\frac{\ell\,{u}_{\yy}}{I_\xx}
\\ \noalign{\smallskip}
\frac{\ell\,{u}_{\xx}}{I_\yy}
\\ \noalign{\smallskip}
\frac{\,{u}_{\psi}}{I_\zz}
\end{array}\right],
\nonumber\\
\end{eqnarray}
\begin{eqnarray}
\label{eq:Inp_vec}
\notag
\left[\begin{array}{c}
{u}_{\xx} \\ {u}_{\yy} \\ {u}_{\zz} \\ {u}_{\psi}
\end{array}\right] &=&
\left[\begin{array}{cccc}
-\kappa_b&0&\kappa_b&0\\
0&\kappa_b&0&-\kappa_b\\
\kappa_b&\kappa_b&\kappa_b&\kappa_b\\
\kappa_\tau&-\kappa_\tau&\kappa_\tau&-\kappa_\tau
\end{array}\right]
\left[\begin{array}{c}
{\omega}_{1}^{2} \\ {\omega}_{2}^{2} \\ {\omega}_{3}^{2} \\ {\omega}_{4}^{2}
\end{array}\right],
\\
\end{eqnarray}
\begin{eqnarray}
\label{eq:qw_vec}
{q}_{w}&=&\omega_1 - \omega_2 + \omega_3 - \omega_4,
\end{eqnarray}
\noindent
where ${u}_{\xx}(t)$, ${u}_{\yy}(t)$, ${u}_{\zz}(t)$, and ${u}_{\psi}(t)$ denote the control signals for the drone;  $\omega_j(t)$ is the angular velocity of the $j$th motor; $m$ is the total mass of the drone; $g$ is the gravitational constant; $\ell$ is the distance from the center of the quadrotor to each motor;  $I_\xx$, $I_\yy$, and $I_\zz$ are the rotational inertia components; $J$ is the total inertia of the motors; and $\kappa_b$ and $\kappa_\tau$ are the thrust and aerodynamic drag factors of the propellers, respectively. In (\ref{eq:Inp_vec}), the matrix relating the vector inputs 
$\left[\begin{array}{cccc}
{u}_{\zz} & {u}_{\yy} & {u}_{\xx} & {u}_{\psi}
\end{array}\right]^{T}
$ with the square angular velocities vector
$\left[\begin{array}{cccc}
{\omega}_{1}^{2}(t) & {\omega}_{2}^{2}(t) & {\omega}_{3}^{2}(t) & {\omega}_{4}^{2}(t) \end{array}\right]^{\mathrm{T}}$ is not singular.  
\par
Equation (\ref{eq:QR_model2}) describes the drone's Euler angles (roll $(\phi)$, pitch $(\theta)$, and yaw $(\psi)$) measured with respect to the axes $o_{B}{\xx}_{B}$, $o_{B}{\yy}_{B}$, and $o_{B}{\zz}_{B}$, with $(o_{B}{\xx}_{B}{\yy}_{B}{\zz}_{B})$ being the body axis system whose origin ${o_{B}}$ is given by the geometric centre of the quadrotor. 
\par
The quadrotor motion given by the model in (\ref{eq:QR_model})-(\ref{eq:qw_vec}) is described w.r.t. to a fixed orthogonal axis set ${(o{\xx}{\yy}{\zz})}$, where ${o{\zz}}$ points vertically up, i.e. opposed to the gravity vector. 
The origin ${o}$ is located at a desired height ${\bar{\zz}}$ with respect to the ground level. The coordinates  $\xx$, $\yy$, and $\zz$ in (\ref{eq:QR_model}) refer to the position of the centre of gravity of the quadrotor in the space where $\zz$ is its altitude \cite{Cook}. In the literature, we find different axis configurations in which the quadrotor motion can be described and each axis configuration provides slightly different models. The time dependence of the variables in equations (\ref{eq:QR_model}) and (\ref{eq:QR_model2}) is not explicitly shown in order to lighten the notation. Further, due to the symmetry of the quadrotor, we have $I_\xx=I_\yy=I$.
\par
More complex models are also possible, which consider external disturbances, such as the wind. Regarding the energy consumption of the quadrotor, there are electric models and physics-based models that are derived using approaches similar to those used for the WMR. In \cite{UAV9}, the energy consumption of a quadrotor is modeled as:
\begin{eqnarray}
\label{eq:UAV:en:1}
E&=&\sum_{j=1}^4\int_t\Bigg[\left(\sum_{k=0}^4c_k\omega_j^k(t)\right)+c_6\dot{\omega}_j(t)+c_7\dot{\omega}_j^2(t)\nonumber\\
&+&c_8\omega_j(t)\dot{\omega}_j(t)+c_9\omega_j^2(t)\dot{\omega}(t)\Bigg]\mathrm{d}t,
\end{eqnarray}
where the coefficients $\{c_k\}_{0}^9$ depend on the parameters of the quadrotor's motors and the geometry of the propellers. 
\par
In \cite{UAV10}, the authors present the following hybrid energy consumption model  based on basic mechanics and completed with some correction factors obtained experimentally:
\begin{eqnarray}
\label{eq:UAV:en:2}
E_c&=&\int_{t_0}^{t_f}\sum_{j=1}^4\tau_j(t)\omega_j(t)\mathrm{d}t,\\
&=&\int_{t_0}^{t_f}\sum_{j=1}^4\left(\dot{\omega}_j(t)+\mathcal{K}_\tau\omega_j^2(t)+D_v\omega_j(t)\right)\omega_j(t)\mathrm{d}t\nonumber,
\end{eqnarray}
where (\ref{eq:UAV:en:2}) is the energy consumed by the four motors of the quadrotor. In other words, (\ref{eq:UAV:en:2}) only describes the amount of energy that is translated into mechanical energy, but disregards the efficiency of the motors. 
\subsection{Fixed-wing UAVs}
\label{sec:fixedwing}
In this section, we briefly discuss fixed-wing UAVs and present a simple, but useful dynamic model for \ac{CaTP} problems. Fixed-wing \acp{UAV} fly using principles that are different from those used by multirotor \acp{UAV} and are consequently modelled in a different manner with different characteristics.
\par
In general, fixed-wing \acp{UAV} are more energy-efficient than multirotor \acp{UAV} mainly because of their ability to glide. They also fly for longer times, longer distances, and at higher speeds. However, they are less agile and cannot land or take-off vertically. As opposed to multirotor UAVs, fixed-wing UAVs are generally not designed to hover. Nevertheless, some special types of fixed-wing UAVs having high thrust-to-weight ratios can hover using complex control techniques \cite{GreenICRA2006}. 
\par
Fixed-wings \ac{UAV} are controlled by the thrust generated by its propeller(s) and by controlling surfaces (aileron, elevator, and rudder). Micro fixed-wing \acp{UAV} usually drive their propellers with an electric motor, while small fixed-wing \acp{UAV} can drive it using gas powered motors.
\par
The fixed-wing \ac{UAV} center of mass position is sometimes described in the North-East-Down (NED) coordinate system, where the down axis points towards the center of the Earth and is aligned with the force of gravity. In this case, the altitude is measured in the opposite direction of the down axis. The attitude of the fixed-wing airplane is expressed using Euler angles in the body frame defined as follows: its origin lies in the center of gravity of the airplane, the $x$ axis points to the nose of the plane, the $y$ axis points to the right wing, and the $z$ axis is orthogonal to those two axes and follows the right-hand rule. The roll $\phi$ describes the rotation about the $x$ axis, the pitch $\theta$ describes the rotation about the $y$ axis, and the yaw $\psi$ describes the rotation about the $z$ axis.

The aerodynamics of airplanes are significantly more complex and nonlinear. Since this is an elementary tutorial, we will present only a high-level simplified dynamic model that can be used for trajectory planing, as well as one linearized dynamic model. In the absence of wind, a simplified nonlinear dynamic model describing the fixed-wing UAV motion is \cite{fixed-wingBook1}:
\begin{equation}
\label{eq:fixed-wing:1}
\left[
\begin{array}{cc}
    \dot{p}_n \\
    \dot{p}_e \\
    \dot{h}
\end{array}
\right]=V_a
\left[
\begin{array}{cc}
    \cos(\psi)\cos(\gamma) \\
    \sin(\psi)\cos(\gamma) \\
    \sin(\gamma)
\end{array}
\right]
\end{equation}
\begin{equation}
\label{eq:fixed-wing:1b}
\left[
\begin{array}{cc}
    \dot{\psi} \\
    \dot{\gamma} 
\end{array}
\right]=
\left[
\begin{array}{cc}
    \frac{F_{lift}}{mV_a}\frac{\sin(\phi)}{\cos(\gamma)} \\
    \frac{F_{lift}}{mV_a}\cos(\phi)-\frac{g}{V_a}\cos(\gamma) 
\end{array}
\right]
\end{equation}
\begin{equation}
\label{eq:fixed-wing:1c}
\left[
\begin{array}{cc}
    F_{lift} \\
    F_{drag}
\end{array}
\right]=
\frac{1}{2}\rho V_a^2 S 
\left[
\begin{array}{cc}
    C_L\\
    C_{D_0}+KC_L^2
\end{array}
\right]
\end{equation}
\begin{equation}
\label{eq:fixed-wing:1d}
    \dot{V}_a=\frac{F_{\rm thrust}}{m}-\frac{F_{\rm drag}}{m}-g\sin(\gamma)
\end{equation}
where $g$ is the gravitational constant, $S$ is the planform area of the wing, $C_{D_0}$ is the zero lift drag, $K$ is the induced drag factor, $m$ is the mass of the airplane, and $\rho$ is the air density. The input to this model is the lift coefficient\footnote{Note that the physical fixed-wing UAV is controlled via its thrust and the three control surfaces (aileron, elevator, and rudder). Hence, even if $C_L$ constitutes the input to the model, in practice it is not directly controlled.} $C_L$, the thrust $F_{\rm thrust}$, and the roll $\phi$. $\dot{p}_n$ and $\dot{p}_e$ are the speeds along the north and east axes, respectively, and $\dot{h}$ is the altitude speed measured w.r.t. to the negative direction of the down axis. $F_{lift}$ and $F_{\rm drag}$ are the lift and drag forces experienced by the plane. Finally, $V_a$ is the air speed of the airplane.

The constant altitude and the airspeed scenario (i.e. $\dot{h}=0$ and $\dot{V}_a=0$) are common and of particular importance for \ac{CaTP} applications.  In this case, we have $\gamma=0$, $\dot{\gamma}=0$, and $F_{thrust}=F_{drag}$. The dynamic model (\ref{eq:fixed-wing:1})-(\ref{eq:fixed-wing:1d}) reduces to the kinematic model:
\begin{equation}
\label{eq:fixed-wing:2}
\left[
\begin{array}{cc}
    \dot{p}_n \\
    \dot{p}_e 
\end{array}
\right]=V_a
\left[
\begin{array}{cc}
    \cos(\psi) \\
    \sin(\psi)
\end{array}
\right]
\end{equation}
\begin{equation}
  \dot{\psi}= ({g}/{V_a})\tan(\phi).
\end{equation}
With this kinematic model, the paths are usually composed of straight lines and circular arcs \cite{Lugo-CardenasICUAS2014}. The simplified models (\ref{eq:fixed-wing:1}) and (\ref{eq:fixed-wing:2}) are practical, but they do not describe the pitch angle $\theta$. This can be an issue for some \ac{CaTP} since the variations in the orientation of the antenna mounted on the fixed-wing \ac{UAV} cannot be fully determined without the pitch $\theta$.

Another simple type of dynamic model for the fixed-wing \ac{UAV} are the linear models. These models are derived after linearizing more complex nonlinear aerodynamic models around small attitude variations. They are separated into two decoupled models, the longitudinal motion model and the lateral motion model. The longitudinal motion model is:
\begin{eqnarray}
\left[
\begin{array}{c}
    \dot{u}\\
    \dot{w}\\
    \dot{q}\\
    \dot{\theta}\\
    \dot{h}
\end{array}
\right]=
\mathbf{A}
\left[
\begin{array}{c}
    u\\
    w\\
    q\\
    \theta\\
    h
\end{array}
\right]+\mathbf{B}
\left[
\begin{array}{c}
    \delta_e\\
    \tau
\end{array}
\right]
\end{eqnarray}
where $\tau$ is the thrust and $\delta_e$ is the elevator angle. For the lateral model, we have that:
\begin{eqnarray}
\left[
\begin{array}{c}
    \dot{v}\\
    \dot{p}\\
    \dot{r}\\
    \dot{\phi}\\
    \dot{\psi}
\end{array}
\right]=
\mathbf{C}
\left[
\begin{array}{c}
    v\\
    p\\
    r\\
    \phi\\
    \psi
\end{array}
\right]+\mathbf{D}
\left[
\begin{array}{c}
    \delta_a\\
    \delta_r
\end{array}
\right]
\end{eqnarray}
where $\delta_a$ and $\delta_r$ are the action of the aileron and the rudder, respectively. The matrices $\mathbf{A}$, $\mathbf{B}$, $\mathbf{C}$, and $\mathbf{D}$ depend on the particular airplane. Since this is an elementary tutorial, we did not consider the effect of wind on the fixed-wing \ac{UAV} motion models, but the interested reader can look into \cite{LangelaanJGCD2011,BeardIEEETransactionsCST2014,JohansenICUAS2015,BorupIEEETransactionsCST2020} for more information regarding this subject.
\par
It can be demonstrated from physical principles that the energy consumption for the fixed-wing \ac{UAV} can be expressed as follows \cite{uav3}:
\begin{eqnarray}
\label{eq:energy:fixed-wing}
    E&=&\int_0^T\left(c_1\|\mathbf{v}\|^3+\frac{c_2}{\|\mathbf{v}\|}\left(1+\frac{\|\mathbf{a}\|^2-\frac{\mathbf{a}^T\mathbf{v}}{\|\mathbf{v}\|^2}}{g^2}\right)\right)\mathrm{d}t\nonumber\\
    &+&\frac{m}{2}(\|\mathbf{v}\|^2(T)-\|\mathbf{v}\|^2(0))
\end{eqnarray}
where $\mathbf{v}$ is the velocity and $\mathbf{a}$ is the acceleration of the \ac{UAV}. 
\subsection{Final Comments on Models for Robots}
\label{sec:FCMMR}
We have presented an overview of some relevant MR's models that are useful for \ac{CaTP} problems, but have omitted some important observations. The models presented in this section are all in continuous time, but it is possible to transform them into discrete time models by transforming differential equations into difference equations. Continuous-time models allow for the utilisation of many analytical tools based on derivatives, such as calculus of variations. On the other hand, discrete-time models allow for numerical techniques like dynamic programming or other related techniques. It is important to mention that, in practice, the control signals for the ground \acp{MR} and  the UAVs are executed in digital computers on board, and thus the control system is implemented in discrete time.
\par
Finally, all the kinematic and dynamic models described in this section are analytical and are mostly derived from physics. But, there are other types of motion models derived through experimentation and machine learning techniques \cite{ML1}.
\section{Communications System}
\label{sec:Mod:comms}
This section mainly addresses researchers who desire to work on \ac{CaTP}, but lack the background in communications systems. Those familiar with the topic can skip this section and directly proceed to section \ref{sec:CATP}.
In this section, we introduce the reader to the basic concepts of communication systems and wireless channel models required to study \ac{CaTP} problems. We begin by introducing a common communication system: {\bf transceiver}. This is a device composed of a transmitter and a receiver. When a communications link is established between two entities, it can take three different forms:

{\bf Simplex Link}: one entity operates exclusively as a transmitter and the other operates exclusively as a receiver. The data flow is always unidirectional.

{\bf Full-Duplex Link}: both entities receive and transmit simultaneously. Two independent data flows in opposite directions occur simultaneously.

{\bf Half-Duplex Link}: both entities receive and transmit in turns, where two independent data flow in opposite directions, but only one is active at a time. This can be implemented using {\bf time duplexing}, where during an interval of time, one entity transmits and the other receives. During the next interval of time, the roles are swapped. This is periodically repeated. \qedsymbol{} 
\par
We now provide a brief overview of the digital transmission. The {\bf source node} generates data as blocks of bits or as a continuous stream of bits. This data is then divided in small groups, which are inserted into packets to form the payload\footnote{The payload's size can be constant or variable depending on the communications protocol selected.}. Each packet has a {\bf header} that contains information such as the destination and/or checkup bits to evaluate the integrity\footnote{In other words checking if the packet has been received with errors.} of the packet at the receiver. There are two main strategies to exploit the checkup bits in the packet's header:

{\bf Retransmission}:
if the receiver does not detect any error in the received data packet, it transmits a confirmation packet back to the source node (containing no payload), indicating that the data packet was correctly received. Thereafter, the source node can transmit new data packets (containing new payload). However, if the receiver detects an error in the received data packet, it does not send back the confirmation packet. When the source node realizes that it did not receive a confirmation packet, it assumes that an error occurred and re-transmits the same data packet.
The number of retransmissions will depend on the particular communications protocol.

{\bf No-retransmission:} if the receiver detects an error in the data packet, it discards the payload. On the other hand, if the packet is correctly received, the receiver does not transmit any confirmation to the source node and simply waits for the next packet. The source node continues to transmit data packets. \qedsymbol{}

The retransmission strategy provides robustness to the transmission of data at the cost of a lowered bit rate and increased latency due to the time spent in confirmation and retransmission of data packets. The no-retransmission strategy can achieve a higher bit rate and lower latency at the cost of more erroneous or missing data. The selection of the transmission strategy will depend on the particular application requirements.
\par
After forming the packets, the transmitter modulates the sequence of bits with a carrier signal of high frequency $f_c$ (or equivalently of short wavelength $\lambda\triangleq c/f_c$ where $c$ is the speed of light) suitable to be radiated as an electromagnetic waves, which are then radiated through the antenna. The propagation environment modifies the radiated wave before arriving to the receiver's antenna, where it is converted back to an electric signal and processed. The wireless channel model describes the effect that the propagation environment has on the transmitted signal until it reaches the receiver.
\subsection{Wireless channel modelling}
\label{Comm:channel}
In the context of \ac{CaTP}, oversimplified wireless channel models can lead the designer to overestimate the communications channel quality and to overlook certain channel behaviours that the \ac{MR} will encounter in real conditions. In such cases, the \ac{MR} might underperform or fail to complete its task due to unexpectedly poor communication quality.
\par
To provide the researcher with the {\em basic} knowledge of communications for \ac{CaTP} problems, we will limit the discussion to the simplest type of communications systems: narrowband\footnote{This means that the bandwidth of the modulated signal is significantly smaller than the carrier frequency $f_c$.}, single-antenna, and single carrier communications systems. In addition, we will not address the issue of interference. The general channel model representing such systems is:
\begin{equation}
\label{chann:1}
y(t)=\mathcal{H}(\mathbf{p}(t),\mathbf{q}(t),t)x(t)+n(t),
\end{equation}
where $x(t)$ and $y(t)$ are the continuous-time transmitted and received complex signals, respectively; $n(t)$ is the noise generated at the receiver, which is usually modelled as a random complex circular white Gaussian process; $\mathbf{p}(t)$ and $\mathbf{q}(t)$ are the positions of the transmitter and the receiver, respectively, at time $t$; and $\mathcal{H}(\mathbf{p}(t),\mathbf{q}(t),t)$ is the complex channel gain. \ac{CaTP} problems require models that describe the spatio-temporal channel gain $\mathcal{H}(\mathbf{p}(t),\mathbf{q}(t),t)$. We further discuss different common models to describe the spatial variations of the channel gain. To model $\mathcal{H}(\mathbf{p}(t),\mathbf{q}(t),t)$, deterministic, stochastic, and machine learning approaches can be used.
\subsubsection{Deterministic Models}
\label{Comm:channel:deterministic}
In the deterministic approach \cite{bookComm1}, the wireless channel models are often derived from physical principles.
One of the simplest and yet most important channel models is the {\it power loss model}. It is the base for many more sophisticated channel models and describes how the mean received signal power varies with the transmitter-receiver distance. It is modelled as \cite{PL1}:
\begin{equation}
\label{chann:1bb}
\mathcal{H}(\mathbf{p}(t),\mathbf{q}(t),t)=L_P^{-1}(\mathbf{p}(t),\mathbf{q}(t)),
\end{equation}
\begin{equation}
\label{eq:PL:1}
L_P(\mathbf{p}(t),\mathbf{q}(t))=K_0\left({\|\mathbf{q}(t)-\mathbf{p}(t)\|_2}/{d_0}\right)^{\alpha/2},
\end{equation}
where $\alpha$ is the power path loss coefficient, $d_0$ is a reference distance, and $K_0$ is the path loss observed at distance $d_0$ that must be in the far field region \cite{MPC1}. This model is valid for distances larger than $d_0$, and is used at distances larger than the radiated signal's wavelength and the physical antenna's dimensions \cite{bookComm1}. Under {\it free space} conditions, the power path loss coefficient becomes $\alpha=2$, but experimental results have reported path loss coefficients as low as $\alpha=1.6$ \cite{comm11,comm3} in some urban environments, buildings, and underground mines. This occurs because some hallways and tunnels behave as {\it giant waveguides} due to their geometry and the presence of metallic objects that act as reflectors. 
\par
Experiments have shown that the power path loss coefficient can change after certain distances. This is modelled using break points \cite{comm11} after which the path loss changes. 
\par
Another deterministic extension to the model in (\ref{eq:PL:1}) is the two-ray model \cite{PL1}, which takes into consideration only the \acf{LoS} wave and the wave that reaches the receiver's antenna after only one reflection on the floor. This model was originally derived in the context of cellular network communications \cite{bookComm5}. It has been tested for scenarios with a transmitter of less than 50m altitude; this model makes the assumption that the distance between the transmitter and the receiver is such that the curvature of the earth can be neglected, and thus the floor is considered flat.
\par
If we consider multiple waves beside the \ac{LoS} and the reflected wave of the two-ray channel model, then we obtain the ray tracing model \cite{RayTracing1}, \cite{RayTracing2}, \cite{RayTracing3}. This method determines the interaction of multiple radiated waves with the environment (e.g. buildings, floor, walls) before arriving to the receiver's antenna and requires a computational map of the area in which both the transmitter and the receiver operate. The accuracy of the ray tracing model increases when the map is more detailed and when more electromagnetic interactions are considered (e.g. reflection, refraction, diffraction). However, this also increases the computational load. 
\par
\par
Let us discuss another deterministic channel model which has been specially devised for indoor operations, and which also requires a map of the building as well as the positions of both transceivers. This model draws a straight line between both nodes, counts the number of walls and floors crossed \cite{Sh3}, and then represents the losses due to those walls and floors as described in \cite{Sh4}. 
\par
\subsubsection{Stochastic Models}
\label{Comm:channel:stochastic}
Stochastic models describe the channel in terms of its mean, variance, and correlation functions, rather than trying to predict the exact channel value, unlike deterministic models. Stochastic models are usually simple and can describe the average behaviour of the channel models accurately as well as its statistics. Their simplicity allows for mathematical analysis that can provide useful insights in many application domains, including \ac{CaTP} problems. 
Using this approach, $\mathcal{H}(\mathbf{p}(t),\mathbf{q}(t),t)$ is modelled as a multi-scale spatial and time-varying stochastic process \cite{comm1} composed of three terms \cite{comm1,comm2}:
\begin{eqnarray}
\label{chann:2}
\mathcal{H}(\mathbf{p}(t),\mathbf{q}(t),t)&=&\frac{s(\mathbf{p}(t),\mathbf{q}(t))h(\mathbf{p}(t),\mathbf{q}(t),(t))}{L_P(\mathbf{p}(t),\mathbf{q}(t))},
\end{eqnarray}
where $s(\mathbf{p}(t),\mathbf{q}(t))$ represents the shadowing \cite{Sh1,Sh2}, $h(\mathbf{p}(t),\mathbf{q}(t),t)$ represents the small-scale fading \cite{MPC1}, and $L_P(\mathbf{p}(t),\mathbf{q}(t))$ represents the path loss \cite{PL1}. We proceed to discuss the physical meaning of these components and their most relevant mathematical models from the perspective of \ac{CaTP} applications.
\par


{\bf Path-loss}: this describes a deterministic component that models the mean power loss variations due to distance between the transmitter and the receiver. It usually takes the form of (\ref{eq:PL:1}). There are also experimentally derived models for the path loss models, such as the Okumura-Hata model \cite{PL2}.
\par
{\bf Shadowing}: this is a random process that models the signal power reduction due to obstructions caused by large objects\footnote{Large w.r.t. the wavelength.}, such as buildings. It is a time-invariant term and depends on the transmitter and receiver positions. The shadowing has been experimentally characterized \cite{Sh1} and is generally modelled as a log-normal real random process with variance $\sigma$ and mean $\mu$. Its spatial autocorrelation has been found experimentally to follow an exponential function \cite{comm1}\footnote{Although this spatial correlation model fits many scenarios, it is by no means a universal model for the shadowing process, see \cite{BeaulieuIEEEWCL2019}}:
\begin{equation}
\label{eq:Sh:1}
r({d}(t))=\exp\left(-d(t)/\beta\right),
\end{equation} 
where $\beta$ is the decorrelation distance that is usually in the order of $10\lambda$, and is thus also called large-scale fading. For small distances (smaller than $\beta$), the shadowing is often considered constant. To give more flexibility to the shadowing model, $\sigma$ and mean $\mu$ can be made dependent on the geographical region \cite{comm2}. The shadowing effect can thus be simulated numerically using mathematical models \cite{CommSim1}. 
\par
{\bf Multipath-fading}: the electromagnetic wave radiated by the transmitter reaches the receiver's antenna after travelling through multiple random paths. These multiple components interact with the environment through reflection, diffraction, and refraction before arriving, with different phases, to the receiver's antenna where they are combined. As a consequence, the receiver observes constructive interference in some locations and destructive interference in others. This results in a random spatial process with large signal strength variations over small distances (smaller than $1\lambda$). This phenomenon is called small-scale fading or multi-path fading \cite{bookComm1,bookComm2}. 
\par
Small-scale fading $h(\mathbf{p}(t),\mathbf{q}(t),t)$ is modelled as a random spatial-temporal process with a certain distribution and spatio-temporal correlations dependent on the environment. The study of small-scale fading is a complex subject. In this elementary tutorial, we focus only on the basic models used in \ac{CaTP} problems.
\par
Three elements influence $h(\mathbf{p}(t),\mathbf{q}(t),t)$: the receiver's position $\mathbf{p}(t)$, the transmitter's position $\mathbf{q}(t)$, and the environment. Any change in any of these elements changes the experienced small-scale fading. In traditional mobile communications, $\mathbf{p}(t)$ and $\mathbf{q}(t)$ are considered uncontrollable { time-variant} random variables. In that context, $h(\mathbf{p}(t),\mathbf{q}(t),t)$ is considered time-variant if the environment varies or if the transceivers move. On the other hand, in \ac{CaTP}, $h(\mathbf{p}(t),\mathbf{q}(t),t)$ is considered time invariant if  $h(\mathbf{p},\mathbf{q},t_1)= h(\mathbf{p},\mathbf{q},t_2)$ for any $t_1\neq t_2$ and for any positions $\mathbf{p}$ and $\mathbf{q}$. In \ac{CaTP}, the small-scale fading is time-variant only if the environment is dynamic. Otherwise, is it considered time-invariant.
\par
When the small-scale fading is time variant, the temporal variation can be characterized with the coherence time $\tau$, which indicates the maximum duration over which the small-scale fading term remains almost constant. One common way to model the  temporal variation of small-scale fading is as follows: $h(\mathbf{p},\mathbf{q},t)$$=$$h_k$, for $t\in[k\tau,(k+1)\tau)$ with $\{h_k\}_k$ being a sequence of independent and identically distributed (i.i.d.) random variables. More dynamic environments have shorter coherence times $\tau$.
\par
After discussing the concept of time-variance for small-scale fading in the context of \ac{CaTP}, we now proceed to first discuss time-invariant small-scale fading models, and then briefly discuss time-variant models.
\par
We start with the statistical distribution of small-scale fading. When there is no line of sight between the transceivers nor any particular strong dominant wave due to some reflection, $h(\mathbf{p},\mathbf{q},t)$ is commonly modelled as a zero-mean complex circular Gaussian random variable. Hence, $|h(\mathbf{p},\mathbf{q},t)|$ is a Rayleigh distributed random variable, which explains why such a model is referred to as Rayleigh fading \cite{MPC1}.
\par
Alternatively, if there is a line of sight between both transceivers, then $h(\mathbf{p},\mathbf{q},t)$ can be modelled as a non-zero-mean complex circular Gaussian random variable:
\begin{equation}
\label{eq:MPC:2}
h(\mathbf{p},\mathbf{q},t)=h_R+jh_I+L,
\end{equation}
where $h_R$ and $h_I$, which are zero-mean independent real Gaussian random variables with variance $1/2$, represent the ensemble of scattered waves arriving to the receiver's antenna, and $L$ is a real number representing the strength of the line of sight component. The ratio $K=L^2/\mathrm{var}(h_R+jh_I)$ is called the Rician factor. If $K\neq 0$, then we say that we have Rician fading \cite{bookComm2} and $|h(\mathbf{p},\mathbf{q},t)|$ is a random variable with a Rician probability distribution function. But, when $K=0$, we have Rayleigh fading.
\par
There are other common probabilistic distributions used to model $|h(\mathbf{p},\mathbf{q},t)|$, such as the Nakagami distribution. Unlike the Rayleigh and the Rician distributions, not all other models have physical interpretations, as some distributions are used only because they fit the experimental results well.
\par
We now address the modeling of the spatial variations of small-scale fading. As the name indicates, the magnitude of $h(\mathbf{p},\mathbf{q},t)$ varies significantly over very small distances. Many experiments show that the small-scale fading coherence distance is usually lower than $\approx \lambda/2$. There is no universal model to describe its spatial correlation \cite{Mob9}, but there is an important theoretical model derived by Jakes \cite{bookComm3} for the case where the receiver is surrounded by a ring of uniformly distributed scatterers. In this scenario, the small-scale fading has a Rayleigh distribution and the following normalized spatial correlation:
\begin{equation}
\label{eq:MPC:3}
r(\mathbf{p},\mathbf{q})=J_0\left({2\pi\|\mathbf{p}-\mathbf{q}\|_2}/{\lambda}\right),
\end{equation}
where $J_0()$ is the Bessel function of the first kind and zero-th order. It is possible to simulate the 2D random field representing the small-scale fading with the Jakes model using numerical techniques such as those of \cite{CommSim2}. If the conditions of the environment differ from those required in the Jakes model, (\ref{eq:MPC:3}) would not be a good model for the spatial correlation function. Alternative models have been developed in the literature { see \cite{bookComm2}}, but as with the Jakes model, their decorrelation distance is generally around $\lambda/2$.
\par
\subsubsection{Data driven Models}
Another approach consists of using machine learning techniques to {\it learn} the communications channel. For instance in \cite{CATP8}, in the context of an iterative \ac{CaTP} problem, the authors use Gaussian processes to construct a radio map of the wireless channel. This radio map is iteratively updated with channel measurements. In \cite{comm7}, the authors predict the \acf{RSS} of the link between a \ac{UAV} and a \ac{BS} using the ensemble method technique, which combines various machine learning models. In \cite{comm8}, the authors propose a machine learning technique to learn the channel map of a defined region using segmentation. 
\par
In \cite{comm9}, the authors developed a machine learning technique that uses satellite images, determines the position of trees, and then uses an artificial neural network to determine the channel loss depending on the transceivers' positions. Some other works also use artificial neural networks to predict the channel loss \cite{comm10}. The radio maps constructed using machine learning techniques can be highly accurate in the regions where the measurements have been taken, but obtaining such measurements can be costly and time consuming. In addition, these are often numerical maps without analytical expressions, thus making them unsuitable for mathematical analysis that could provide useful insight into how to solve \ac{CaTP} problems. In the rest of this section, we will focus on analytical channel models. The reader interested in machine learning techniques for channel modelling can find more information in \cite{UAV12}.
\par
\subsection{UAV Channel Models}
\label{Comm:UAVchannel}
The channel models described in the previous section were originally developed for mobile communications considering ground users and are well-suited to ground \acp{MR} applications. In general, they are not appropriate for \acp{UAV} applications mainly because of the changing altitude of the \ac{UAV}. Given the increasing attention being paid to the integration of \ac{UAV}s into 5G networks \cite{Mishra2020CN,uav26b}, for the sake of completeness of this tutorial, we next discuss briefly the modeling of \ac{UAV} wireless channels, which is currently an active research topic \cite{uav15,UAVChannelsurvey2}. 
Before presenting the \ac{UAV} channel models, let us discuss some particularities of \ac{UAV} communications. For small-sized UAVs, the receiver is close to the \ac{UAV}'s power electronics and to the motors which run continuously. As a consequence, sometimes the \ac{UAV}'s motors generate electromagnetic noise that can interfere with the \ac{UAV}'s own receiver \cite{UAVnoise1}. 
\par
 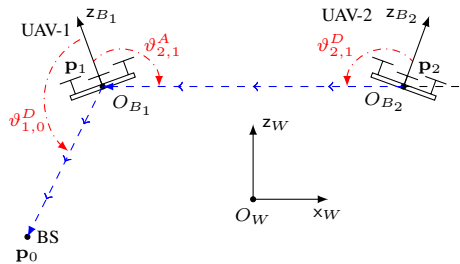
\begin{figure}
    \begin{center}
        \scalebox{1}{
		\begin{tikzpicture}
		    \tikzset{->-/.style={decoration={
                markings,
                mark=at position #1 with {\arrow{>}}}, postaction={decorate}}
            }
            
            \node (p_UAV1) at (-0.35,0.25) {\scriptsize $\mathbf{p}_1$};
            \node (p_UAV2) at (4.35,0.25) {\scriptsize $\mathbf{p}_2$};
            \node (p_BS) at (-1,-2.25) {\scriptsize $\mathbf{p}_0$};
		    
		    \node (UAV1) [quadcopter side, fill=white, draw=black, minimum width=1.5cm, rotate=20] at (0,0) {};
		    \node at (-0.75,0.75) [text centered]{\scriptsize UAV-1};
		    
			\node (UAV2) [quadcopter side, fill=white, draw=black, minimum width=1.5cm, rotate=-20] at (4,0) {}; 
			\node at (3.25,0.95) [text centered]{\scriptsize UAV-2};
			
			\draw[red, dash dot, latex-] (0.75, 0) arc (0:138:0.5cm) node at (0.80, 0.25) [above]{\scriptsize $\vartheta^A_{2,1}$};
			
			\draw[red, dash dot, -latex] (-0.3, 0.625) arc (118:230:0.90cm) node at (-1, -0.75) [above]{\scriptsize $\vartheta^D_{1,0}$};
			
			\draw[red, dash dot, latex-] (3.25, 0) arc (180:42:0.5cm) node at (3.10, 0.25) [above]{\scriptsize $\vartheta^D_{2,1}$};
			\draw[dashed] (4,0) -- (4.75,0);
			
			\node (circle) at (UAV1) [circle, draw, scale=0.2, fill=black] {};
			\node (circle) at (UAV2) [circle, draw, scale=0.2, fill=black] {};
			\node (circle) at (-1,-2) [circle, draw, scale=0.2, fill=black] {};
			\draw (-1,-2) node[right]{\scriptsize $\text{BS}$};
			\draw[-latex, dashed, blue, ->-=.5, ->-=.25, ->-=.75] (UAV2) -- (UAV1);
			
			\draw[dashed, blue, ->-=.5, ->-=.75, ->-=.25, -latex] (UAV1) -- (-1,-2); 
			
			\draw[-latex] (UAV2) node at (3.75,0.05) [below]{\scriptsize $O_{B_2}$} -- ($ (UAV2) + (0.342020143, 0.939692621)$) node[left]{\scriptsize $\zz_{B_2}$};
			\draw[-latex] (UAV1) node[below right]{\scriptsize $O_{B_1}$} -- ($ (UAV1) + (-0.342020143, 0.939692621)$) node[right]{\scriptsize $\zz_{B_1}$};
			
			\draw[-latex] (2,-1.5) node[below]{\scriptsize $O_W$} -- (2,-0.5) node[right]{\scriptsize $\zz_W$};
			\draw[-latex] (2,-1.5) -- (3,-1.5) node[below]{\scriptsize $\xx_W$};
			\node (circle) at (2,-1.5) [circle, draw, scale=0.2, fill=black] {};
			
		\end{tikzpicture}
		}
	\end{center}
    \caption{A chain of relays composed of two \acp{UAV} and one ground \ac{BS}. In blue the \ac{LoS} components, and in red the \ac{AoA} and the \ac{AoD} for each \ac{UAV} \cite{BonillaEUSIPCO21}.}
    \label{fig:UAV:AoA}
    \end{figure} 
\par
Another phenomenon is airframe shadowing \cite{uav15}. This occurs when the frame of the \ac{UAV} itself partially blocks the \ac{LoS}. 
Consider a multirotor \ac{UAV} with an antenna on the top surface (i.e. the \ac{UAV}'s surface facing the sky) that is communicating with a ground node. Assume that the \ac{UAV} moves away from the node. To do this, the multirotor \ac{UAV} has to tilt in such a way that its bottom surface (i.e. the \ac{UAV}'s surface facing the ground) is slightly orientated towards the ground node, see Fig. \ref{fig:UAV:AoA}. This can fully or partially block the \ac{LoS} between the antennas of the ground node and of the \ac{UAV}. In the case of fixed-wing \acp{UAV}, airframe shadowing can occur when the \acp{UAV} turn. In turning, they usually change their roll by controlling their ailerons. During this manoeuvre, one wing tilts up and the other tilts down. This tilting might temporarily block the \ac{LoS} with other communication nodes. The airframe shadowing severity, for both types of \acp{UAV}, depends on the airframe or wings material, its size, its shape, antenna location on the \ac{UAV}'s frame, and \ac{UAV} trajectories. This phenomenon has been observed in practice; but, as mentioned in \cite{uav15,UAVChannelsurvey2}, it has not yet been fully studied.
\par
The communications channel gain depends on the relative orientation of the transmitting and receiving antennas. During the flying phase, a multirotor \ac{UAV} must tilt, thus changing its antenna orientation. As a consequence, the communication channel observed when a multirotor \ac{UAV} hovers is different than when they move \cite{UAVantenna1}, see Fig. \ref{fig:UAV:motion}. Furthermore, the contribution on the antenna channel gain will vary with the motion of the \ac{UAV}, see \cite{BonillaEUSIPCO21} for more details. Similarly, during turning manoeuvres, a fixed-wing \ac{UAV} has to tilt, thus changing its antenna orientation, see Fig. \ref{fig:Aircraft}. The communications channel observed when fixed-wing \acp{UAV} move on a straight line is different than when they are turning. We also note that the location and orientation of the antenna on the \ac{UAV} has a significant impact on the communications channel, as shown experimentally in \cite{comm6,UAVantenna2,UAVantenna3,UAVantenna4}.

\par
    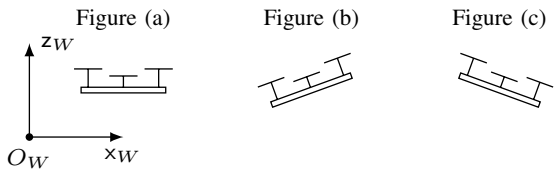
\begin{figure}
    \scalebox{1.25}{
		\begin{tikzpicture}
		    \tikzset{->-/.style={decoration={
                markings,
                mark=at position #1 with {\arrow{>}}}, postaction={decorate}}
            }

		    \node (UAV1) [quadcopter side, fill=white, draw=black, minimum width=1.5cm, rotate=0] at (-2,0) {};
		    \node at (-2.00,0.75) [text centered]{\scriptsize Figure (a)};
		    
		    \node (UAV1) [quadcopter side, fill=white, draw=black, minimum width=1.5cm, rotate=20] at (0,0) {};
		    \node at (0.,0.75) [text centered]{\scriptsize Figure (b)};
		    
			\node (UAV2) [quadcopter side, fill=white, draw=black, minimum width=1.5cm, rotate=-20] at (2,0) {}; 
			\node at (2.00,0.75) [text centered]{\scriptsize Figure (c)};
			
			\draw[-latex] (-3,-0.5) node[below]{\scriptsize $O_W$} -- (-3,0.5) node[right]{\scriptsize $\zz_W$};
			\draw[-latex] (-3,-0.5) -- (-2,-0.5) node[below]{\scriptsize $\xx_W$};
			\node (circle) at (-3,-0.5) [circle, draw, scale=0.2, fill=black] {};
			
		\end{tikzpicture}
	}
    \caption{(a) quad-rotor hovering, (b) quad-rotor moving to the left, and (c) quad-rotor moving to the right.}
    \label{fig:UAV:motion}
    \end{figure} 
\par

The \ac{UAV} channel models can be divided into two types. The first consists of air-to-ground channels, which characterize the channels between a \ac{UAV} and ground users or ground \ac{BS}. The second consists of air-to-air channels, which characterize the channels between flying \acp{UAV}.

\begin{figure}[ht]
    \centering
   \includegraphics[scale=0.36]{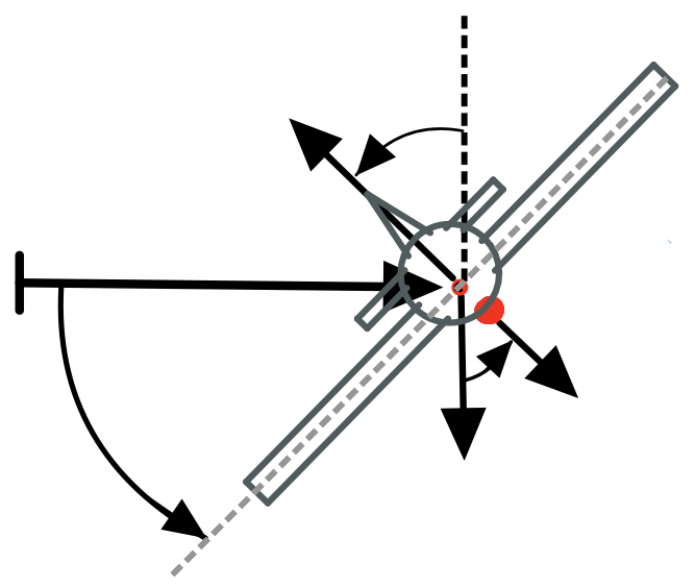}
\put(-60,84){\footnotesize $\varphi_{e}$}
\put(-86,86){\footnotesize $F_{\ell}$}
\put(-90,56){\footnotesize $R$}
\put(-100,30){\footnotesize $\varphi_{e}$}
\put(-34,24){\footnotesize $\varphi_{e}$}
\put(-20,40){\footnotesize $\vec{\eu{n}}$}
\put(-56,10){\footnotesize $mg$}

    \caption{Fixed-wing UAV; the UAV antenna is indicated by a red dot. Front view of fixed-wing \ac{UAV} performing a steady turn with a bank angle ${\varphi}_{e}$. 
    }
\label{fig:Aircraft}
\end{figure}
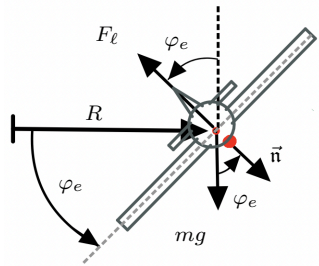

\subsubsection{Air-to-ground channels}
\label{A2GCommChannels}
The air-to-ground channel communication models the link between a \ac{UAV} and a ground user, such as a control station or a 5G \ac{BS}. The properties of this channel depend not only on the distance between both nodes, but also on the \ac{UAV} altitude and on the {\bf elevation angle}\footnote{The angle of vector $\mathbf{p}_{UAV}-\mathbf{p}_{G}$ measured w.r.t. the horizontal plane, where $\mathbf{p}_{UAV}$ and $\mathbf{p}_{G}$ is the position of the ground node.}. As the elevation angle increases, the probability of \ac{LoS} between both nodes also increases. As such, \ac{LoS} links present lower losses than non-\ac{LoS} links. As the \ac{UAV} altitude increases, the elevation angle increases along with the distance between the nodes. For instance, one illustrative model that describes such effects is \cite{UAV13}:
\begin{eqnarray}
\label{eq:UAVcomm:1}
\mathcal{H}(\mathbf{p}(t),\mathbf{q}(t))[dB]&=&-20\log\left(4\pi \lambda \|\mathbf{p}(t)-\mathbf{q}(t)\|_2\right)\nonumber\\
&+&\xi(\mathbf{p}(t),\mathbf{q}(t)),
\end{eqnarray}
where the first term represents the average path loss and $\xi(\mathbf{p}(t),\mathbf{q}(t))$ represents the shadowing. If there is line of sight, then $\xi(\mathbf{p}(t),\mathbf{q}(t))\sim\mathcal{N}(\mu_{LOS},\sigma_{\rm LOS}^2)$. Otherwise,   $\xi(\mathbf{p}(t),\mathbf{q}(t))\sim\mathcal{N}(\mu_{NLOS},\sigma_{NLOS}^2)$. To complete this channel model, we must also consider the probability of having line of sight, which can be expressed as:
\begin{equation}
\label{eq:UAVcomm:2}
\mathrm{P}_{\rm LOS}(\mathbf{p},\mathbf{q})=\frac{1}{1+a\exp\left(-b\left[\frac{180}{\pi}\sin^{-1}\left(\frac{z_{UAV}}{\|\mathbf{p}-\mathbf{q}\|}\right)-a\right]\right)},
\end{equation}
where $z_{UAV}$ is the altitude of the \ac{UAV}, $\mathbf{p}_{UAV}$ is its position and $\mathbf{q}$ is the position of the ground node. The coefficients $a$ and $b$ are environment-dependent. 
\par
The model (\ref{eq:UAVcomm:1})-(\ref{eq:UAVcomm:2}) does not take into account small-scale fading. As mentioned in \cite{UAVChannelsurvey2}, small-scale fading in air-to-ground channels often follows a Rician distribution whose properties depend on the \ac{UAV} altitude and the surroundings of the ground node. There are also approaches which use 3D numerical maps of the region of operation to determine the channel model \cite{Hajar2018}. We refer the reader to the survey in \cite{UAVChannelsurvey2} for more detailed information about the air-to-ground \ac{UAV} communications channels.
\subsubsection{Air-to-air channels}
The modeling of ground-to-air channels is not yet well studied, but the situation for air-to-air channels (communications channels between flying \acp{UAV}) is worse still as there are fewer studies and measurement campaigns focusing on these types of channels \cite{uav15}. When the altitude of two \acp{UAV} is low, the floor and other objects on the surface (e.g. hills, buildings, and trees) influence the channel. However, as the altitude of the \acp{UAV} increases, such influences weaken. As the altitude of both \acp{UAV} increases, the channel tends to consist of a \ac{LoS} and tends to behave like the {\it free-space} channel described in section \ref{Comm:channel:deterministic}.
\par
It has been observed experimentally that when the \acp{UAV} operate in an open field with a floor that is flat enough, the propagation channel follows a two-ray model \cite{uav15}. As the altitude of the \acp{UAV} increases, the strength of the reflected path decreases. Further, when the \acp{UAV} are operating over bodies of water, such as lakes, the strength of the reflected path is stronger than when operating over land.
\par
The constructive and destructive interference between the \ac{LoS} and reflected waves described by the two-ray model can statistically be described with the small-scale fading. This is the approach taken in \cite{UAV14}, where it was shown experimentally that the air-to-air channel model can be modelled as:
\begin{equation}
\label{eq:UAVcomm:3}
\mathcal{H}(\mathbf{p}(t),\mathbf{q}(t),t)=\frac{h(\mathbf{p}(t),\mathbf{q}(t),t)}{L_P(\mathbf{p}(t),\mathbf{q}(t))},
\end{equation}
where the pathloss term $L_P(\mathbf{p}(t),\mathbf{q}(t))$ follows the {\it free space} model, and the small-scale fading term is well described by a Rician distribution with the height-dependent parameters:
\begin{equation}
p_h(x)=\frac{x}{\sigma_0^2}\exp\left(\frac{-x^2-\rho^2}{2\sigma_0^2}\right)I_0\left(\frac{x\rho}{2\sigma_0^2}\right),
\end{equation}
where $x\geq 0$; $\rho$ and $\sigma$$=$$ah^b+c$ are the strength of the dominant and scattered components, respectively; $h$ is the altitude of the UAVs (both \acp{UAV} are assumed to have the same altitude); and with $a$, $b<0$, and $c$ being parameters to be fitted numerically according to the environment. Finally, we refer the reader to the survey in \cite{uav15} for more detailed information about the air-to-air \ac{UAV} communication channels.

\subsection{Performance metrics}
\label{Comm:metrics}
In \ac{CaTP}, the communication aspects are taken into consideration in the optimization problem in either the target function or the constraints. This requires the use of metrics to quantify the performance of wireless communication. Next, we discuss the metrics that are most relevant to \ac{CaTP}.

{\bf Channel quality}: one way in which the channel quality can be assessed is by measuring its power by using the \acf{CNR} and the \acf{SNR}. The instantaneous \ac{CNR} \cite{CATP3} at the receiver is given by:
\begin{equation}
\mathrm{CNR}(\mathbf{p}(t),\mathbf{q}(t),t)={|\mathcal{H}(\mathbf{p}(t),\mathbf{q}(t),t)|^2}/{\sigma^2},
\end{equation}
where $\mathbf{p}(t)$ and $\mathbf{q}(t)$ are the positions of the transmitter and the receiver, respectively, and $\sigma^2$ is the power of the thermal noise generated at the receiver. The mean \ac{CNR} is $\mathbb{E}\left[\mathrm{CNR}(\mathbf{p}(t),\mathbf{q}(t),t)\right]$, where the expected value is taken w.r.t. small-scale fading and shadowing. The \ac{SNR} is the \ac{CNR} multiplied by the transmission power.
\par
Another important metric is the \acf{RSS} \cite{Zanella2016IEEESurvey}. It measures the power of the received signal, with the reason behind its importance being that many receivers in the market provide the \acf{RSSI} (a discretized version of the \ac{RSS}). In practice, many algorithms resort to working with the \ac{RSSI}. 
\par
{\bf Error rate}: in some cases, one can be more interested in the performance of the communication system, rather than in the channel quality. The principal way of evaluating the performance of a communication system is by quantifying the errors that occurred in the transmission. This is done by calculating the \acf{BER} or by the \acf{PRR} \cite{CATP9}, which measures how many packets were successfully received without errors. These metrics evaluate the performance of the communication system, and thus strongly depend on the specifications of the system, including the  modulation scheme and the communication protocols.

{\bf Channel capacity}: the channel capacity was first introduced by Claude Shannon in 1948 in \cite{Shannon1948}. It is an upper bound on the maximum bit rate, given channel conditions, for which an arbitrarily low \ac{BER} can be reached. Its mathematical expression depends on the information that the transmitter and the receiver have about the channel, as well as on the power transmission strategy. When the channel is estimated at the receiver and the transmitter uses constant transmission power, then the channel capacity is given by \cite{bookComm4}:
\begin{equation}
C=B\int_{0}^\infty\log_2\left(1+\gamma\right)p(\gamma)\mathrm{d}\gamma,
\end{equation}
where $\gamma$ is the \ac{SNR}, $p(\gamma)$ is its p.d.f., and $B$ is the bandwidth of the transmitted signal.
\subsection{Energy consumption}
\label{sec:comm:energ}
Although often the \ac{MR}'s energy consumption is mostly due to motion, this is not always the case, as shown in the experiments presented in \cite{Mob2}. The $\mu$controller, the CPU, and the sensing systems can consume an amount of energy which may be comparable to the amount of energy spent in motion. The same applies to the \ac{MR}'s communication system. The energy consumed by the communication system is mainly determined by the transmission power. Next, we discuss transmission strategies.

{\bf Constant power transmission}: all signals are transmitted with constant average power. Thus, the transmitter's energy consumption depends on the transmission time, but is independent of the transmitter and receiver positions. In this scenario, the \ac{SNR} depends on the positions of the transmitter and the receiver through the spatial variations of the channel.

{\bf Adaptive power transmission}: the transmitter adapts its power to ensure a certain power $P_{ref}$ for the received signal. The power is adapted as follows:
\begin{equation}
\label{eq:Energy:1}
P={P_{\rm ref}}/{|\mathcal{H}(\mathbf{p}(t),\mathbf{q}(t),t)|^{2}}.
\end{equation} 
$P_{\rm ref}$ is selected to ensure a certain quality of service. The transmission power in (\ref{eq:Energy:1}) can take any arbitrarily large value. In practice, the transmission power is limited to a certain value $P_{max}$ by the hardware limitations, or by legal regulations. If the intended transmission power is smaller than $P_{max}$, then the actual transmission power is (\ref{eq:Energy:1}). Alternatively, if the intended transmission power is larger than $P_{max}$ (usually due to a poor channel gain), then it is not possible to satisfy the desired power $P_{ref}$ at the receiver. The most common action in this scenario is to refrain from transmitting and say that an outage has occurred. Thus, the transmission power can be written as:
\begin{equation}
\label{eq:Energy:2}
P=\Bigg\{
\begin{array}{cc}
\frac{P_{\rm ref}}{|\mathcal{H}(\mathbf{p}(t),\mathbf{q}(t),t)|^{2}}     & \mathrm{for}\  \frac{P_{\rm ref}}{|\mathcal{H}(\mathbf{p}(t),\mathbf{q}(t),t)|^{2}}\leq P_{\rm max},\\
0     &  \mathrm{otherwise}.
\end{array}
\end{equation}
\par
For mathematical simplicity and tractability, it is common to assume $P_{\rm max}\rightarrow\infty$ (i.e. disregard the power limitation).
\par
Finally, the energy consumption at the receiver is generally considered constant. It is usually significantly lower than that of the transmitter, and is hence often neglected in the analysis.
\par
\subsection{Final comments on communications models}
\label{sec:comm:FinalComments}
We provided an overview of communication systems and basic channel models for both ground \acp{MR} and \acp{UAV}. We also discussed the special characteristics of \ac{UAV} channel models. Another special scenario that can be important for some \ac{CaTP} applications, such as rescue and search operations, regards communication inside tunnels and underground mines. These types of channel models are outside the scope of this tutorial, although interested readers are referred to \cite{YarkanIEEESurveyComm2009,comm3,HrovatIEEESurveyComm2014}.

The models presented in this section are narrowband models, which are the most basic models. More complex and elaborate models include wideband communication channel models, which take into consideration how the channel varies w.r.t. the frequency. Also, we must mention that when one of the transceivers moves quickly, Doppler shift occurs and affects the received signals. This phenomenon can be of particular importance when dealing with fixed-wing \acp{UAV} as they are capable of high speed flight. 

{ We conclude this section by bringing to the attention of the reader that one of the most determinant elements for the adequate selection of the communications channel model is the environment where the system will operate. To illustrate this, lets consider a pair of \acp{UAV} hovering at low altitude over a body of water while communicating. On one hand, if the body of water is a calm lake then it will act as a mirror and reflect well the electromagnetic waves. Consequently, the two-ray model described in section \ref{Comm:channel:deterministic} could be a good selection for the communications channel model since it would  consider the specular reflection occurring on the surface of the lake. On the other hand, if the body of water is a rough sea, then the irregular and dynamic surface of the rough sea would provoke that the characteristics and the number of reflected waves would also be time variant. But the two-ray model would fail to consider this, and thus it would most probably be highly inaccurate in this case. Hence the importance of taking into account the environment for the selection of the communications channel model.}
{
\section{Communications-aware Trajectory Planning}
\label{sec:CATP}
As mentioned in section \ref{sec:Intro}, \ac{CaTP} refers to designing trajectories where both communications aspects (e.g. communication energy consumption, quality of the communications link, number of transmitted bits) and robotics aspects (e.g. kinematic and dynamic constraints, motion energy consumption) are simultaneously considered. We begin with some preliminary concepts. A particular case of \ac{CaTP} is the \acf{CaPP} problem. The trajectory and path are defined as follows:
\begin{definition}[Path]
The {\bf path} followed by a robot from time instant $t_1$ to $t_2$ is defined as $\mathcal{P}(\mathbf{p}(t),t_1,t_2)\triangleq\{\mathbf{p}(t)|t\in[t_1,t_2]\}$, where $\mathbf{p}(t)$ is the position of the robot at time $t$.
\end{definition}
\begin{definition}[Trajectory]
The {\bf trajectory} followed by a robot from time instant $t_1$ to $t_2$ is the description of the robot's position at each instant $t\in[t_1,t_2]$ and can then be described as $\mathcal{T}(\mathbf{p}(t),t_1,t_2)=\{\mathbf{p}(t),t|t\in[t_1,t_2]\}$.
\end{definition}
The {\bf path} consists of the collection of all the points visited by the \ac{MR} without specifying when and without any velocity information. The {\bf trajectory} gives not only information about the points visited by the \ac{MR}, but also a temporal description of the change of the \ac{MR}'s position over time. The trajectory can also be considered as the combination of the path and the velocity profile.
\par
From the application point of view, a \ac{CaTP} problem can be categorized as either a \ac{CaR} or a \ac{RaC} task. {This classification pertains only the nature of the application and it is not related to the mathematical formulation directly. In \cite{GeraciIEEECST2022}, the authors present the similar categorization in the context of \ac{UAV} communications, although they don't use the terms \ac{CaR} and \ac{RaC}, and they describe both categories very concisely: {\it 'what can \acp{UAV} do for networks'} and  {\it 'what can networks do for \acp{UAV}'}.  Thus, we can follow a similar strategy, and say that \ac{RaC} are applications that result from the question '{\it what can robots do for networks?}' while \ac{CaR} are applications that result from the question '{\it what can networks do for robots?}'.
}
\par
We can also classify \ac{CaTP} problems depending on what is being optimized into three categories. In the first category, the objective is to design the \ac{MR}'s full trajectory, i.e., path and velocity profile; see for example \cite{CATP19}. In the second category, the translational velocity is given (e.g. constant speed) and the objective is to design the MR's path; see for example \cite{UAV23}; this is a \ac{CaPP} problem. In the third category, the path is given and the objective is to design the translational speed profile with which the \ac{MR} will follow the given path; see for example \cite{CATP9}.
\par
\ac{CaTP} problems can also be divided according to the trajectory type: {\bf predetermined} and {\bf adaptive} (also referred to as reactive). Predetermined trajectories are designed before they are executed by the \ac{MR}, e.g. \cite{CATP19}. Adaptive trajectories are created and modified as the robot executes them, e.g. \cite{BonillaEUSIPCO21}; the \ac{MR} may either use a predetermined trajectory as an initial guide and then implement an adaptive mechanism to adapt it in real-time in response to new information obtained from the environment, or create the trajectory in real-time while operating. Due to the elementary nature of this tutorial, we will mostly focus on predetermined trajectories, whose theory constitutes the base for adaptive ones.
\par
The optimization method and the resulting optimal trajectory strongly depend on the available information about the wireless channel. The following are some possibilities.


{\bf i. Physical map available}: in this case the \ac{RF} map or its estimate is known. In other words, we know $\mathcal{H}(\mathbf{p},\mathbf{q}(t),t)$ in Eq. (\ref{chann:1}) for all $\mathbf{p}\in\mathcal{W}$ and all $t\in[0,T]$ where $\mathcal{W}$ is the workspace where the \ac{MR} operates and $\mathbf{q}(t)$ is the position of the other node. 
\par
For instance, in \cite{UAV24}, the authors study the performance of different 3D path planners for a single \ac{UAV} communicating over a cellular network. They resort to a realistic \ac{RF} map to accurately evaluate the connectivity and the interference resulting from each path planning algorithm. This map is constructed as follows. First, a public 3D map of the region of Flanders, Belgium  \cite{GeopuntSite} is used. This map was generated by Lidar, with a latitude and longitude resolution of 1m$\times$1m, and each pixel contains the height of the surface. 
Then, using information from the Belgian Institute for Postal Services and Telecommunication, they located nineteen \acp{BS} of the Belgian mobile provider, {\it Proximus}, within the area of interest. The map was then fed into a 3D radio coverage simulator \cite{ColpaertSensors2018}. 
The latter uses both the 3D map and the location of the transceivers to determine whether or not a \ac{LoS} exists, thus accounting for possible shadowing. To account for multipath fading, the simulator adds a Gaussian random process to the overall channel with a variance dependent on the altitude of the \ac{UAV}. The resulting 3D \ac{RF} map allowed the authors to observe how changes in altitude can improve communications performance.
\par
If we have a physical 3D map of the workspace, then we can also obtain a 3D \ac{RF} map by implementing ray tracing \cite{RayTracing1}, \cite{RayTracing2}, \cite{RayTracing3}. This type of \ac{RF} map captures the shadowing (due to obstruction of the \ac{LoS}), and can also capture multipath fading. 2D physical maps can also be used and can be useful particularly for ground \acp{MR} operating indoors to accurately determine when walls obstruct the \ac{LoS}. 

{\bf ii. \ac{RF} channel measurements available}: in this case, we do not have a physical map of the workspace where the \acp{MR} operates, but we have a number of \ac{RF} channel measurements taken at different locations in the workspace. Thus, prior to solving the \ac{CaTP}, we feed the channel measurements into a predictor to create an \ac{RF} map that will be used by the trajectory planner. In \cite{YanACC2013}, the authors consider a ground \ac{MR} that must visit a number of points of interest, gather data there, and then transmit all collected data to a \ac{BS}. This must be done while minimizing the total energy consumed. The \ac{MR} initially disposes of few channel measurements taken in the workspace. These measurements are then fed into the probabilistic channel assessment framework developed by the authors in \cite{comm1} and \cite{comm2}. This framework is used by the path planning algorithm to predict the channel gain throughout the workspace. The predicted probabilistic \ac{RF} map is constructed using the pathloss model, the \ac{PDF} of the multipath fading, and the \ac{PDF} of the shadowing as well as its spatial correlation. 

{\bf iii. Mathematical model available}: in this case, we only dispose of a mathematical model of the communications channel. In our previous work \cite{CATP19}, we considered the problem of a multirotor \ac{UAV} that must reach some goal while transmitting data to a \ac{BS}. The only information about the communications channel used for solving the communications-aware trajectory planning was the pathloss model and the \ac{PDF} of the shadowing. In \cite{UAV18}, the authors considered the problem of optimizing the position of a \ac{UAV} operating as a \ac{BS}. To solve this, they considered the pathloss model, which is complemented by the \ac{PMF} of the \ac{LoS}. In \cite{MD1}, we considered the problem of mitigating the small-scale fading in an \ac{MR} communications link by leveraging the knowledge of its \ac{PDF} and spatial correlation. \qedsymbol{}

\subsection{General Structure} 
\label{sec:generalStructure}
The \ac{CaTP} optimization problem to determine a predetermined trajectory of a single \ac{MR} can be formulated as:
\begin{equation}
\label{TPA:eq:1}
\begin{array}{l}
\displaystyle\minimize_{\mathcal{C},\mathcal{T}}\  J(\mathcal{C},\mathcal{T})\\
{\rm s.t.}\\
{\rm motion\ model},\\
{\rm channel\ model},\\
{\rm trajectory\ constraints},\\
{\rm communications\ constraints},
\end{array}
\end{equation}
where $\mathcal{C}$ is the set of all communication-related parameters to be optimized, such as modulation order or transmission power\footnote{If all of the communications parameters are fixed, then $\mathcal{C}=\emptyset$.}; and $\mathcal{T}$ is the set of variables related to the trajectory of the robot, such as the completion duration of the trajectory or the waypoints that form the robot's path.
\par
For instance, in \cite{UAV6}, 
 $\mathcal{T}$ consists of the set of \ac{UAV}'s 3D hovering locations as well as their visiting order, and communications-related inputs were considered, i.e., $\mathcal{C}=\emptyset$. 
In \cite{YanACC2013}, 
$\mathcal{T}$ consists of the set of waypoints forming the robot's path, and $\mathcal{C}$ describes the spectral efficiency (i.e. the bit rate normalized by the bandwidth) of the robot's transmitter. In \cite{UAV18}, 
 $\mathcal{T}$ describes the 3D position of the \ac{UAV}, and $\mathcal{C}$ is the set of users that the \ac{UAV} communicates with. 
In \cite{Ahmed2020CL}, 
the inputs forming $\mathcal{T}$ are the position, velocity, and acceleration profiles of the \ac{UAV}, and  those forming $\mathcal{C}$ are the transmission powers of the \ac{UAV} and the \ac{BS}. 
In \cite{CATP19}, 
$\mathcal{T}$ consists of the control signals for the \ac{UAV} motors, and 
$\mathcal{C}=\emptyset$. In \cite{HeIEEECL2018}, 
$\mathcal{T}$ consists of the altitude of the \ac{UAV},  and $\mathcal{C}$ consists of the beamwidth of its antennas.

\par
The generic \ac{CaTP} problem in (\ref{TPA:eq:1}) is composed of five elements: the optimization target $J(\mathcal{C},\mathcal{T})$, a  motion model for the MR describing how its {\it state} behaves according to the control signal $\mathbf{u}$ (discussed in Section \ref{sec:RMod}), a wireless channel model describing how the received and transmitted signals behave depending on the position and orientation of the \ac{MR} (discussed in section \ref{sec:Mod:comms}), some trajectory related constraints, and constraints related to the communication system performance and/or goals. We will next discuss  the elements of the \ac{CaTP} problem that have not been discussed in previous sections. 



\par
\subsection{ Optimization target}
\label{sec:optimizeationTarget}
By convention, optimization problems are usually stated as minimization problems \cite{boyd_vandenberghe_2004}, as in (\ref{TPA:eq:1}). However, sometimes (e.g. when the focus is on energy efficiency or on maximizing the number of users served), they are initially cast as maximization problems \cite{Ahmed2020CL}, \cite{mobb10}, \cite{UAV18}. Nevertheless, any  maximization problem can be converted into an equivalent\footnote{Two optimization problems are equivalent if, and only if, they have exactly the same solution.} minimization problem by proceeding as follows \cite{boyd_vandenberghe_2004}:
\begin{equation}
\label{eq:opt1}
    \begin{array}{ccc}
     \displaystyle\maximize_{\mathcal{C},\mathcal{T}}\  J(\mathcal{C},\mathcal{T})    & \rightarrow & \displaystyle\minimize_{\mathcal{C},\mathcal{T}}\  -J(\mathcal{C},\mathcal{T}).
    \end{array}
\end{equation}
Another possibility is the following transformation:
\begin{equation}
\label{eq:opt2}
    \begin{array}{ccc}
     \displaystyle\maximize_{\mathcal{C},\mathcal{T}}\  J(\mathcal{C},\mathcal{T})    & \rightarrow & \displaystyle\minimize_{\mathcal{C},\mathcal{T}}\  J^{-1}(\mathcal{C},\mathcal{T}).
    \end{array}
\end{equation}
The linear transformation in (\ref{eq:opt1}) maintains the properties of the original optimization target regardless of its nature. The nonlinear transformation (\ref{eq:opt2}) deforms the original optimization target by expanding it for values in the region $[-1,1]$ and compressing it outside of this interval. Furthermore, if the original optimization target $J(\mathcal{C},\mathcal{T})$ can take a value of zero, then $J^{-1}(\mathcal{C},\mathcal{T})$ will not be defined at this value. Nevertheless, if $J(\mathcal{C},\mathcal{T})>0$ (or $J(\mathcal{C},\mathcal{T})<0$), then the nonlinear transformation (\ref{eq:opt2}) is not necessarily an issue. For instance, the authors in \cite{HeIEEECL2018} transform a minimization problem into a maximization problem by applying the nonlinear transformation (\ref{eq:opt2}) to a strictly positive optimization target. The authors in \cite{KalogeriasICASSP2016} move in the opposite direction of the nonlinear transformation (\ref{eq:opt2}); instead of minimizing the power consumption, they maximize the inverse of the power consumption to simplify the problem.
\par
If the inputs of the optimization problem (i.e. $\mathcal{C}$ and $\mathcal{T}$) are variables, then the optimization target is a function, e.g.  \cite{LiceaEUSIPCO2019b,CATP18}. On the other hand, if the inputs of the optimization problem are functions, then the optimization target is a functional \cite{Luenberger}, e.g. \cite{MD1,MD2,BonillaIEEETSP2016}. Further, if the optimization target is a function to be minimized, it is referred to cost function. Alternatively, if the optimization target is a function to be maximized, it is referred to as utility function \cite{boyd_vandenberghe_2004}. Similarly, if the optimization target is a functional to be minimized, it is referred to a cost functional \cite{CATP3,Ali2019TCNS}.
\par
If the set of inputs $\{\mathcal{C},\mathcal{T}\}$ contains continuous variables/functions and discrete variables/functions, the optimization problem is referred to as mixed-integer. A common class of mixed-integer optimization problems which often appears in \ac{CaPP} is the \ac{MILP} \cite{YanACC2013,LiuCDC2017,FlushingSSRR2016,GhaffarkhahGlobecom2012}.
\par
We next describe some functionals that have been used as  optimization targets. We start with the optimization target initially considered in \cite{CATP19} which aims to solve a \ac{CaTP} problem for a multirotor \ac{UAV} \footnote{We simplify the writing of the original expression of the optimization target for clarity purposes.}:
\begin{equation}
\label{eq:opt3}
   J=\vartheta \int_{0}^{t_f}\|\mathbf{u}(t)\|_2^2\mathrm{d}t+(1-\vartheta)w(\mathbf{p}), 
\end{equation}
where $t_f$ is the trajectory duration and $w(\mathbf{p})$ is an arbitrary functional related to a communications criteria which depends on the trajectory of the \ac{UAV}. The optimization target consists of the sum of two terms, the first one measuring the motion-induced energy consumption, while the second one is a communications-related term which depends on the bit rate over the \ac{UAV} trajectory. Parameter $\vartheta\in[0,1]$ determines the importance of one term over the other. The control signal for the \ac{UAV} is $\mathbf{u}(t)\in\mathbb{R}^4$, defined in $t\in[0,t_f]$ (i.e. $\mathbf{u}$ is a four dimensional vectorial function defined over a closed support). The 3D position of the \ac{UAV} is $\mathbf{p}(t)\in\mathbb{R}^3$, defined in $t\in[0,t_f]$ (i.e. $\mathbf{p}$ is a three dimensional vectorial function defined over a closed support). The model in \cite{CATP19} used to describe the relationship between the \ac{UAV} control signal $\mathbf{u}$ and its position $\mathbf{p}$ is as described in subsection \ref{sec:uav}. Therefore, the optimization target $J$ depends on two functions: the \ac{UAV} control signal $\mathbf{u}$ and the \ac{UAV} position $\mathbf{p}$ (which itself depends on $\mathbf{u}$). Furthermore, the optimization target is the convex combination of two functionals.
\par
Another interesting example can be found in \cite{BonillaIEEETSP2016}. In this case, a \ac{MR} disposes of a fixed time $T$ to harvest radio energy from a \ac{BS}. The channel experiences small-scale fading, and the radio energy harvested varies greatly over very small distances. Thus, the \ac{MR} uses the following procedure divided in three phases: during the first phase, the \ac{MR} moves along a line of length $L$ for a duration of $\alpha T_s$ with $\alpha\in(0,1)$ and $T_s<T$; during the second phase, the \ac{MR} moves to the position in the line with the highest channel gain; lastly, the \ac{MR} remains for the rest of the time (i.e. $T-T_s$) at that position harvesting radio energy. The objective is to optimize the velocity profile of the \ac{MR} to maximize the expected value of the harvested energy. Therefore, the optimization target is: 
\begin{eqnarray}
\label{eq:opt4}
    J&=&E_s(v^{I}_{_{L,\alpha, T_s}})+E_s(v^{II}_{_{L,\alpha, T_s}})+E_H(L,T_s)\nonumber\\
    &-&E_{m}(v^{I}_{_{L,\alpha, T_s}})-E_{m}(v^{II}_{_{L,\alpha, T_s}}),
\end{eqnarray}
where $v^{I}_{_{L,\alpha, T_s}}$ and $v^{II}_{_{L,\alpha, T_s}}$ are the speed profiles of the \ac{MR} in the first and second phases, respectively. They are both unidimensional continuous functions parametrized by the variables $L$, $\alpha$, and $T_s$. The support for $v^{I}_{_{L,\alpha, T_s}}$ is $t\in[0,\alpha T_s]$ and the support for $v^{II}_{_{L,\alpha, T_s}}$ is $t\in[\alpha T_s, T_s]$. The first term in (\ref{eq:opt4}) is the energy harvested during the first phase while using the speed profile $v^{I}_{_{L,\alpha, T_s}}$; the second term is the energy harvested during the first phase while using the speed profile $v^{II}_{_{L,\alpha, T_s}}$; the third term is the energy harvested at the optimum position found by the \ac{MR}. This energy depends on the length $L$ (the space where the \ac{MR} searched for the position with the largest channel gain) and the time spent harvesting energy at the position with the largest channel gain; the fourth term is the energy spent by the \ac{MR} in motion during the first phase; the fifth term is the energy spent by the \ac{MR} in motion during the second phase. From the above description, we can observe that the optimization target (\ref{eq:opt4}) is the sum of four functionals ($E_s(v^{I}_{_{L,\alpha, T_s}})$, $E_s(v^{II}_{_{L,\alpha, T_s}})$, $E_{m}(v^{I}_{_{L,\alpha, T_s}})$, and $E_{m}(v^{II}_{_{L,\alpha, T_s}})$) and one function $E_H(L,T_s)$ (which depends on the variables $L$ and $T_s$). In this problem, the functions $v^{I}_{_{L,\alpha, T_s}}$ and $v^{II}_{_{L,\alpha, T_s}}$ are actually parameterized by the variables $L$, $\alpha$, and $T_s$. Nevertheless, in \cite{BonillaIEEETSP2016}, we use calculus of variations \cite{KirkOptimalControl} to find the optimum shape of the speed profiles (i.e. the functions $v^{I}_{_{L,\alpha, T_s}}$ and $v^{II}_{_{L,\alpha, T_s}}$) parameterized with $L$, $\alpha$, and $T_s$. After this step, the functionals in (\ref{eq:opt4}) become functions of $L$, $\alpha$, and $T_s$. 

The general strategy for converting functionals into functions is to fix the shape of the function that constitutes the input to the functional. One way to do this is to derive the optimum shape, as described in the previous example. However, this is not always possible, at least not analytically, due to the complexity of the problem. An alternative is to use the method presented in \cite{CATP18}. In the latter, we considered a \ac{UAV} acting as a data ferry between a source and a destination. The objective was to optimize the UAV's periodic trajectory, of period $T$, to maximize an upper bound of the number of bits transferred. During the first half of the trajectory, the \ac{UAV} collects data from the source and delivers it to the destination in the second half. The corresponding functional\footnote{Written in a simplified manner.} to maximize is:
\begin{equation}
\label{eq:opt5}
    J=\int_0^{T/2}r({x}(t),{y}(t))\mathrm{d}t,
\end{equation}
where functions $x(t)$ and $y(t)$ represent the trajectory of the \ac{UAV} at time instant $t$ in the $x-y$ plane, and the integrand in (\ref{eq:opt5}) is the maximum instantaneous bit rate achievable at $(x,y)$. The integral represents the maximum amount of bits collected by the \ac{UAV} from the source.
The functional (\ref{eq:opt5}) depends on the unidimensional continuous functions $x(t)$ and $y(t)$, with support $t\in[0,T]$. Since the problem indicates that the optimum \ac{UAV} trajectory is periodic, one way to transform the functional (\ref{eq:opt5}) into a function is by constraining the shape of the functions $x(t)$ and $y(t)$ using a Fourier series of order $N$. 
After adding this constraint 
into the optimization problem, the functional (\ref{eq:opt5}) becomes a function of the Fourier series coefficients. 
The resulting optimum trajectory is suboptimal. It gets closer to the optimum trajectory as the order of the Fourier series grows, but this also increases the number of parameters to optimize.

Below we briefly review some of the communications-related and robotics-related terms used in the optimization target of \ac{CaTP} problems. 
In \cite{ZhangIEEECL2018},  
the communications-related term is the outage probability of the communication link. In \cite{UAV18}, the optimization target is the number of users served by a \ac{UAV} operating as an aerial \ac{BS}. In \cite{Al-HouraniIEEEWCL2014}, the communications-related term is the coverage radius of a \ac{LAP} acting as a \ac{BS} for ground users. 
In \cite{HeIEEECL2018}, 
the communications-related terms are the time that the \ac{UAV} takes to transmit a certain amount of data, and the total amount of bits transmitted by the \ac{UAV}. In \cite{MD1}, 
the optimization target includes the expected value of the channel gain. In \cite{YanACC2013}, 
the communications-related term considered in the optimization target is the energy spent in data transmission.

In \cite{UAV6}, the authors consider a multirotor \ac{UAV} and a cost function which has two robotics-related terms: the time that the \ac{UAV} takes to complete the trajectory and the time that it spends hovering over ground nodes. In \cite{YanACC2013}, the authors consider an \ac{MR} moving at constant speed, and a cost function which consists of the motion-induced energy consumption in the form of (\ref{eq:En:H2}). In \cite{Ahmed2020CL}, the authors aim to maximize the efficiency of a fixed-wing \ac{UAV} acting as a communications relay; the cost function represents the propulsion-induced energy consumption in the form of (\ref{eq:energy:fixed-wing}).


\par

\subsection{Trajectory constraints}
\label{sec:trajConstraints}
These constraints directly restrain the shape of the trajectory and the time in which it is completed. 
\par
One simple constraint in this category defines the area where the \ac{MR} can operate. 
For example, the authors in \cite{Mob9} constrain the area where the \ac{MR} can move by 
 $   \mathbf{p}(t)\in \mathcal{W}/\mathcal{Q}_t$,
where 
 $\mathbf{p}(t)\in\mathbb{R}^2$, 
$\mathcal{W}\subset\mathbb{R}^2$ is the workspace and $\mathcal{Q}_t$ is the set of points where channel measurements have been taken up to the instant $t$. 
In \cite{LiuCDC2017}, 
the authors define the environment $\mathcal{X}$ as a large convex polygonal subset of the 2-D Euclidean space $\mathbb{R}^2$, define $\mathcal{X}_{obs}\subseteq\mathcal{X}$ as the
regions in the environment occupied by polygon obstacles. 
\par
The shape of the trajectory can be constrained to polynomials \cite{CowlingECC2007} or splines \cite{RichterSTAR2016}. One reason for using such constraints is that the resulting optimum trajectory can be feasible for real-life robots \cite{BonillaEUSIPCO21}. Various types of polynomials can be used, each with its own particular properties (e.g. Chebyishev polynomials \cite{VlassenbroeckIEEETAC1988}, Laguerre polynomials \cite{huzmezan2001multivariable}). More information about the relationship between this type of constraint and the  trajectory's feasibility and smoothness can be found in \cite{BrescianiniIEEETR2018} and the references therein.


\par
We can also constrain the initial and/or final positions of the robots \cite{CATP9}, \cite{mobb10} by adding the trivial constraints:
\begin{eqnarray}
    \label{eq:ini_fin}
    \mathbf{p}(0)=\mathbf{p}_{\rm ini}, & \mathbf{p}(t_f)=\mathbf{p}_{\rm fin}.
\end{eqnarray}
If the initial and final positions differ (see \cite{mobb10,HuangSPAWC2018}), we have an {\bf open trajectory}. If they are the same (\cite{OnoIEEETWC2016, CATP18}), we have a {\bf closed trajectory}. We find closed trajectories in repetitive tasks, such as a multirotor \ac{UAV} operating as a data ferry \cite{CATP18}, an \ac{MR} doing periodic data collection from sensor nodes \cite{GhaffarkhahGlobecom2012}, or a fixed-wing \ac{UAV} acting as a communications relay \cite{OnoIEEETWC2016,BonillaIEEECL2023}. 
Further, a predetermined trajectory that is periodic and smooth can be well approximated by truncated Fourier series.  


\par
Another type of constraint aims at limiting the velocity and/or acceleration of the robot. In \cite{CATP10}, the authors consider a group of $N$ \acp{MR} 
and limit the speed of each one using the following constraint:
\begin{equation}
\label{TPA:eq:2b}
\|\dot{\mathbf{p}}_n\|_{\infty}\leq V_n,
\end{equation}
where $\dot{\mathbf{p}}_n\in\mathbb{R}^2$ is the velocity of the $n$th \ac{MR}, and $V_n$ is a design parameter. The authors mention that they used the $H_{\infty}$-norm to limit the \acp{MR} speed instead of the $\ell_2$-norm in order to keep their particular optimization problem as a linear program. Nevertheless, the use of the $\ell_2$-norm to directly limit the speed is very common \cite{HuangSPAWC2018, Wu2018TWC}. In the case of multirotor \acp{UAV}, it is also common to separately limit the horizontal speed and the vertical speed \cite{YouIEEETWC2019}, \cite{HuaIEEETC2020}. 
Sometimes, constraints limit not only the speed, but also the acceleration of the \ac{MR} \cite{UAV11}. One may also set lower and upper bounds on the velocities and accelerations  \cite{LiuIEEERAL2022}.

The previous constraints directly limit the velocity of the \ac{UAV}, but other constraints indirectly perform this by acting on the inputs to the dynamic model of the \ac{UAV}. For instance, the authors in \cite{BezerraIEEECSL2022} limit the rotational speed and acceleration of the rotors. 

\subsection{Communications constraints}
\label{sec:CommConstraints}
These constraints act on the overall communication system performance or directly on the wireless channel experienced by the \ac{MR} along the whole trajectory. When stochastic channel models are used, the constraints related to the wireless channel are probabilistic. 
For example, the authors in \cite{UAV6} consider the problem of a multirotor \ac{UAV} in a dense \ac{IoT} network that visits $M$ hovering locations where it collects data from the connected nodes. The ground-to-air channels experience Nakagami-$m$ fading. Depending on the \ac{SINR} which itself depends on the hovering locations, the probability of successfully collecting data from the covered sensors can vary. Therefore, the authors add the following probabilistic constraint to their problem:
\begin{equation}
\label{eq:bitrateConstraint1}
   \sum_{\ell=1}^M\mathbb{E}[K_{\ell}]\geq N,
\end{equation}
where $K_{\ell}$ is the number of successfully collected observations at the $\ell$th hovering location, and $N$ is the minimum desired {\bf average} number of successful observations that the \ac{UAV} must collect from the \ac{IoT} network. In \cite{YanACC2013}, the authors consider the problem of an \ac{MR} that must visit $m$ \ac{POI}, gather their data, and then transmit it to a \ac{BS}. The \ac{MR} can transmit with $n_r$ different spectral efficiencies. The \ac{MR} can choose to transmit data from the current \ac{POI} or from other $n_s$ communication points close by to compensate for small-scale fading. The authors include the following constraint:
\begin{equation}
\label{eq:bitrateConstraint2}
    \sum_{i=2}^m\sum_{j=1}^{n_s+1}\sum_{k=1}^{n_r} R_k t_{i,j,k}=V/B,
\end{equation}
where $R_k$ is the $k$th spectral efficiency, $V$ is the total number of bits collected from the \acp{POI}, $B$ is the bandwidth used in the transmission, and $t_{i,j,k}$ is the time spent by the \ac{MR} transmitting at the $j$th communication point of the $i$th \ac{POI} using the spectral efficiency $R_k$. Note that even if both \cite{UAV6} and \cite{YanACC2013} consider a stochastic channel model, the constraint (\ref{eq:bitrateConstraint1}) is stochastic but the constraint (\ref{eq:bitrateConstraint2}) is deterministic. This is due to the fact that the constraint (\ref{eq:bitrateConstraint1}) acts on the received bits, which is a stochastic process (due to the stochastic channel model), whereas the constraint (\ref{eq:bitrateConstraint2}) acts on the transmitted bits, which is a deterministic process. In \cite{UAV11}, the authors consider a multirotor \ac{UAV} acting as a \ac{BS} and providing communications to ground users. The transmitter of the \ac{UAV} has a bandwidth $B$ divided into $N_F$ orthogonal subcarriers. The authors include the following constraint to ensure that each user is provided with at least a specified minimum bit rate:
\begin{equation}
\label{eq:bitrateConstraint3}
    \sum_{i}^{N_F}\left(\frac{s_k^i[n]B}{N_F}\right)\log_2\left(1+\gamma_k^i[n]\right)\geq R_k^{req}, \,\,\forall k,n
\end{equation}
where $s_k^i[n]=1$ if the $k$th user is assigned the $i$th subcarrier by the \ac{UAV} at time $n$, and $s_k^i[n]=0$ otherwise, $B/N_F$ is the bandwidth of each subcarrier, $\gamma_k^i[n]$ is the \ac{SNR}, of the channel between the $k$th user and the \ac{UAV} on the $i$th subcarrier at instant $n$, and $R_k^{req}$ is the minimum bit rate required by the $k$th user. The constraint (\ref{eq:bitrateConstraint3}) ensures that, at all times, the achievable data rate of each user is at least $R_k^{req}$. In \cite{UAV11}, the optimization problem was formulated in discrete time. We can find similar constraints in \cite{XiaoIEEEIoTJ2020} where a \ac{UAV} serves $K$ users and a minimum bit rate for each one must be ensured. Similarly, in \cite{ChenICC2017}, a \ac{UAV} acts as a communications relay between a ground user and a \ac{BS} and a constraint is added to ensure a minimum bit rate on each communications link.

\par
We can also constrain the average number of bits that the robot must transmit to have a lower bound:
\begin{equation}
\label{TPA:eq:3}
\int_{0}^T\mathbb{E}[R({\rm SNR}(\mathbf{p}(t),\mathbf{q}(t),t))]\mathrm{d}t\geq N,
\end{equation}
where ${\rm SNR}(\mathbf{p}(t),\mathbf{q}(t),t)$ is the instantaneous \ac{SNR} at time $t$ over the link between the \ac{MR} located at $\mathbf{p}(t)$ and the other node located at $\mathbf{q}(t)$. The function $R(\cdot)$ determines the transmission bit rate as a function of the instantaneous SNR. {
This function $R(\cdot)$ depends on the modulation scheme as well as on the bandwidth of the transmitted signal. The expected value in (\ref{TPA:eq:3}) is taken w.r.t. the joint \ac{PDF} of the shadowing and small-scale fading}. To introduce this constraint, the designer needs to know the first-order statistics of the wireless channel model. Because of the stochastic nature of the wireless channel, the constraint (\ref{TPA:eq:3}) does not ensure that the \ac{MR} will be able to always transmit at least $N$ bits, but rather that following the same trajectory in the same environment it will transmit at least $N$ bits on average. Note that (\ref{TPA:eq:3}) assumes that the transmission process is continuous 
(the integrand in (\ref{TPA:eq:3}) is treated as a continuous function of time). Even if the transmission is not continuous (e.g. when data is transmitted in packets and not in a continuous and uninterrupted flow), we can assume in practice that it is a continuous process since the dynamics of the motion is generally much slower than those of data  transmission.
\par
When formulating the problem in discrete time and dealing with \ac{DF} relays, there is a common constraint called \ac{ICC} \cite{comm5, LiuICCC2019,Ahmed2020CL,HuIEEETWC2019}: at time slot $n$, the \ac{DF} relay receives data from a source, processes it, and then transmits it at the next time slot to the destination. 
For instance, in \cite{comm5}, the \ac{ICC} is included in the discrete-time optimization problem as follows:
\begin{equation}
\label{eq:bitrateConstraint4}
    \sum_{i=2}^n\log_2\left(1+p_r[i]\gamma_{rd}[i]\right)\leq \sum_{i=1}^{n-1}\log_2\left(1+p_s[i]\gamma_{sr}[i]\right),
\end{equation}
where $\gamma_{sr}[i]$ and $\gamma_{rd}[i]$ are the \ac{CNR} at discrete time slot $i$ of the source-to-relay and relay-to-destination links, respectively, and $p_s[i]$ and $p_r[i]$ are the transmission powers at discrete time slot $i$ of the source and of the relay, respectively. The argument of the left term represents the achievable bit rate in the relay-to-destination link at slot $i$, while the argument of the right term represents the achievable bit rate in the source-to-relay at slot $i$. The \ac{ICC} (\ref{eq:bitrateConstraint4}) ensures that at time slot $n$, the relay cannot transmit to the destination more bits than the maximum number of bits it had already processed from the source, hence the term {\it causality}. We must mention that the \ac{AF} relays operate differently from the \ac{DF} relays, and thus the \ac{ICC} should be adapted accordingly.

\par
There are also constraints acting directly on the wireless channel; the advantage of such constraints resides in the fact that it makes the optimisation problem independent of the used communication system (e.g. modulation type, \ac{AF} or \ac{DF} relay). For instance, when the channel model is stochastic, we cannot ensure that the \ac{CNR} will take a certain value $\gamma_0$, but we can add a constraint to ensure that it achieves this value $\gamma_0$ with a desired probability $\eta$:
\begin{equation}
\label{TPA:eq:4}
\mathrm{Pr}\left({\rm CNR}(\mathbf{p}(t),\mathbf{q}(t),t)\geq \gamma_0\right)=\eta,\ \forall\ t\in[0,T],
\end{equation}
where $\mathbf{p}(t)$ and $\mathbf{q}(t)$ are the positions of the transmitter and receiver at time $t$, and $T$ is the duration over which the constraint must be enforced. We find a similar constraint in \cite{KalogeriasICASSP2016,MozaffariIEEETC2017} where the authors 
use expected values rather than probabilities to deal with the stochasticity of the communications channels. In \cite{UAV19}, the authors introduce a constraint limiting the maximum path loss allowed, which acts only on the deterministic part of the wireless channel.
\par
Another class of constraints act on the communications coverage. They can act either in space \cite{UAV17,UAV18} or in time \cite{BulutIEEEICC2018}. The authors in \cite{UAV18} consider a \ac{UAV} acting as an aerial \ac{BS} for a group of ground users $\mathcal{U}$ and aim to maximize the number of users served. To solve this problem, the authors use the following constraint:
\begin{equation}
\label{eq:coverage1}
    R(h)\geq r_i-M(u_i-1),\,\, \forall\, \, i=1,\cdots,|\mathcal{U}|,
\end{equation}
where $R(h)$ is the communications coverage radius of the \ac{UAV}, which depends on its altitude $h$, $r_i$ is the horizontal distance between the $i$th user and the \ac{BS}, $\{u_i\}_{i=1}^{|\mathcal{U}|}$ are binary allocation variables to be optimized (if $u_i=1$, then the $i$th user is served by the \ac{UAV} and $u_i=0$ otherwise), and $M>\displaystyle\max_i\{ r_i\}$ is a design parameter. The constraint (\ref{eq:coverage1}) works as follows. If we allocate the $i$th user to the aerial \ac{BS} (i.e. $u_i=1$), then (\ref{eq:coverage1}) is satisfied only if that user is within the coverage area. On the other hand, if we do not allocate the $i$th user to the aerial \ac{BS} (i.e. $u_i=0$), then the constraint (\ref{eq:coverage1}) becomes satisfied regardless of its horizontal distance $r_i$ to the aerial \ac{BS}. In \cite{BulutIEEEICC2018}, the authors consider a \ac{UAV} navigating in an area sparsely covered by ground \acp{BS}. The objective is to reach a final destination in minimum time $T$, but to ensure that the \ac{UAV} does not spend more than $\tau_{\rm max}$ continuous seconds outside the coverage area the following constraint is used:
\begin{equation}
\label{eq:coverage2}
    \rho(t)-t-1\leq \tau_{\max},
\end{equation}
where $\rho(t)=\argmin_{\forall s>t} \{c_s=1\}$ with:
\begin{equation}
\label{eq:coverage2b}
c_t=\Bigg\{
\begin{array}{cc}
1     & \mathrm{if}\,\exists\, \mathbf{g}_k\in\mathcal{G}\,: \|\mathbf{g}_k-\mathbf{p}(t)\| \leq R,\\
0     & \mathrm{otherwise},
\end{array}
\end{equation}
where 
$\mathcal{G}$ is set containing the positions of all the \acp{BS}, $\mathbf{g}_k$ is the position of the $k$th \ac{BS}, and $R$ is the coverage range of the \acp{BS}.
\par
Other constraints act on the transmission power. For example, the authors in \cite{HuangSPAWC2018} consider a \ac{UAV} equipped with a transmitter that has power control capacity 
and constrain the average transmission power:
\begin{equation}
\label{eq:powerConstraint1}
    \frac{1}{N}\sum_{n=1}^Np[n]\leq P,
\end{equation}
where $p[n]$ is the power transmitted by the \ac{UAV} at discrete time $n$, $N$ is the duration of the \ac{UAV} trajectory, and $P$ is the maximum average power allowed. 
In \cite{UAV11}, the authors consider a \ac{UAV} with a transmitter with $N_F$ subcarriers to serve $K$ users. The authors constrain the maximum instantaneous transmission power as:
\begin{equation}
\label{eq:powerConstraint2}
    \sum_{i=1}^{N_F}\sum_{k=1}^K s_k^i[n]p_k^i[n]\leq P_{\max},
\end{equation}
where $s_k^i[n]$ is a binary allocation variable. If at time $n$, the $i$th subcarrier is allocated to the $k$th user, then $s_k^i[n]=1$; otherwise, $s_k^i[n]=0$. The power transmitted by the \ac{UAV}  to the $k$th user through the $i$th subcarrier is $p_k^i[n]$. The overall sum in the left term of (\ref{eq:powerConstraint2}) is the total power transmitted by the \ac{UAV} at time $n$. Both constraints (\ref{eq:powerConstraint1}) and (\ref{eq:powerConstraint2}) make sense for diverse reasons, among them we find the properties of the power amplifier of the \ac{RF} transmitter. Indeed, the \ac{RF} amplifier is a nonlinear device. Hence, while it is possible that the average power of the signal transmitted operates within the linear region of the amplifier, if the maximum instantaneous power is unconstrained, then the peak power may drive the \ac{RF} amplifier to operate in the nonlinear region and thus lead to distortion. This will degrade the performance of the communication \cite{LeiIEETC2004,MorganIEEETSP2006}.
\par
The energy consumed in communications can sometimes be comparable to the energy spent in motion. This is reflected in \cite{UAV11}, where the authors consider a \ac{UAV} equipped with solar panels for energy harvesting and providing communication to ground users. To ensure the continuous operation of the \ac{UAV}, the authors add the following constraint:
\begin{equation}
    \label{eq:totalConsumption}
    \left(\frac{1}{\epsilon}P_{\rm tx}[n]+P_{\rm UAV}[n]+P_{\rm static}\right)\Delta_T\leq q[n],\, \forall n,
\end{equation}
where $P_{\rm tx}[n]$ is the power transmitted at discrete time $n$, $P_{\rm UAV}[n]$ is the aerodynamic power consumption of the \ac{UAV}, $P_{\rm static}$ is the static power required to maintain the operation of the \ac{UAV}, $\Delta_T$ is the sampling duration of the \ac{UAV} controller, and $q[n]$ is the energy level of the battery. The energy harvested from the sun does not appear in (\ref{eq:totalConsumption}) because in this system the solar panels are only connected to the battery, so they cannot directly power the other systems of the \ac{UAV}. \qedsymbol{}

\subsection{Soft Constraints}
\label{sec:SoftConstraints}
The constraints discussed above are {\bf hard constraints}, i.e., either they are satisfied or they are not. In some cases, hard constraints can cause feasibility issues in the optimization problem \cite{AskariIEEETIA2017}. It might be that some combination of parameter values makes it impossible to satisfy every single hard constraint in the optimization problem, rendering it unfeasible. In complex \ac{CaTP} problems 
it might be hard to identify this situation. 
This can be addressed by relaxing some hard constraints and transforming them into {\it soft constraints} \cite{SchepkerACM2020,KerriganUKACC2000,ChenICISCE2016}. The idea of soft constraints is to allow violations but impose a penalty depending on the extent of the violation. 

In \ac{CaTP}, some constraints are imposed by the physical world. For example, the energy constraint (\ref{eq:totalConsumption}) states that the \ac{UAV} cannot consume more energy than the energy stored within its battery, as this constraint is imposed by the laws of physics and cannot be relaxed. On the other hand, other constraints are arbitrarily imposed by the designer. For example, for the constraint in (\ref{eq:ini_fin}), the final position of the robot is arbitrarily imposed by the designer. Thus, if the robot finishes its trajectory at $\mathbf{p}_{fin}+\boldsymbol{\epsilon}$ with a small $\|\boldsymbol{\epsilon}\|_2$, rather than at exactly  $\mathbf{p}_{fin}$, there would probably not be a problem. Similarly, in the constraint (\ref{eq:bitrateConstraint3}), the minimum achievable bit rate $R^{req}_k$ is arbitrarily imposed by the designer. If a user experiences a bit rate slightly inferior to the desired minimum bit rate, it might pass unnoticed. These constraints are the ones that can be relaxed and converted into soft constraints.
\par
There are various methods to implement soft constraints, but we will discuss only three of them. The first method consists of the use of {\it slack variables} \cite{KerriganUKACC2000,AskariIEEETIA2017,boyd_vandenberghe_2004}. We can use them to relax inequality constraints. For instance, consider the following generic optimization problem with hard constraints:
\begin{eqnarray}
\label{eq:hardConstraint}
    \minimize_x&J(x)&\nonumber\\
    \mathrm{s.t.}&&\nonumber\\
    &g(x)\leq 0,\quad f(x)=c,&
\end{eqnarray}
where $x,J(x),g(x),f(x),c$$\in$$\mathbb{R}$. We can relax the hard inequality constraint imposed on $g(x)$ with a slack variable $s$:
\begin{eqnarray}
\label{eq:softConstraint1}
    \minimize_{x,s}&J(x)+k_1s^2&\nonumber\\
    \mathrm{s.t.}&&\nonumber\\
    &g(x)\leq s,\quad s\geq 0,\quad f(x)=c,&
\end{eqnarray}
where $k_1>0$ is a design parameter. In (\ref{eq:softConstraint1}), we can have $g(x)>0$ since the slack variable $s$ will take the necessary value to satisfy the constraint $g(x)\leq s$. Nevertheless, this adds a penalty $k_1s^2$ to the cost function. If $g(x)\leq0$, then $s=0$. The relaxed optimization problem needs to optimize not only the variable $x$ but also $s$. In \cite{mobb10}, we find an interesting example where a \ac{CaTP} has its cost function exclusively composed of a single slack variable.
\par
The second method consists of the use of exponential penalty functions \cite{DorMohammadiSMO2013} to relax the inequality functions:
\begin{eqnarray}
\label{eq:softConstraint2}
    \minimize_{x}&J(x)+\frac{k_1}{k_2}\left(\exp{\left(k_2\,g(x)\right)}-1\right)&\nonumber\\
    \mathrm{s.t.}&&\nonumber\\
    &f(x)=c,&    
\end{eqnarray}
where $k_1,k_2>0$. If $g(x)=0$, the penalization term becomes exactly zero. If $g(x)<0$ and $k_2\rightarrow +\infty$, then the penalization term tends to zero and its effect vanishes. If $g(x)>0$, then the penalization term becomes larger. In \cite{CATP19}, we used exponential penalization to avoid obstacles in a \ac{CaTP} problem.
\par
The third method is to transfer the equality constraints to the optimization target as quadratic penalizations \cite{ChenICISCE2016,SchepkerACM2020}: 
\begin{eqnarray}
\label{eq:softConstraint3}
    \minimize_{x}&J(x)+k_1|f(x)-c|^2&\nonumber\\
    \mathrm{s.t.}&&\nonumber\\
    &g(x)\leq 0,&
\end{eqnarray}
where $k_1>0$. The penalization term in (\ref{eq:softConstraint3}) is small if $f(x)$ is close to the desired value $c$, and the farther $f(x)$ from $c$, the larger the penalization. In \cite{AliIEEETCNS2019}, the authors consider a \ac{CaTP} with multiple \acp{MR} and they add to their cost functional the same type of penalization as in (\ref{eq:softConstraint3}). 
\par
We can simultaneously relax the inequality and equality constraints by combining the use of slack variables and the quadratic penalization terms: 
\begin{eqnarray}
\label{eq:softConstraint4}
    \minimize_{x,s}&J(x)+k_1|f(x)-c|^2+k_2s^2&\nonumber\\
    \mathrm{s.t.}&&\nonumber\\
    &g(x)\leq s,\quad s\geq 0,& 
\end{eqnarray}
\par
If the original problem (\ref{eq:hardConstraint}) is feasible, then all of its hard constraints are satisfied by the optimum solution $x^*$. Consequently, the relaxed problems (\ref{eq:softConstraint1}), (\ref{eq:softConstraint3}), and (\ref{eq:softConstraint4}) are equivalent to (\ref{eq:hardConstraint}), and (\ref{eq:softConstraint2}) will tend to be equivalent to (\ref{eq:hardConstraint}) as $k_2\rightarrow +\infty$. Alternatively, if the original problem (\ref{eq:hardConstraint}) is unfeasible, the relaxed problem (\ref{eq:softConstraint4}) is still feasible; by tuning $k_1$ and $k_2$, we can get as close as possible to satisfying the original hard constraints.

\par
\section{\ac{CaTP} Examples}
\label{sec:ExamplesDetails}
Next, we first describe in detail three illustrative examples of \ac{CaTP} problems that were carefully selected to showcase the variety of mathematical forms that \ac{CaTP} problems can take. Then, we present a survey of {\it minimum energy} \ac{CaTP} problems,
a survey of {\it energy efficient}  \ac{CaTP} problems,
and finally a survey of the different types of \ac{CaTP} problems.

\begin{example}
\label{example1}
An \ac{MR} is tasked to move in a finite time $t_f$ from an initial position $\mathbf{p}_{ini}$ to a final position $\mathbf{p}_{fin}$ while transmitting a given number of bits $c$ to a \ac{BS} with a certain bit error probability. The objective is to optimize the trajectory and the instantaneous bit rate to balance the motion energy and the energy spent in transmission. 
\par
The input to the \ac{MR} is the control signal $\mathbf{u}(t)\in\mathbb{R}^2$, which directly controls its acceleration. The communications input is the instantaneous transmission bit rate $R(t)\in\mathbb{R}$. Note that $\mathbf{u}(t)$ and $R(t)$ are both functions with support $[0,t_f]$.
\par
In this example, the state vector of the system is $\mathbf{x}^{\mathrm{T}}=[\mathbf{x}_{1}^{\mathrm{T}}(t)\,\,\mathbf{x}_{2}^{\mathrm{T}}(t)\,\, {x}_3(t)]$, where $\mathbf{x}_{1}(t)\in\mathbb{R}^2$ and $\mathbf{x}_{2}(t)\in\mathbb{R}^2$ are the \ac{MR} position and velocity, and ${x}_3(t)\in\mathbb{R}$ is the number of bits transmitted by the \ac{MR}.
\par
The optimization problem is cast as a minimization problem with the following cost functional:
\begin{eqnarray}
\label{example1:costFunctional}
 J(\mathbf{x},\mathbf{u},R) &=& \int_{0}^{t_f}\left(\frac{(2^{R(t)}-1)s(\mathbf{x}_1(t))}{K}\right)\mathrm{d}t\nonumber\\
   &+& \gamma\int_{0}^{t_f}p_2\left(\|\mathbf{x}_2(t)\|_2,\|\mathbf{u}(t)\|_2\right)\mathrm{d}t\nonumber\\
   &+&C_1\|\mathbf{x}_1(t_f)-\mathbf{p}_{\rm fin}\|^2+C_2\|\mathbf{x}_2(t_f)\|^2\nonumber\\
   &+&C_3\|{x}_3(t_f)-c\|^2.
\end{eqnarray}
 The first term in the functional corresponds to the energy spent in transmission, where the transmitter of the \ac{MR} has a power control mechanism that sets $K=-1.5/\log(5p_b)$ to adapt the transmission power to ensure a \ac{BER} $p_b$. The integrand of this first term corresponds to the power required to meet the targeted \ac{BER} with the transmission bit rate $R(t)$, as well as the \ac{CNR} estimation $s(\mathbf{x}_1(t))$ at position $\mathbf{x}_1(t)$. The second term is the motion-induced energy, weighted by parameter $\gamma>0$, with $p_2(a,b)$ being a second-order polynomial in $a$ and $b$; the integrand is the instantaneous power consumed in motion. The last terms are quadratic penalties on the desired final states of the system. $C_1,C_2,C_3>0$ are design parameters that determine the importance of reaching the desired final states. The first penalty moves the final \ac{MR} position $\mathbf{x}_1(t_f)$ to the desired final position $\mathbf{p}_{\rm fin}$. The second penalty makes the \ac{MR} to end the trajectory with the lowest possible speed. The third penalty drives the \ac{MR} to transmit data close to $c$ bits by the end of the trajectory. The initial state value of the state variables is set with the following constraints:
 \begin{eqnarray}
 \label{example1:initial}
     \mathbf{x}_1(0)=\mathbf{p}_{\rm ini},&  \mathbf{x}_2(0)=\boldsymbol{0}, & x_3(0)=0,
 \end{eqnarray}
 and the inputs are bounded according to:
 \begin{eqnarray}
  \label{example1:boundedInputs}
  0\leq \|\mathbf{u}\|_2\leq u_{\max},& 0\leq R(t)\leq R_{\max}.
 \end{eqnarray}
 \par
 The system is described by the following linear dynamic model:
 \begin{eqnarray}
 \label{example1:modela}
     \dot{\mathbf{x}}_1(t)=\mathbf{x}_2(t),&\quad& 
     \dot{\mathbf{x}}_2(t)=\mathbf{u}(t),\\
 \label{example1:modelc}
     \dot{x}_3(t)&=&R(t),
 \end{eqnarray}
The robotics-related aspects are described in (\ref{example1:modela}). From a systems theory perspective, one may state that (\ref{example1:modelc}) alone represents the  communications model. However, if we interpret the communications model as the full mathematical description of how the communications system behaves, then it includes not only (\ref{example1:modelc}) but also the bounds on the bit rate (\ref{example1:modelc}) and the transmitted power description in the first term of the functional (\ref{example1:costFunctional}), as well as the function $s(\mathbf{x}_1(t))$ describing how the \ac{CNR} estimation varies with the position of the \ac{MR}. In summary, this \ac{CaTP} problem takes the following form:
 \begin{eqnarray}
     \minimize_{\mathbf{u},R} & &J(\mathbf{x},\mathbf{u},R)\nonumber\\
     \mathrm{s.t.} & & \nonumber\\
    & & \dot{\mathbf{x}}_1(t)=\mathbf{x}_2(t),\quad \dot{\mathbf{x}}_2(t)=\mathbf{u}(t),\quad \dot{x}_3(t)=R(t),\nonumber\\    
    & &\mathbf{x}_1(0)=\mathbf{p}_{\rm ini},\,\,  \mathbf{x}_2(0)=\boldsymbol{0},\,\,  x_3(0)=0,\nonumber\\
&&    0\leq \|\mathbf{u}\|_2\leq u_{\max},\,\, 0\leq R(t)\leq R_{\max}.\nonumber\\
 \end{eqnarray}
 It is worth noting that the cost functional of this optimization problem, with the exception of the first term, is quadratic w.r.t. the state vector $\mathbf{x}$ and the input $R$. 
\end{example}
\par



\begin{example}
    \label{example3}
An \ac{IoT} sensor network is randomly deployed in a region, say $\mathcal{A}$, 
and a multirotor \ac{UAV} is tasked with collecting at least an expected amount of $\zeta$ successful transmissions from the network. The \ac{UAV} is equipped with an antenna of a fixed beamwidth $\phi$. It visits $M$ \acp{HL} to collect data from the sensor nodes. The coordinates of the $m$th {\ac{HL}} are $\ell_{m}=[x_{m}\,y_{m}\,h]$, where $x_{m}$ and $y_{m}$ are the $x-y$ coordinates of  the $m$th {\ac{HL}} and $h$ is its altitude, which is the same for all of the \acp{HL}. When the \ac{UAV} is at the $m$th {\ac{HL}}, its communication system covers a circular area $\mathcal{A}_{m}$ of radius:
\begin{equation}
\label{example3:CoverageRadius}
    R=h\tan(\phi/2).
\end{equation}
At the $m$th \ac{HL}, the \ac{UAV} spends a time $T_{hover}^{m}$ performing the following process. It first broadcasts a signal to activate the nodes in $\mathcal{A}_{m}$. The nodes that get activated transmit, with a constant spectrum efficiency $\log_2(1+\beta)$, over a common radio spectrum using a slotted ALOHA medium access scheme with transmission probability $p_{tx}$. The transmission of the $n$th node is successful if the \ac{SINR} at the \ac{UAV} is larger than $\beta$. This criterion comes from the definition of Shannon's channel capacity and its implications; if the \ac{SINR} of the communications channel is $\beta_0$, then the maximum bit rate ensuring that the data can be recovered without errors is $B\log_2(1+\beta_0)$, where $B$ is the bandwidth. Thus, if we try to transmit over this channel using a bit rate of $B\log_2(1+\beta)$ with $\beta_0<\beta$, the transmission will fail. The set of nodes in the region $\mathcal{A}_{m}$ that successfully transmit their data is denoted as $\tilde{\boldsymbol{\Psi}}_{m}$.  
\par
The transmission of the node with the largest \ac{SINR} is treated as the intended transmission by the \ac{UAV}; the transmissions of other nodes are treated as interference. The \ac{SINR} at the \ac{UAV} receiver related to the transmission from a node $n$ in the set $\tilde{\boldsymbol{\Psi}}_{m}$ is:
\begin{eqnarray}
\label{example3:SINR}
    \mathrm{SINR}_{m}^n=\frac{PG_nD^{-\eta}_n}{\sum_{\mathbf{x\in\tilde{\boldsymbol{\Psi}}_{m}}/\mathbf{x}_n} PG_xD^{-\eta}_x+\sigma^2},
\end{eqnarray}
where $P$ is the transmission power, $\eta$ is the power path-loss coefficient, $D_n$ is the distance between the \ac{UAV} and the $n$th node in $\tilde{\boldsymbol{\Psi}}_{m}$, $G_n$ is the power gain due to small-scale fading (considered here to follow a Nakagami-$m$ distribution), and $\sigma^2$ is the noise power at the receiver. 
\par
When the \ac{UAV} moves from one \ac{HL} to another \ac{HL}, it moves on a straight line. Its acceleration rate, its deceleration rate and its maximum speed are denoted by  $\hat{a}$,  $\Check{a}$ and $v_{max}$, respectively. Thus, the time that the \ac{UAV} takes to travel from the $m$th \ac{HL} to the $(m+1)$th \ac{HL} is:
\begin{equation}
\label{example3:travelTime}
    \tau_{\mu}=\Bigg\{
    \begin{array}{ccc}
      \sqrt{\frac{d_{m}}{\hat{{a}}+\Check{{a}}}}  & \mathrm{if} & d_{m}\leq \hat{d}+\Check{d},  \\
       \hat{t}+\Check{t}+\frac{d_{m}-\hat{d}-\Check{d}}{v_{\max}}  & \mathrm{if} & d_{m}> \hat{d}+\Check{d},
    \end{array}
\end{equation}
where $d_{m}$ is the distance between the $m$th \ac{HL} and the $(m+1)$th \ac{HL}, $\hat{t}=v_{max}/\hat{a}$,  $\Check{t}=v_{max}/\Check{a}$, $\hat{d}=1/2\hat{a}\hat{t}^2$, and $\Check{d}=1/2\Check{a}\Check{t}^2$. The first line in (\ref{example3:travelTime}) is the time it takes for the \ac{UAV} to move between the two \acp{HL}. If $d_{m}$ is too short, then the \ac{UAV} cannot reach its maximum speed. The second line is the time it takes when the distance is large enough so that the \ac{UAV} reaches its maximum speed.
\par
The trajectory-related inputs in the optimization problem are the total number of \acp{HL}, say $M$, and their 3D locations, given by $\mathbf{L}=[\boldsymbol{\ell}_1,\,\boldsymbol{\ell}_2,\cdots,\boldsymbol{\ell}_{M}]$ with $\boldsymbol{\ell}_{m}\in\mathbb{R}^3$. The flight path of the \ac{UAV} is encoded in the binary matrix $\mathbf{Z}\in\{0,1\}^{M\times M}$ where $z_{i,j}=1$ if the \ac{UAV} departs from the $i$th \ac{HL} to arrive to the $j$th \ac{HL}, and  $z_{i,j}=0$ otherwise.
\par
The communications-related input is the coverage radius $R$. However, given that in this scenario the beamwidth $\phi$ of the \ac{UAV} antenna is fixed, the altitude of the \acp{HL} uniquely determines the coverage radius $R$, see (\ref{example3:CoverageRadius}). Thus, the inputs $R$ and $\mathbf{L}$ cannot be chosen independently. 
\par
The path constraints of the optimization problem are as follows. Each \ac{HL} must be reached from only one \ac{HL}:
\begin{equation}
    \label{example3:pathConstraint1}
    \sum_{i=1,i\neq j}^Mz_{i,j}=1,\,\,\,\forall\, j\in [1,M].
\end{equation}
At each \ac{HL}, the \ac{UAV} can only reach one \ac{HL}:
\begin{equation}
    \label{example3:pathConstraint2}
    \sum_{j=1,i\neq j}^Mz_{i,j}=1,\,\,\,\forall\, i\in [1,M].
\end{equation}
Note that constraints (\ref{example3:pathConstraint1}) and (\ref{example3:pathConstraint2}) imply that the path for the \ac{UAV} must be closed. The last path constraint is the \ac{MTZ} constraint \cite{MillerJACM1960} which ensures that the path does not include subloops :
\begin{eqnarray}
\label{example3:pathConstraint3}
    \phi_i-\phi_j+Mz_{i,j}\leq M-1,\quad
    2\leq i\neq j\leq M.
\end{eqnarray}
where $\boldsymbol{\phi}=[\phi_2,\,\phi_3,\cdots,\phi_M]\in\mathbb{R}^{M-1}$ are auxiliary variables for the \ac{MTZ} constraint exclusively.
\par
The communications constraints imposed are as follows. The expected value of the number of successful transmissions collected at all of the $M$ \acp{HL} by the \ac{UAV} must be at least $\zeta$:
\begin{equation}
\label{example3:DataConstraint}
    \sum_{m=1}^M\mathbb{E}[K_{m}]\geq \zeta,
\end{equation}
where $K_{m}$ is the number of successful transmissions collected at the $m$th \ac{HL}. The \ac{UAV} must cover the full region $\mathcal{A}$ where the \ac{IoT} sensor network is deployed:
\begin{equation}
    \label{example3:CoverageConstraint}
    \mathcal{A}\subseteq \displaystyle\cup_{m=1}^M\mathcal{A}_{m}.
\end{equation}
The cost function is the total flying time of the \ac{UAV} to perform one full round of the collection process:
\begin{equation}
    T_{total}=\sum_{m=1}^MT_{\rm hover}^{m}+T_{\rm travel},
\end{equation}
where $T_{\rm hover}^{m}$ is the time that the \ac{UAV} hovers over the $m$th \ac{HL}, which depends on the number of \acp{HL}, the channel capacity, and the probability of receiving a successful transmission, see \cite{BushnaqIEEETWC2019} for the details; $T_{\rm travel}$ is the time that the \ac{UAV} spends in motion which depends mainly on $\mathbf{L}$ and on $\mathbf{Z}$. In summary, this \ac{CaTP} problem takes the following form:
\begin{eqnarray}
     \minimize_{M,R,\mathbf{L},\mathbf{Z},\boldsymbol{\phi}} & T_{\rm total}&\nonumber\\
     \mathrm{s.t.} & & \nonumber\\
     \sum_{m=1}^M\mathbb{E}[K_{m}]\geq \zeta &\mathcal{A}\subseteq \displaystyle\cup_{m=1}^M \mathcal{A}_{m},&\nonumber\\    
     (101),\quad (102),&\nonumber\\ 
    \phi_i-\phi_j+Mz_{i,j}\leq M-1,&\,\,   2\leq i\neq j\leq M,& \nonumber\\
     R=h\tan(\phi/2).& &
 \end{eqnarray}
 Note that this optimization problem is a \ac{MIP} problem since it has continuous variables ($\mathbf{L}$, $\boldsymbol{\phi}$ and $R$) and discrete variables ($M$ and $\mathbf{Z}$); in this example, the discrete variables determine the \ac{UAV} path.
\end{example}

\begin{example}
\label{example4}
In \cite{UAV11}, a multirotor \ac{UAV} equipped with solar panels to recharge its battery and with a multicarrier transmitter having $N_F$ subcarriers, is tasked with providing communications to $K$ ground users. This problem is formulated in discrete time and considers the joint optimization of the 3D \ac{UAV} trajectory and the wireless resource allocation to maximize the system's sum throughput over a period of time. The time is divided into $N_T$ time slots of duration $\Delta_T$. The inputs to the \ac{UAV} is its 3D position $\mathbf{r}\in\mathbb{R}^{N_T\times3}$ and its 3D velocity $\mathbf{v}\in\mathbb{R}^{N_T\times3}$. Another input in this problem is the energy level of the \ac{UAV} battery $\mathbf{q}\in\mathbb{R}^{N_T\times1}$.  The communications-related inputs are the power transmitted by the \ac{UAV} and the subcarriers allocation schedule. The power transmitted to the $k$th user through the $i$th subcarrier at time slot $n$ is denoted by $p_k^i[n]$; if the $i$th subcarrier is allocated to the $k$th user at time slot $n$, then $s_k^i[n]=1$, and $s_k^i[n]=0$ otherwise. We also define $\mathbf{p}\in\mathbb{R}_{>0}^{N_T\times N_F\times K}$ and $\mathbf{p}\in\{0,1\}^{N_T\times N_F\times K}$ which contain all the set of variables $p_k^i[n]$ and $s_k^i[n]$ respectively.
\par
The \ac{UAV} motion model is a first-order model: 
\begin{equation}
    \mathbf{r}[n+1]=\mathbf{r}[n]+\mathbf{v}[n+1]\Delta_T,
\end{equation} 
where $\mathbf{r}[n]\in\mathbb{R}^3$ is the position of the \ac{UAV} at time slot $n$, and $\mathbf{v}[n]\in\mathbb{R}^3$ is the velocity of the \ac{UAV}. The \ac{UAV} model also includes the aerodynamic power consumption model: 
\begin{eqnarray}
    P_{\rm UAV}[n]&=&\frac{k_1}{\sqrt{v_{x,y}^2[n]+\sqrt{v_{x,y}^4[n]+4k_2^4}}}\nonumber\\
&+&k_3v_z[n]+k_4v_{x,y}^3[n],
\end{eqnarray}
where $v_{x,y}[n]$ and $v_z[n]$ are the horizontal and vertical speeds of the \ac{UAV}, where $k_1$, $k_2$, $k_3$, and $k_4$ are constants; see \cite{UAV11} for more details. The power generated by the solar panel is upper bounded according to:
\begin{equation}
    \underline{P}_{\rm solar}[n]=\frac{c_1}{1+\exp{-c_2(z[n]-c_3)}}+c_4,
\end{equation}
\par The communications model is  described by the following equations. The signal received by user $k$ assigned to the $i$th subcarrier at time slot $n$ is:
\begin{equation}\label{example4:channelModel}
    u_k^i[n]=\frac{\sqrt{\zeta p_k^i[n]}h_k^i[n]}{\|\mathbf{r}[n]-\mathbf{r}_k\|}d_k^i[n]+n_k^i[n],
\end{equation}
where $\mathbf{r}_k\in\mathbb{R}^3$ is the position of user $k$, $\zeta$ captures the antenna gain,  $h_k^i[n]\in\mathbb{C}$ is the channel gain between the \ac{UAV} and user $k$ on subcarrier $i$ and it accounts for the multipath fading and the shadowing, $n_k^i[n]\in\mathbb{C}$ is the noise generated at user $k$'s receiver on subcarrier $i$, and $d_k^i[n]\in\mathbb{C}$ denotes the symbol transmitted from the \ac{UAV} to user $k$ through subcarrier $i$th. Since $d_k^i[n]$ and $n_k^i[n]$ are stochastic processes, (\ref{example4:channelModel}) would be an incomplete channel model without including their statistical distributions. In \cite{UAV11}, the authors assume that  $\mathbb{E}[d_k^i[n]]=1$ and that $n_k^i[n]$ is a zero-mean \ac{AWGN} with variance $\sigma^2$. They describe an offline trajectory planning problem and its online version. Here, we discuss the offline version only, where a non-causal knowledge  of $h_k^i[n]$ is assumed\footnote{In the online version of the problem, which is more practical, only causal knowledge can be leveraged to design the trajectory.}, thus making this term deterministic; see \cite{UAV11} for the details of this issue.  
\par
The communications model is completed with a formula for the user $k$'s achievable bit rate through subcarrier $i$:
\begin{equation}
    R_k^i[n]=s_k^i[n]B\log_2\left(1+\frac{\zeta^2|h_k^i[n]|^2p_k^i[n]}{\|\mathbf{r}[n]-\mathbf{r}_k\|^2\sigma^2}\right),
\end{equation}
This optimization problem is cast as a maximization problem, and the utility function is the total throughput averaged over the number of subcarriers, $N_F$: 
\begin{equation}
    J(\mathbf{p},\mathbf{s},\mathbf{r})=\frac{1}{N_FB}\sum_{n=1}^{N_T}\sum_{i=1}^{N_F}\sum_{k=1}^{K}R_k^i[n].
\end{equation}
The acceleration, the speed, and the altitude of the \ac{UAV} are constrained as follows:
\begin{eqnarray}
    \|\mathbf{v}[n+1]-\mathbf{v}[n]\|_2\leq a_{\max}\Delta_T,\quad\quad\quad\\
    v_{x,y}[n]\leq V^{x,y}_{\max},\quad
    v_{z}[n]\leq V^{z}_{\max},\quad Z_{\min}\leq z[n] \leq Z_{\max}.     
\end{eqnarray}
The energy level of the \ac{UAV} battery is constrained by:
\begin{eqnarray}
    0\leq q[n] \leq q_{\max},\quad
    q[1]=q_0, \,\,\, q[N_T+1]=q_{\rm end},
\end{eqnarray}
where $q_0$ and $q_{\rm end}$ are the initial and final desired states.
\par
We next describe the communications-related constraints. First, since $p_k^i[n]$ represent a power, we have that $p_k^i[n]\geq 0$.
At each time slot, the total transmit power is limited, i.e.:
\begin{equation}
    \sum_{i=1}^{N_F}\sum_{k=1}^K s_k^i[n]p_k^i[n]\leq P_{\max},
\end{equation}
where $P_{\max}$ is the maximum transmit power. The subcarriers allocation should be done so that each subcarrier is allocated to only one user at most:
\begin{equation}
    \sum_{k=1}^K s_k^i[n]\leq 1,\\
    s_k^i[n]\in\{0,1\},
\end{equation} 
Further, the \acp{UAV} must warrant each user a minimum bit rate:
\begin{equation}
    \sum_{i=1}^{N_F}R_k^i[n]\geq R_k^{\rm req},
\end{equation} 
where $R_k^{\rm req}$ is the minimum bit rate required by user $k$. 

The energy consumption-related constraints are as follows. The energy consumed cannot surpass the energy stored in the battery:
\begin{equation}
  \left[\sum_{i=1}^{N_F}\sum_{k=1}^K\frac{1}{\epsilon} s_k^i[n]p_k^i[n]+\underline{P}_{\rm UAV}[n]+P_{\rm static}\right]\Delta_T\leq  q[n],
\end{equation}
where $\epsilon\in(0,1)$ is the efficiency of the power amplifier of the \ac{UAV} transmitter, and $P_{static}$ is the constant power to keep the \ac{UAV} operational.The increase of energy in the battery between consecutive time slots is bounded according to:
\begin{eqnarray}\label{example4:energyUpperBound}
    q[n+1]&\leq&q[n]+\underline{P}_{\rm solar}[n]\Delta_T-P_{\rm static}\Delta_T\nonumber\\
    &-&\left[\sum_{i=1}^{N_F}\sum_{k=1}^K\frac{1}{\epsilon} s_k^i[n]p_k^i[n]+P_{\rm UAV}[n]\right]\Delta_T.\nonumber\\
\end{eqnarray}
Depending on the charging state, the \ac{UAV}'s battery might not store all of the power generated by its solar panel.  The resulting \ac{CaTP} problem thus takes the following form:
\begin{equation}
\maximize_{\mathbf{p},\mathbf{s},\mathbf{r},\mathbf{v},\mathbf{q}} \quad J(\mathbf{p},\mathbf{s},\mathbf{r})
\end{equation}
\begin{equation}
    \begin{array}{l}
    \mathrm{s.t.}\\
    \\
\|\mathbf{v}[n+1]-\mathbf{v}[n]\|_2\leq a_{\max}\Delta_T,\\
   v_{x,y}[n]\leq V^{x,y}_{\max},\quad v_{z}[n]\leq V^{z}_{\max},\quad
    Z_{\min}\leq z[n] \leq Z_{\max}, \\  
    \\
R_k^i[n](\mathbf{p},\mathbf{s},\mathbf{r})=s_k^i[n]B\log_2\left(1+\frac{\zeta^2|h_k^i[n]|^2p_k^i[n]}{\sigma^2}\right)\\ 
\sum_{i=1}^{N_F}\sum_{k=1}^K s_k^i[n]p_k^i[n]\leq P_{\max},\quad p_k^i[n]\geq 0,\\
    \sum_{k=1}^K s_k^i[n]\leq 1, \quad  s_k^i[n]\in\{0,1\},\\
\sum_{i=1}^{N_F}R_k^i[n](\mathbf{p},\mathbf{s},\mathbf{r})\geq R_k^{\rm req},\\
\\
{(120), \quad (121)},\\
    0\leq q[n] \leq q_{\max},\quad q[1]=q_0, \quad q[N_T+1]=q_{\rm end}.\\  
     \end{array}
\end{equation}
In this problem, there are continuous variables ($\mathbf{p}$, $\mathbf{r}$, $\mathbf{v}$ and $\mathbf{q}$) and discrete  variables ($\mathbf{s}$), i.e. this is \ac{MIP} problem. The discrete variables are used to allocate the subcarriers of the \ac{UAV} transmitter to the users served. In addition, the problem is formulated in discrete time, where the time between consecutive discrete instants corresponds to the duration of the symbols transmitted by the multircarrier transmitter of the \ac{UAV}. This assumes that the communications channel remains more or less constant during the slot time, i.e., the coherence time of the communications channel is assumed larger than the time slot duration. 
\end{example}
\par




\subsection{Minimum Energy Trajectories}
\label{sec:CATP:ME}
The {\bf Minimum Energy Trajectory with Communications Objectives} problem is one of the most illustrative \ac{CaTP} problems. It considers an \ac{MR} that accomplishes a certain task involving both motion and communications. The objective is to design the \ac{MR}'s trajectory so as to perform the desired task while minimizing the overall energy consumption (i.e. communications energy plus motion energy). This is important since \acp{MR} usually draw their energy from an onboard battery which feeds not only their locomotion system, but their communications system as well. By minimizing the overall energy consumption, we may either reduce the MR's battery size (and thus its cost) or increase the MR's operational time, allowing it to fulfil more and longer tasks. Note that this problem may be irrelevant for tethered \acp{MR} since they receive energy from an external source \cite{RossiECC2019}. 
\par
At first glance, we could say that the energy spent by the \ac{MR} in communications is negligible compared to the energy spent in motion, but as mentioned in section \ref{sec:comm:energ}, this is not always true. The energy consumed by the communication system may be comparable to the energy spent in motion \cite{CATP5}. This depends on various aspects, such as the amount of data transmitted, the distance to the node communicating with the \ac{MR}, and the adopted power transmission strategy. In addition, the size of the \ac{MR} also plays a role in this comparison \cite{UAV25}. For a small \acp{MR}, the power spent in communications can be significant w.r.t. the power consumed in motion; however, for larger and heavier robots, the opposite might be true. Mathematically, the Minimum Energy Trajectory with Communications Objectives problem is stated as an optimization problem with the following optimization target:
\begin{equation}
\label{eq:TP:en:1}
J=E_{\rm motion}(\mathbf{u},0,T)+E_{\rm comm}(\mathbf{u},0,T),
\end{equation}
where $E_{\rm motion}(\mathbf{u},0,T)$ and $E_{\rm comm}(\mathbf{u},0,T)$ are the energies spent by the \ac{MR} in motion and in communications, respectively, from time 0 up to time $T$ using the control signal $\mathbf{u}$. The power transmission policy used (see section \ref{sec:comm:energ}) drastically affects the optimum trajectory and the problem complexity.
\par
 If the transmission power is constant  and the transmission is continuous, the communication energy consumption can be expressed as:
\begin{equation}
\label{eq:TP:en:2}
E_{\rm comm}(\mathbf{u},0,T)=PT.
\end{equation}
If the final time $T$ and the transmission power $P$ are fixed, then the communications term in the cost function (\ref{eq:TP:en:1}) becomes constant and the problem becomes that of minimizing the motion energy subject to communications constraints, often related to the quality of the received signal that itself depends on the \ac{MR}'s trajectory. 
On the other hand, if a transmission power control is used (see section \ref{sec:comm:energ}), then the communication energy consumption in (\ref{eq:TP:en:1}) depends on the particular trajectory followed by the \ac{MR}. If the transmission power is given by (\ref{eq:Energy:1}), the received power is always $P_{ref}$, and thus the quality of the signal received is independent of the trajectory. However, (\ref{eq:Energy:1}) assumes no bound on the maximum transmission power. If we consider maximum transmission power, then the actual transmission power consumption is given by (\ref{eq:Energy:2}). Additional constraints are needed to ensure that the robot can communicate with a \ac{BS}. We illustrate this with the following example: an \ac{MR} must depart from a point $A$ and arrive at a point $B$ while communicating with a \ac{BS} using the power control strategy (\ref{eq:Energy:2}). We seek to optimize its trajectory so that the total energy (\ref{eq:TP:en:1}) is minimized. If the channel observed along the minimum motion energy path between points $A$ and $B$ is so poor that no transmission is possible along this path (i.e. $E_{\rm comm}(\mathbf{u},0,T)=0$), {this would be the optimum path for our problem,} however the issue is that no communication would occur. Therefore, when using the transmission power policy in (\ref{eq:Energy:2}), we must also include constraints to ensure that communication requirements are satisfied.
\par
There is an instance of this problem in \cite{CATP1} where the authors consider a ground \ac{MR} that needs to communicate with a \ac{BS} and must reach a predefined point. The path is optimized to minimize the total energy consumption while completing the entire task. 
In \cite{LiceaEUSIPCO2013,BonillaISP2013}, we consider a ground \ac{MR} that must communicate with a \ac{BS} through a communication channel experiencing small-scale fading. We optimize the \ac{MR} trajectory to find a good location for transmission while minimizing the communications and motion energies.
\par\bigskip
\subsection{Energy Efficient Trajectories}
\label{sec:CATP:EFF}
In this scenario, the goal is to use as little energy in motion as possible to complete a communications-related task.
 This results in Energy Efficient Communications-aware trajectories that can be obtained using the following optimization target:
\begin{equation}
\label{eq:TP:eff:1}
J=\theta E_{\rm motion}(\mathbf{u},0,T)+(1-\theta)f_{\rm comm}(\mathbf{u},0,T)
\end{equation}
where $E_{\rm motion}(\mathbf{u},0,T)$ is the motion energy consumed by the \ac{MR}, $f_{\rm comm}(\mathbf{u},0,T)$ is a communication metric to be minimized, and $\theta\in[0,1]$ is a design parameter that determines the importance of one objective with respect to the other. Since both terms have generally different physical units, we need to normalize them to ensure that they vary, more or less, in the same numerical range to avoid problems when solving the \ac{CaTP} optimization problem numerically. {
We used this approach in \cite{CATP19}.} Another way of formulating an energy-efficient communications-aware trajectories problem is by maximizing the following objective function:
\begin{equation}
\label{eq:TP:eff:2}
J=\frac{g_{\rm comm}(\mathbf{u},0,T)}{E_{\rm motion}(\mathbf{u},0,T)}
\end{equation}
where $g_{\rm comm}(\mathbf{u},0,T)$ is a communications metric to be maximized. Here, (\ref{eq:TP:eff:2}) maximizes the ratio of two metrics. The advantage of doing this is that we can easily combine metrics having different units and do not need to normalize them, nor to select an adequate weight parameter. Nevertheless, this optimization target lacks flexibility by being able to produce only a single trajectory, rather than a whole Pareto front of possible trajectories as with (\ref{eq:TP:eff:1}). 
We used this approach in \cite{CATP17}. Yet another way to tackle energy-efficient communications-aware trajectories is to define the optimization target as the motion energy and cast the communications objective as constraints. By doing this, the \ac{MR} will reach the communications objective using  the minimum motion energy possible. The complementary form is also possible, where the optimization target is a communications metric and constraints are imposed on the motion energy. In this case, the \ac{MR} efficiently expends a certain amount of energy to optimize the communications metric.
\par
For instance, in \cite{Mob9}, the authors consider the problem of an \ac{MR} that must sense some area while maintaining communication with a \ac{BS}, and design a trajectory to complete this task while maintaining appropriate channel quality throughout the entire trajectory. In \cite{CATP14}, the authors consider the problem of a \ac{MR} that must find a position where the channel gain is high enough to allow for communication with a certain quality, while minimizing the motion energy. In \cite{CATP15}, the authors consider a problem in which a team of \acp{MR} all have a copy of the same message and need to transmit it to a \ac{BS}, which combines the signals received from all of the \acp{MR}. The goal is to maximize the total received signal power. To achieve this, the \acp{MR} need to find positions, using as little energy as possible, where the channel gain is high enough. In \cite{UAV23}, the authors design the path for an energy-limited \ac{UAV} that must reach a certain point while satisfying certain channel constraints to transmit video. In \cite{UAV24}, the authors highlight the importance of considering the wireless communication links for \acp{UAV} when designing paths and considering the communication coverage of the \ac{UAV}'s path. In \cite{LiceaWINCOM2016,LiceaSpringer2017}, we consider the problem of an \ac{MR} that must reach a certain target point and transmit data to a \ac{BS}; the trajectory is optimized so as to spend as little energy in motion as possible while transmitting as much data as possible. The objective function in \cite{LiceaWINCOM2016,LiceaSpringer2017} takes the form of (\ref{eq:TP:eff:1}) where the communication-related term represents the number of bits transmitted by the \ac{MR}. In \cite{CATP19}, we extended this by considering a general communications term that can represent various communications-related criteria.
\subsection{Other Trajectory Design Strategies}
\label{sec:OTDS}

We now discuss some other types of  interesting and exciting  \ac{CaTP} problems, and in the next subsection we briefly present some research opportunities in this field. Additional examples of \ac{CaTP} can be found in \cite{MuralidharanARCRAS2021}.

{\bf i. Trajectory design for Robotic relays}. A robotic relay is an \ac{MR} that relays data between communicating nodes. They can be used to establish communication between two nodes that cannot communicate directly because their direct communication channel is too weak. This can happen if both nodes are far from each other or if there are obstacles between them. In this case, one or more robotic relays can establish communications between the nodes. The advantage of using robotic relays instead of static ones is that they can adapt their positions to maintain or improve the performance of the system, regardless of any change in the environment or in the positions of the nodes. 
\par
In this problem, the objective is to optimize the trajectory of one or more robotic relays to have good end-to-end communication performance. For instance, in \cite{CATP10}, the authors considered the scenario in which a chain of robotic relays must ensure communication between a static robot and an exploring robot. In \cite{CATP11}, the authors addressed the problem of designing trajectories for robotic routers to ensure communication within an indoor environment. In \cite{CATP12}, the authors considered the problem of a fixed-wing \ac{UAV} relay to help a ground robot. In \cite{comm5}, the authors too considered a fixed-wing \ac{UAV} relay between ground nodes and optimized its trajectory to maximize the end-to-end throughput. The authors of \cite{Mob11} developed an experimental platform for robotic relays. In \cite{CATP16}, we optimized the position of a robotic relay to transfer data from various sensors to a fusion center.
\par\bigskip
{\bf ii. Trajectory Design for Data Ferries}. Robotic data ferries operate similarly to robotic relays, except that they move between the nodes to physically transport data. Data ferries are suitable only in scenarios where the application tolerates the delay introduced by the movement of the data ferry. In some cases, the source node produces data continuously, which requires designing a periodic (or quasi-periodic) trajectory. These periodic trajectories are generally optimized by maximizing the end-to-end throughput or maximizing the energy efficiency of the data transfer process (i.e. minimizing the energy cost of transferring one bit of data).
\par
In \cite{CATP8}, the authors designed a closed trajectory for a \ac{UAV} acting as a data ferry to maximize the end-to-end throughput. In \cite{LiceaEUSIPCO2019b}, we studied the problem of a ground \ac{MR} acting as a ferry that must gather data from various sensors and then move towards a fusion center to deliver the data. We studied the tradeoff between the energy consumption of the \ac{MR} and the end-to-end throughput. In \cite{CATP18}, we considered the problem of a \ac{UAV} acting as a data ferry and we optimized its closed trajectory to maximize the end-to-end throughput. 
In \cite{mobb10}, the authors optimize the trajectory of a \ac{UAV} that collects the data from sensor nodes in an energy efficient manner. In \cite{CATP17}, we optimized a 3D trajectory for a quadrotor \ac{UAV} to collect data in an energy efficient manner. In this scenario, as the \ac{UAV}  increases its altitude, the region that it covers grows, but the channel losses also grow. Thus, the \ac{UAV} altitude has to be carefully optimized. 
In \cite{LiuIEEETWC2021}, the authors consider a multirotor \ac{UAV} that must collect data from a sensor network. They optimize the location of the collection points and the 2D \ac{UAV} trajectory to minimize the \ac{AoI} of the data collected. The \ac{AoI} is a {\it freshness measure of the data}, it measures the time since its generation up to its delivery to the intended recipient. In \cite{ChenIEEETVT2023}, the authors consider a similar problem where a \ac{UAV} must collect data from some nodes and then deliver it to a fusion center via a cellular network. Since the cellular network is designed for ground users, it does not have coverage everywhere in the sky, and the \ac{UAV} may thus encounter various coverage holes at the altitude where it operates. In this problem, the authors optimize the \ac{UAV} trajectory to minimize the worst \ac{AoI} from all the nodes.

\par\bigskip
{\bf iii. Communications-aware Trajectory for PV equipped robots}. 
Solar PV panels can be added to \acp{UAV} to extend their flying time. This allows the \acp{UAV} to recharge their batteries through the solar panels. Both the power production of the PV panels and the quality of the communications depend on the \ac{UAV} position. This makes the \ac{CaTP} problem more complex by considering not only the motion and communications aspects but also the energy harvesting aspect as well.
\par
In \cite{UAV11}, the authors optimize the 3D trajectory of a \ac{UAV} equipped with a PV solar panel that must provide communications, see section \ref{sec:ExamplesDetails} for details. 
A variation of this problem is presented in \cite{UAV21} where the \ac{UAV} can sell the energy generated and the users are charged by the amount of data processed by the \ac{UAV}. The objective is to optimize the \ac{UAV} trajectory to maximize the revenue. The energy and data transmitted are both translated into money, and thus the optimization target representing the revenue is a weighted sum of the data received by the UAV from the users, the energy consumed, and the energy sold. 
The authors in \cite{UAV5} optimize the behaviour of \acp{UAV} which are equipped with transceivers and PV panels to minimize the overall energy consumption of a heterogeneous communications network.
\par\bigskip
{\bf iv. Trajectory Design for Small-Scale Fading Compensation}. The small-scale fading provokes large variations in the channel gain over small distances (compared to the wavelength used) and they can significantly degrade the performance of the communication system, see section \ref{Comm:channel}. Thus, it is important to compensate for this phenomenon. One way to compensate for it is to control the position of the robot's antenna by controlling the position of the robot itself. This way, the robots can leverage the spatial variations of the small-scale fading to improve the experienced channel gain. This problem can take two forms. In the first form, the objective is to find a static position from which the robot will transmit its data. The robot explores a small region to find a suitable point for transmission. In its second form, the robot must follow a certain path and can only control its linear velocity. In this case, a speed controller is designed to compensate for the small-scale fading. The strategy of such controllers is to reduce the speed when the channel gain is high and increase it otherwise, see \cite{CATP9}.
\par
In \cite{MD3}, the authors experimentally control the position of the robot to compensate for the small-scale fading. In \cite{MD4}, the authors also experimentally demonstrate this same principle using an antenna mounted on a turntable. In \cite{BonillaIEEETSP2016}, we considered a ground \ac{MR} harvesting \ac{RF} energy and developed a trajectory planner that exploits small-scale fading to maximize the energy harvested. The path explored by the \ac{MR} was restricted to a straight line whose length is optimized. In \cite{LiceaEUSIPCO2018}, we extended this trajectory planner to optimize the shape of the path. In \cite{MD1}, we developed a predetermined trajectory planner for a ground \ac{MR} to find points with high channel gain in its vicinity in an energy-efficient manner. In this technique, the \ac{MR} explores a set of stopping points to determine the best position for communication. 
We expanded upon this work in \cite{MD2} by making the trajectory planner adaptive. At each time, the next point to be visited by the \ac{MR} is calculated as a function of the channel measured in the current and previous points. This technique outperforms the technique presented in \cite{MD1}, but requires an estimate of the shadowing as well. We further investigated this technique in \cite{LiceaEUSIPCO2019} by considering an \ac{MR} equipped with an antenna mounted on a rotary platform that follows a predefined trajectory while communicating with a \ac{BS}. We developed a feedback controller that continuously optimizes the position of the antenna under time-varying fading. In \cite{CATP16,LiceaEUSIPCO2019b}, we optimize the position of a ground \ac{MR} to compensate for the small-scale fading of multiple communication channels simultaneously.
\par\bigskip
{\bf v. Communications-aware UAV Placement}. 
In Communications-aware \ac{UAV} placement problems, the focus is on the stopping (or final) positions of the \ac{UAV} to provide communication to ground users, rather than on the full trajectory. In some cases, the energy required by the UAV to reach a convenient final position is also of interest. 
For these problems, the wireless channel model usually considers the effect of the altitude and uses similar models to the ones described in section \ref{A2GCommChannels}. 
It is expected that drones will become a part of the cellular networks to improve their performance, and thus this particular problem is of interest. Even if this is not a \ac{CaTP} problem, it requires similar tools and models. 
\par
In \cite{CATP13}, the problem of positioning \acp{UAV} for improving a cellular network is considered. In \cite{UAV16}, the authors optimize the altitude of a \ac{UAV} to provide service to a cellular network. In \cite{UAV17}, the authors optimize the 3D position of a \ac{UAV} to maximize the number of covered users. In \cite{UAV18}, the authors optimize the 3D position of \ac{UAV} to maximize the number of users covered by the network. In \cite{UAV19}, the authors optimize the position of a \ac{UAV} providing communications to a user inside a vertical building. In \cite{UAV22}, the authors propose a distributed algorithm to iteratively optimize the position of a team of \ac{UAV} providing service to users in a network.
\par
{\bf vi. UAVs and physical layer security}. A recent application is the implementation of physical layer techniques \cite{RodriguezIEEECM2015} through the use of \ac{UAV} to solve cybersecurity issues \cite{XiaofangIEEEWC2019}. One example is the joint optimization of the \ac{UAV}'s trajectory and communications strategy to improve the security of the network against eavesdroppers. The strategy is to use one or more \acp{UAV} to either (i) jam the eavesdropper to degrade its \ac{SNR} or (ii) act as relays between the \ac{BS} and the targeted node so that the latter reduces its transmission power, thus decreasing the strength of the signals captured by eavesdroppers.

In \cite{BonillaICASSP2024} we optimized the position of an omnidirectional \ac{UAV}, acting as relay, which was under attack attack by a malicious jammer. In \cite{XiaofangGlobecom2018}, the authors consider 
a source node that must transmit data to a destination node via a \ac{UAV} acting as a relay, but an eavesdropper is present. They jointly optimize the \ac{UAV} trajectory and the communications parameters to maximize the secrecy rate. They  degrade the communications channel between the \ac{UAV} and the eavesdropper while improving the quality of the end-to-end legitimate communications link. 
In \cite{LiIEEEWCL2019}, the authors optimize the trajectory and communication parameters of a \ac{UAV} that operates as a jammer. 
The \ac{UAV} jammer degrades the \ac{SINR} at an eavesdropper while minimizing the impact on the legitimate nodes. Similar cybersecurity \ac{CaTP} problems are discussed in \cite{YunlongIEEEJSAC2018, ZhouIEEEToVT2018, LiuIEEEToWC2019, XiaofangICCW2019}.

\subsection{Research opportunities} Some of the key technologies for { \ac{UAV} communications in 6G are\footnote{See \cite{GeraciIEEECST2022} and the references therein for a more detailed discussion on these technologies in the context of \ac{UAV} communications.} mmWave \ac{UAV} communications, THz \ac{UAV} communications, and \ac{UAV} communications with Reconfigurable Intelligent Surfaces (RIS). These technologies create new research opportunities in \ac{CaTP} for \acp{UAV}.

mmWave \ac{UAV} communications uses highly directional antennas. This high directionality makes those communications links very sensitive to antenna misalignments. Thus, the tilting that underactuated \acp{UAV} experience when moving, see Fig. \ref{fig:UAV:motion}, creates a challenge that must be overcome if we plan to use mmWave communications in flying \acp{UAV}. This requires the development of new trajectory planning techniques, adapted to the large bandwidth of mmWave communications and focused in the antenna misalignment caused by the \ac{UAV} tilting, similar to the one studied in \cite{BonillaEUSIPCO21}.

THz communications uses extremely small wavelengths and large bandwidths. These extremely small wavelengths allow to equip small \acp{UAV} with antenna arrays. This calls for the development of new techniques for the joint optimization of trajectory planning and beamforming for \acp{UAV} equipped with antenna arrays.

RIS are passive planar structures electronically manipulated that control how they reflect the electromagnetic waves. According to \cite{GeraciIEEECST2022}, there is still a lack of trajectory planning techniques for \acp{UAV} equipped with RIS to improve the communications network.

Omnidirectional \acp{UAV} \cite{TognonIEEERAL2018} is another interesting technology that opens new research opportunities in \ac{CaTP}. These \acp{UAV} can control its position and orientation independently, as opposed to underactuated multirotor \acp{UAV} which are the focus of most of the current research in \ac{CaTP} nowadays. This new capability introduced by omnidirectional \acp{UAV} can be exploited in creative ways to develop new techniques such as making a \ac{UAV} orient itself in the air to direct the nul of its antenna radiation pattern towards a jammer to neutralize its attack, see \cite{BonillaICASSP2024}.

Finally, all the \ac{CaTP} described in the previous sections can be revisited using these new technologies while considering both robotics and communications aspects to create new and exciting developments.}

\subsection{Useful numerical tools to solve \ac{CaTP} problems}
\label{sec:solutions}
{Our focus in this paper is on the formulation and design of the \ac{CaTP} problem, with limited discussion on resolution methodologies. This is attributed to the inherent variability of the \ac{CaTP} problem, which assumes a diverse array of forms.
 However, for the sake of completeness, we briefly discuss software tools and algorithms that prove beneficial in addressing various instances of \ac{CaTP} problems.

The first tool is CVX \cite{cvx} which is a Matlab-based modeling system for convex optimization. It can be used to solve some standard convex problems such as linear programming, quadratic programming and even \ac{MIP} among other types of convex problems. CVX can be used with various solvers such as MOSEK \cite{mosek} and Gurobi \cite{gurobi}. A comparable tool for convex optimization is CVXPY \cite{diamond2016cvxpy}, designed for Python. Another optimization tool is CPLEX \cite{cplex2009v12} which can work with C/C++ as well as with Python. CPLEX can be used to solve linear programming problems, quadratic programming problems as well as \ac{MIP} among others. 

When the \ac{CaTP} problem is convex and the optimization target is a function, the aforementioned tools can be employed. If the optimization target is a function, but the \ac{CaTP} problem is non-convex, then we can rely on successive convex approximation or on heuristic algorithms such as simulated annealing or genetic algorithms to solve the problem.

Finally, when the optimization target of the \ac{CaTP} problem is a functional and the optimization problem involves optimizing a function instead of a set of variables, then CasADi \cite{CasADi}, an open-source tool for nonlinear optimization and algorithmic differentiation, could be helpful. CasADi is a tool commonly used in optimal control problems. 
Alternatively, reinforcement learning, applicable to control problems \cite{BUSONIU20188}, can be implemented using dedicated Python libraries {such as TensorFlow or PyTorch}.
} 

\section{\ac{CaTP} Summary and Final Comments}
\label{sec:summary}
We conclude this tutorial by presenting a mathematical formalization of the \ac{CaTP} description (\ref{TPA:eq:1}). This will help the reader to understand the mathematical structure of \ac{CaTP} problems and can be used as a guide to formulate a specific \ac{CaTP} problem. In addition, this section constitutes a high-level map that indicates in which part of the tutorial is discussed each element of the \ac{CaTP} problem.
\par
\begin{eqnarray}
\label{formalization:1}
    \minimize_{\mathcal{C},\mathcal{T}}&\quad& J(\mathcal{C},\mathcal{T})
\end{eqnarray}
s.t.
\begin{subequations}
\begin{eqnarray}
\label{formalization:2a}
  \dot{\mathbf{x}} & = &\mathbf{f}_{\rm dyn}(\mathbf{x},\mathcal{T}),\\
  \label{formalization:2d}
  \mathbf{f}_{\rm kyn}(\mathbf{x}) &=&\boldsymbol{0},\\
  \label{formalization:2dd}
  \mathbf{g}_{\rm kyn}(\mathbf{x}) &\leq&\boldsymbol{0},\\
  \label{formalization:2e}
  \mathbf{f}_{\rm lim}(\mathcal{T})&=&\boldsymbol{0},\\
    \label{formalization:2ee}
  \mathbf{g}_{\rm lim}(\mathcal{T})&\leq&\boldsymbol{0},\\
  \label{formalization:2f}
  E_{\rm mech}&=&{f}_{\rm mech}(\mathbf{x},\mathcal{T}),
\end{eqnarray}
\end{subequations}
\begin{subequations}
\begin{eqnarray}
\label{formalization:3a}
\mathbf{r}&=&\mathbf{f}_{\rm chan}(\mathbf{s}(\mathcal{C}),\mathbf{p},\mathbf{q},\mathbf{n}_{com};\mathbf{h},t,),\\
\label{formalization:3d}
\mathbf{f}_{\rm rep}(\mathbf{h})&=&\mathbf{f}_{\rm desc}(\mathbf{p},\mathbf{q},t)\\
\label{formalization:3e}
\mathbf{f}_{\rm rep}(\mathbf{n}_{com})&=&\mathbf{f}_{\rm nc},\\
\label{formalization:3b}
\boldsymbol{\Gamma}&=&\mathbf{f}_{\rm pr}(\mathbf{p},\mathbf{q},\mathbf{h},t,\mathcal{C}),\\
\label{formalization:3c}
\mathbf{d}_{\rm rx}&=&\int_0^{t_f}\mathbf{f}_{\rm rate}(\boldsymbol{\Gamma},\mathcal{C})\mathrm{d}t,\\
\label{formalization:3f}
  \mathbf{f}_{lim}(\mathcal{C})&\leq&\boldsymbol{0},\\
\label{formalization:3g}
E_{\rm comm}&=&{f}_{\rm comm}(\mathbf{\Gamma},\mathcal{C}),
\end{eqnarray}
\end{subequations}
\begin{subequations}
    \begin{eqnarray}
    \label{formalization:4a}
     \mathbf{f}_{\rm traj}(\mathbf{x},\mathcal{T})&=& \boldsymbol{0},\\
         \label{formalization:4b}
     \mathbf{g}_{\rm traj}(\mathbf{x},\mathcal{T})&\leq& \boldsymbol{0},
    \end{eqnarray}
\end{subequations}
\begin{subequations}
    \begin{eqnarray}
     \label{formalization:5a}
     \mathbf{f}_{\rm comm}(\mathbf{r},\boldsymbol{\Gamma},\mathbf{d}_{rx},\mathbf{p},\mathbf{q},\mathcal{C})&=& \boldsymbol{0},\\
     \label{formalization:5b}
     \mathbf{g}_{\rm comm}(\mathbf{r},\boldsymbol{\Gamma},\mathbf{d}_{rx},\mathbf{p},\mathbf{q},\mathcal{C})&\leq& \boldsymbol{0},
    \end{eqnarray}
\end{subequations}
\begin{eqnarray}
   \label{formalization:6} 
    \mathbf{g}_{en}(E_{\rm mech},\dot{E}_{\rm mech},E_{\rm comm},\dot{E}_{\rm comm})\leq\mathbf{0}.
\end{eqnarray}

We divide (\ref{formalization:1})-(\ref{formalization:5b}) into the optimization target (\ref{formalization:1}), discussed in section \ref{sec:optimizeationTarget}, and five blocks.
\par
The first block (\ref{formalization:2a})-(\ref{formalization:2f}) constitutes the model of the robot. In (\ref{formalization:2a}), the function $\mathbf{f}_{dyn}$ describes the dynamic model of the robot, and $\mathbf{x}$ is the state vector; see sections \ref{sec:MR:Dyn},  \ref{sec:uav}, and \ref{sec:fixedwing}. In placement problems, discussed in section \ref{sec:OTDS}, the dynamic model is not needed.
\par
The equality constraint (\ref{formalization:2d}) describes the forward kinematic model of the robot; see section \ref{sec:MR:Kin}. 
While  (\ref{formalization:2dd}) represents the inequality kinematic constraints. For example, in a wheeled ground \ac{MR}, depending on its physical configuration, the steering angle might be restrained to some range. These kinematic inequality constraints are not discussed in this tutorial.  If the kinematic constraints of the particular robot are not relevant to the problem, then we can put (\ref{formalization:2d}) and (\ref{formalization:2dd}) aside. 
\par
The constraints (\ref{formalization:2e}) and (\ref{formalization:2ee}) directly act on the input related to the robot's control. Section \ref{sec:trajConstraints} provides some examples; see (\ref{TPA:eq:2b}). 
\par
Equation (\ref{formalization:2f}) is the mechanical energy consumption model of the robot; see sections \ref{Rob:ener}, \ref{sec:uav}, and \ref{sec:fixedwing}. This model is not always necessary, but it becomes essential if we are interested in the total operation time of the robot or we are interested in energy efficiency.
\par
The second block (\ref{formalization:3a})-(\ref{formalization:3g}) constitutes the model of the communications channel. In (\ref{formalization:3a}), $\mathbf{r}$ is the signal (or signals) received, $\mathbf{s}(\mathcal{C})$ is the signal (or signals) transmitted which depends on the communications parameters $\mathcal{C}$, $\mathbf{p}$ and $\mathbf{q}$ are the poses (position and attitude) of the transmitter and of the receiver respectively, $\mathbf{n}_{\rm comm}$ is the noise generated at the receiver, $\mathbf{h}$ represents the small-scale fading and the shadowing,  and $t$ is the time. If the communications channel is time-invariant we drop the dependency on $t$ in the function $\mathbf{f}_{\rm chan}$. Note that $\mathbf{f}_{\rm chan}$ is a function, parameterized by $\mathbf{h}$ and $t$, which takes as input the signal transmitted $\mathbf{s}(\mathcal{C})$, the poses of the transmitter(s) and the receiver(s) and the noise $\mathbf{n}_{\rm comm}$ to calculate the signal(s) received $r$; see section \ref{Comm:channel} for more information.
\par
Equation (\ref{formalization:3d}) describes the model of the shadowing and/or small-scale fading, i.e. $\mathbf{h}$. If the channel model selected is stochastic, then the function $\mathbf{f}_{\rm rep}$ is the statistical representation chosen for $\mathbf{h}$, and $\mathbf{f}_{\rm desc}$ is the actual description for $\mathbf{f}_{\rm rep}(\mathbf{h})$. For example, let $h\in\mathbb{R}$ be the small-scale fading, if  we represent $h$ with its mean and its variance, then $\mathbf{f}_{\rm rep}({h})=[\mathbb{E}[h],\quad \mathrm{Var}[h]]^{\mathrm{T}}$. Thus,  $\mathbf{f}_{\rm desc}$ would be the formulas of the vector $[\mathbb{E}[h],\quad \mathrm{Var}[h]]$. We could also choose to represent $h$ with its \ac{PDF} and its temporal correlation. 
If we are not using a stochastic model and $h$ is time-invariant, then we could represent $h$ with a 2D \ac{RF} map. In this case, we could have $\mathbf{f}_{\rm rep}({h})=h(\mathbf{p},\mathbf{q})$ and $\mathbf{f}_{\rm desc}$ could be a numerical function that returns the corresponding value of $h$ for each pair $\{\mathbf{p},\mathbf{q}\}$, see section \ref{Comm:channel}. If the channel model used in (\ref{formalization:3a}) disregards the small-scale fading and the shadowing, then (\ref{formalization:3d}) is not required; this occurs when the channel model used is deterministic; see section \ref{Comm:channel:deterministic}.  
\par
Equation (\ref{formalization:3e}) is the statistical description of the noise generated at the receiver. It acts similarly as (\ref{formalization:3d}) with the exception that its description is always the \ac{PDF} and the temporal autocorrelation. This model is rarely explicitly written in the equations, and the authors rather describe it in the text, e.g. saying that the noise is white Gaussian with zero-mean and variance $\sigma^2$.
\par
In (\ref{formalization:3b}), $\Gamma$ is the \ac{SNR}; {see section \ref{Comm:metrics} (or \ac{SINR} like in Example \ref{example3} of section \ref{sec:ExamplesDetails}) at the receiver,} (\ref{example3:SINR}) is one instance of the function $\mathbf{f}_{pr}$. (\ref{formalization:3b}) can be calculated from (\ref{formalization:3a}), and so it is not essential to add it if  (\ref{formalization:3a}) is already included in the model. It is also possible to include (\ref{formalization:3b}) and leave out (\ref{formalization:3a}); see Example \ref{example3} in section \ref{sec:ExamplesDetails}.
\par
In (\ref{formalization:3c}), $\mathbf{d}_{\rm rx}$ is the data received, and $\mathbf{f}_{\rm rate}$ is the data rate which is related to the channel capacity; see section \ref{Comm:metrics}. Here, {\it data} can be the number of bits received or the number of successful transmissions/packets; see Examples \ref{example1} and \ref{example3} in section \ref{sec:ExamplesDetails}. 
 \par
The constraint (\ref{formalization:3f}) acts directly on the communications-related inputs; see section \ref{sec:generalStructure}.
 \par
 The constraint (\ref{formalization:3g}) describes the energy consumption model for the communications system; see section \ref{sec:comm:energ}. If the energy consumption of the communications system is significantly inferior to the energy consumed by the motion of the robot, then we might drop this model. 
 \par 
The third block (i.e., (\ref{formalization:4a}) and (\ref{formalization:4b})) and the fourth block (i.e., (\ref{formalization:5a}) and (\ref{formalization:5b})) constitute the constraints introduced by the designer, excluding the constraints due to physical reasons, on the trajectory/path (see section \ref{sec:generalStructure}) and on the communications system (see section \ref{sec:generalStructure}) respectively.
\par 
The fifth block (\ref{formalization:6}) can also be added to limit the energy and/or power consumption of the system. It is possible to  have a single inequality constraint which will limit the total energy consumption or the total power consumption as in Example \ref{example4} in section \ref{sec:ExamplesDetails}. It is also possible to limit the total energy consumption and the total power consumption simultaneously. It is also possible to introduce inequality constraints to limit the transmission power of the communications system (e.g., (\ref{eq:powerConstraint1}) and (\ref{eq:powerConstraint2}) in section \ref{sec:CommConstraints})), and separately introduce another inequality to limit the power consumption of the robot motors.
\par
The \ac{CaTP} problem formulation presented above does not intend to be a {\it one model fits all}, but rather a guide to the reader about how to mathematically formulate their \ac{CaTP} problem. Depending on the specific needs, this formulation would require adjustments. For instance, (\ref{formalization:1})-(\ref{formalization:5b}) is formulated in continuous time, but an equivalent discrete-time formulation is also possible; see Example \ref{example4}.
The trajectory planning problem (\ref{formalization:1})-(\ref{formalization:5b}) can be considered a \ac{CaTP} problem if all of the following conditions are satisfied: i) either (\ref{formalization:3a}) or (\ref{formalization:3b}) are present; ii) the function $\mathbf{f}_{\rm chan}$, or the function $\mathbf{f}_{\rm pr}$ if the former is not present, should depend on the position of the robot; iii) the cost function (\ref{formalization:1}) depends on $\mathcal{C}$ or there is at least one constraint in (\ref{formalization:5a})-(\ref{formalization:5b}) that depends on $\mathcal{C}$.
These conditions establish the bridge that links the robotic block (i.e., (\ref{formalization:2a})-(\ref{formalization:2f}) and (\ref{formalization:4a})-(\ref{formalization:4b})) with the communications block (i.e., (\ref{formalization:3a})-(\ref{formalization:3g}) and (\ref{formalization:5a})-(\ref{formalization:5b})). In other words, a \ac{TP} problem is a \ac{CaTP} problem if it includes a communications channel model that depends on the position of the agent and if at least one communications-related term appears either in the optimization target or in any constraint.
}

\section{Discussion and Concluding Thoughts}
\label{sec:Discussion}
The growing interest for merging ground \acp{MR} and \acp{UAV} with communication systems is gaining momentum in both industry and academia. One of the main reasons for this occurrence is the surge of 5G technologies, which aim to integrate \acp{UAV} into the cellular networks. Another important reason involves the growing popularity of multi-robot systems where the \acp{MR} must maintain communication links. In these applications, the communications and robotics aspects are closely interrelated. In section \ref{sec:Intro}, we discussed some of the consequences of addressing these problems without an interdisciplinary approach. 
\par
Due to the strong entanglement between the robotics and communications aspects, an interdisciplinary approach is fundamental to fully exploit the maximum potential of this research area. In the hope of contributing to the quantity of material and quality of this research area, this tutorial provides the basic theory behind this still underdeveloped, yet promising research area. We have described different motion models for various types of ground and aerial \acp{MR}, as well as various channel models for different scenarios relevant to the \ac{CaTP} problems. Additionally, we have provided a general mathematical formulation for \ac{CaTP} problems for predetermined trajectories. Different ways have been shown in which the robotic motion models and the communications channel models can interact with each other within the formulation of \ac{CaTP} problems. Finally, we have provided a brief application-oriented classification of different \ac{CaTP} problems and other related problems.

 \begin{IEEEbiography}[{\includegraphics[width=1in,height=1.25in,clip,keepaspectratio]{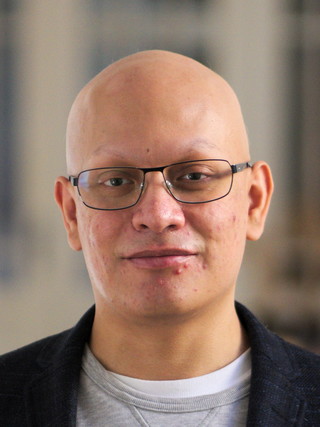}}]{Daniel Bonilla Licea} received his M.Sc. degree in 2011 from the Centro de Investigaci\'on y Estudios Avanzados (CINVESTAV), Mexico City. From May 2011 until June 2012, he worked as an intern in the signal processing team of Intel Labs in Guadalajara, Mexico. He received his PhD degree in 2016 from the University of Leeds, U.K. In 2016, he was invited for a short research visit at the Centre de Recherche en Automatique de Nancy (CRAN), France. In 2017, he collaborated in a research project with the Centro de Investigaci\'on en Computacion (CIC) in Mexico. From 2017 to 2020, he held  a postdoctoral position at the International University of Rabat, Morocco. From 2020 to 2023, he held a full-time a postdoctoral position at the Czech Technical University in Prague, Czech Republic. Currently he is a full-time assistant professor at the Mohammed VI Polytechnic University, Morocco, and a part-time senior researcher at Czech Technical University in Prague, Czech Republic. His main research interests are signal processing and communications-aware robotics. 
 \end{IEEEbiography}
\begin{IEEEbiography}[{\includegraphics[width=1in,height=1.25in,clip,keepaspectratio]{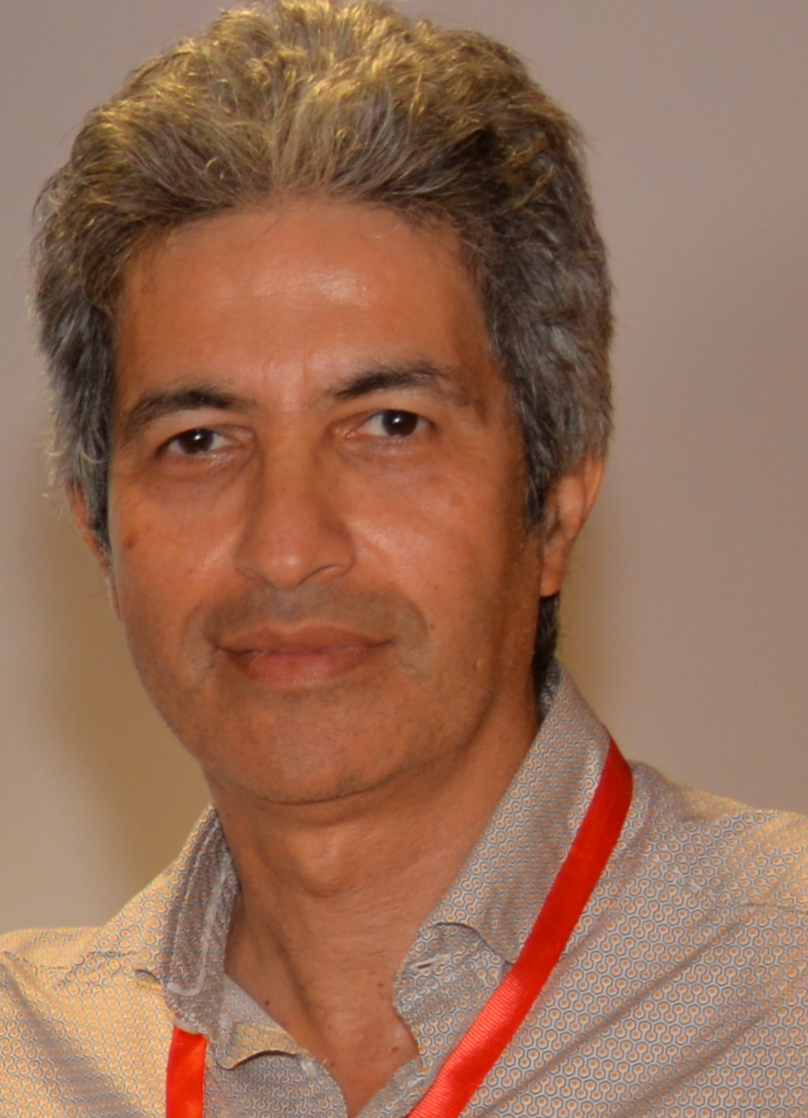}}]{Mounir Ghogho }
Mounir Ghogho (F’18) has received the PhD degree in 1997 from the National Polytechnic Institute of Toulouse (France). He was an EPSRC Research Fellow with the University of Strathclyde (Scotland) from Sept 1997 to Nov 2001. In Dec 2001, he joined the University of Leeds (England), where he was promoted to full Professor in 2008. While still affiliated with the University of Leeds, in 2010 he joined the International University of Rabat (UIR) where he is currently Dean of the College of Doctoral Studies and Director of TICLab (ICT Research Lab). He is a Fellow of IEEE and AAIA (Asia-Pacific AI Association), a recipient of the 2013 IBM Faculty Award, and a recipient of the 2000 UK Royal Academy of Engineering Research Fellowship. His research interests are in Machine Learning, Signal Processing and Wireless Communication. He has coordinated around 20 research projects and supervised over 50 PhD students in the UK and Morocco. He has served as Associate Editor of many journals including the IEEE Signal Processing Magazine and the IEEE Transactions on Signal Processing. He has also been a member of The Signal Processing Theory and Methods (SPTM) and The Sensor Array and Multichannel (SAM) technical committees of the IEEE Signal Processing Society. He has chaired many international conferences and workshops including the IEEE 11th Workshop on Signal Processing for Advanced Wireless Communication (SPAWC 2010) and the 21st European Signal Processing Conference (EUSIPCO 2013).
\end{IEEEbiography}
\begin{IEEEbiography}[{\includegraphics[width=1in,height=1.25in,clip,keepaspectratio]{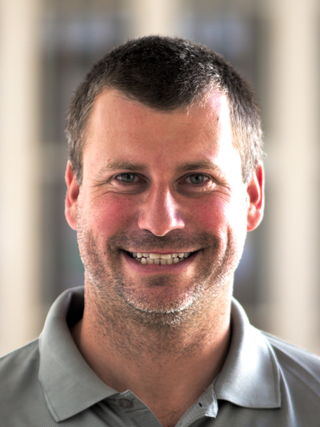}}]
{Martin Saska} received his Ph.D. degree at University of Wuerzburg, Germany, within PhD program of Elite Network of Bavaria. He founded and heads the Multi-robot Systems lab (http://mrs.felk.cvut.cz/) at Czech Technical University in Prague with more than 40 researchers. He was a visiting scholar at University of Illinois at Urbana-Champaign and at University of Pennsylvania, USA. He is co-author of >200 publications in conferences and impacted journals, including IJRR, AURO, JFR, ASC, EJC, with >6800 citations indexed by Scholar and H-index 45. His team won multiple robotic challenges in MBZIRC 2017, MBZIRC 2020 and DARPA SubT competitions (http://mrs.felk.cvut.cz/projects/mbzirc, http://mrs.felk.cvut.cz/mbzirc2020,   http://mrs.felk.cvut.cz/projects/darpa). 
\end{IEEEbiography}
\bibliographystyle{IEEEtran}
\bibliography{main.bib}

\end{document}